\documentclass[11pt,letterpaper]{article}
\usepackage[margin=1in]{geometry}
\usepackage[T1]{fontenc}
\usepackage{lmodern,microtype}
\usepackage{amsmath,amssymb,amsthm,mathtools}
\usepackage{booktabs,tabularx,array}
\usepackage{enumitem}
\providecommand{\citep}[1]{\cite{#1}}
\providecommand{\citet}[1]{\cite{#1}}
\usepackage{xcolor}
\usepackage{graphicx}
\usepackage[hypertexnames=false,colorlinks=true,linkcolor=blue!55!black,citecolor=blue!55!black,urlcolor=blue!55!black]{hyperref}
\hypersetup{pdftitle={Computational Complexity of Data-Driven Configuration for Integer Programming, Clustering, and Learning Algorithms},pdfauthor={Dravyansh Sharma},pdfsubject={STOC DDAD manuscript}}
\usepackage{tikz}
\usetikzlibrary{arrows.meta,positioning,calc}
\setlist{nosep,leftmargin=1.4em}
\newtheorem{theorem}{Theorem}[section]
\newtheorem{lemma}[theorem]{Lemma}
\newtheorem{proposition}[theorem]{Proposition}
\newtheorem{corollary}[theorem]{Corollary}
\newtheorem{definition}[theorem]{Definition}
\newtheorem{remark}[theorem]{Remark}
\newcommand{\F}{\mathcal F}
\newcommand{\X}{\mathcal X}

\newcommand{\bits}{\{0,1\}}
\newcommand{\fat}{\operatorname{fat}}
\newcommand{\Pdim}{\operatorname{Pdim}}
\newcommand{\Tree}{\operatorname{TreeSize}}
\newcommand{\OPT}{\operatorname{OPT}}
\newcommand{\ALG}{\operatorname{ALG}}
\newcommand{\osc}{\operatorname{osc}}
\newcommand{\one}{\mathbf 1}
\newcommand{\poly}{\operatorname{poly}}
\newcommand{\E}{\mathbb E}
\newcommand{\ERM}{\operatorname{ERM}}

\newcommand{\Enc}{\operatorname{Enc}}
\newcommand{\Unsat}{\operatorname{Unsat}}
\newcommand{\ind}{\mathbf{1}}
\newtheorem{recipe}[theorem]{Recipe}

\DeclareMathOperator*{\argmin}{arg\,min}

\title{Computational Complexity of Data-Driven Configuration for Integer Programming, Clustering, and Learning Algorithms}
\author{Dravyansh Sharma\thanks{Most of this work was carried out while the author was at Toyota Technological Institute at Chicago (TTIC).}\\Microsoft}
\date{}
\usepackage{longtable}
\usepackage{bm}
\usepackage{alphalph}

\newenvironment{insight}{\begin{quote}\textbf{Intuition.} }{\end{quote}}

\begin{document}
\pagenumbering{roman}
\setcounter{page}{1}
\maketitle
\thispagestyle{plain}
\begin{abstract}
Data-driven algorithm design selects configurations for application-specific inputs, but computing an empirically optimal configuration can itself be difficult. We study this problem for integer programming, clustering, and learning algorithms. For one rank-one Chv\'atal--Gomory cut, choosing shared multipliers across integer programs is NP-complete and exhibits constant-additive approximation hardness. Choosing a power-mean linkage exponent that realizes a prescribed root split (target 2-cluster) is NP-hard. Tuning one scalar Lasso penalty to achieve zero standard squared validation error on a compact interval bounded away from zero is NP-complete.  We establish NP-hardness of constant-step-size gradient-descent ERM on piecewise-quadratic objectives and statistical capacity lower bounds for step sizes, schedules, and initialization. We complement our worst-case hardness results with some positive results under smoothed analysis. Independent Gaussian design perturbations yield polynomial expected Lasso path length, we obtain sharper bounds than Li and Singer. In power-mean linkage clustering, independent bounded-density distances yield polynomial expected number of distinct cluster trees and oracle-efficient ERM. Conversely, an enhanced aggregated-CG construction  remains hard under independently perturbed nonzero coefficients. 
\end{abstract}
\clearpage
\pagenumbering{arabic}
\section{Introduction}
Automatic algorithm configuration replaces manual trial and error with
choosing algorithm parameters from representative problem instances. This
view was given a statistical foundation by Gupta and Roughgarden
\cite{GuptaRoughgarden2017,GuptaRoughgarden2020} and developed in the
broader study of data-driven algorithm design
\cite{BalcanDDADChapter2021}.  A parameter may govern a mixed-integer solver's branching
or cutting strategy, a clustering rule's linkage exponent, the regularization
strength of a regression estimator, or the learning rate of an iterative
optimization method.  Such parameters change the computation an algorithm performs,
the solution it returns, and its running time.  Even a one-dimensional parameter can have a substantial
impact because its effect is amplified by a sequence of adaptive decisions.

ParamILS
showed that automatic search over solver configurations can improve SAT and
mixed-integer programming performance~\cite{HutterHoosLeytonBrownStutzle2009}.
Model-based configuration and racing methods provide alternatives to manual
trial and error~\cite{HutterHoosLeytonBrown2011,LopezIbanez2016}.
In machine learning, automatic selection of models and preprocessing pipelines,
as well as hyperparameter optimization and resource-aware evaluation, are
central parts of AutoML~\cite{FeurerEtAl2015,LiEtAl2018}.
Generalization bounds measure how many instances are needed to select a
reliable configuration. Our question is the computational effort required
to find an empirically optimal configuration once those instances are given.

To formalize the question, let an algorithm $A_\theta$ with configuration
$\theta\in\Theta$ act on an instance $x$, and write
\[
\F=\{f_\theta:x\mapsto\ell(x,A_\theta(x)):\theta\in\Theta\}.
\]
Given a training collection $S=(x_1,\dots,x_m)$, empirical risk minimization
(ERM) chooses
\[
\widehat\theta\in\argmin_{\theta\in\Theta}\ \widehat L_S(\theta),\qquad
\widehat L_S(\theta)=\frac1m\sum_{i=1}^m f_\theta(x_i).
\]
The empirical average models a configuration used repeatedly on future
instances drawn from the same distribution.  Data-driven algorithm design (DDAD) addresses generalization through
pseudodimension, fat-shattering, and the structure of algorithm-dependent
loss functions~\cite{GuptaRoughgarden2017,prior:balcan2017,prior:balcan2018,prior:balcannguyensharma2023}.
Computing an
empirical minimizer depends additionally on the structure of the loss as
the configuration varies. Output-sensitive ERM methods process
constant-behavior regions of parameter space~\cite{BalcanSeilerSharma2024},
which can be exponentially numerous even for one numerical parameter.

For a real-valued function family $\mathcal F$, its pseudodimension
$\Pdim(\mathcal F)$ measures the largest number of threshold-labeled
instances it can shatter. Its margin-sensitive analogue
$\fat_\gamma(\mathcal F)$ requires a gap of at least $\gamma$ from
each threshold. Both measure sample complexity rather than the
computational effort required for exact ERM.

\paragraph{From decision sequences to hard empirical optimization.}
A scalar parameter may lie in one of $2^T$ short intervals, each
corresponding to a $T$-bit decision sequence. When the empirical loss
separates into one test per bit, minimization is coordinatewise. If
training instances can test pairs or triples of bits through the
algorithm's ordinary operation and cost, minimizing their total loss
can encode \textsc{Max-2-SAT} or \textsc{3-SAT}. Establishing this
connection requires a construction of the algorithm's actual behavior
and a soundness proof for every candidate parameter.

\paragraph{Main computational hardness results.}
We establish the following results for standard algorithmic operations.
\begin{enumerate}[label=\textbf{H\arabic*.},itemsep=3pt]
\item \textbf{One aggregated cutting plane with small coefficients.}
When the \emph{same vector of multipliers} must work on many tiny integer
programs, deciding whether a standard rank-one CG cut closes every
root relaxation gap is NP-complete. Coefficients are at most eight.
Each individual IP is simple, while the shared multipliers must satisfy
constraints across the training collection
(Theorem~\ref{thm:singlecg:main}).
\item \textbf{One Lasso parameter with standard validation loss.}
For full-rank rational regression, deciding whether some regularization
strength on a compact interval bounded away from zero attains zero
standard squared validation error is NP-complete.
The proof builds an exponentially long regularization path and uses
dual conditions to show that \emph{every} zero-loss configuration
encodes a valid partition (Theorem~\ref{thm:lasso-ordinary-validation}).
The construction uses exponentially separated rational coefficients.
\item \textbf{An unmodified hierarchical clustering rule.}
For power-mean linkage on a rational metric with distances in $[1,2]$,
it is NP-hard to choose an exponent achieving a target root split.
Two growing clusters read assignment choices, and a failed clause
forces an irreversible merge across the required split
(Theorem~\ref{thm:linkage:main}). The assignment clock uses
exponentially fine offsets.
\item \textbf{Scalar-step-size gradient descent.}
For capped gradient-norm stopping time on specified two-dimensional,
piecewise-quadratic, potentially discontinuous objectives, exact ERM over
one scalar step size is NP-hard (Theorem~\ref{thm:gd-summary}). Separate
statistical capacity lower bounds cover step sizes, schedules, and
initialization. The four-example ReLU capacity construction is continuous
and does not establish ERM hardness for that neural training problem.
\end{enumerate}
\paragraph{Precision and tractability.}
In a one-variable batch-cut model, worst-case tuning with large integer
coefficients is NP-hard, while a pseudo-polynomial algorithm and an
expected-polynomial algorithm under independent relative noise identify
a tractable regime (Theorem~\ref{thm:scabs:overview}). For Lasso,
Gaussian perturbations yield a polynomial bound on expected path length
(Theorem~\ref{thm:lasso:kac-rice}). The main constructions use integer
rounding for cutting planes, sign patterns and dual conditions for Lasso,
and irreversible merges for linkage. An expanding gradient-descent map also gives both capacity lower bounds and an ERM-hardness construction for a scalar step size. Table~\ref{tab:first-results} summarizes the
results and the respective configuration models.

\begin{table}[htbp]\centering\small
\caption{Representative tuning and optimization results. The precise theorem assumptions appear in the main text and appendices.}
\label{tab:first-results}
\begin{tabularx}{\linewidth}{@{}>{\raggedright\arraybackslash}p{.26\linewidth}X@{}}\toprule
\textbf{Tuning problem} & \textbf{What the result says and why it is informative}\\\midrule
One aggregated cutting plane & Zero-loss empirical tuning is NP-complete
for a shared vector of multipliers, even with small integer data. Integer
rounding creates interacting tests.\\
Scalar Lasso regularization & Zero ordinary squared validation loss is
NP-complete to attain with one positive regularization parameter. Dual
conditions exclude spurious zero-loss configurations.\\
Power-mean linkage & It is NP-hard to choose an exponent giving a
specified root split on a rational metric with distances in $[1,2]$.
Forced merges and clause tests enforce the target split.\\
Gradient descent & One constant step size can encode $\Omega(H\log k)$
independent choices in a piecewise-quadratic model. A one-weight ReLU
trained on four examples already gives $\Omega(H/\log H)$ statistical
capacity. Computational hardness uses a separate, possibly
discontinuous objective family.\\
Online matching and structured optimization & A scalar history-dependent
score can produce many decisions. Individual-choice tests are easy, while
valid clause tests make empirical tuning NP-hard. Restricted
interaction width permits dynamic programming.\\
Perturbed configuration & Relative noise yields an expected-polynomial
algorithm for one-variable binary cut closure. Gaussian noise gives an
expected Lasso path-length bound under its specified assumptions.\\\bottomrule
\end{tabularx}
\end{table}

The results use the same elementary proof structure. A parameter encodes
choices, and legal training inputs measure combinations of them. The
application-specific work is to realize the tests with the original
algorithm, rule out unintended low-loss configurations, and quantify
any precision required. The formal framework and supporting code
lemmas are in the appendices. We present the aggregate-cut, linkage,
and Lasso results first, followed by the effects of perturbations,
gradient descent, and complementary applications.

\section{Computational hardness of cutting-plane configuration}\label{sec:complexityextension}
\paragraph{One aggregated CG cut with shared row multipliers.}
We use the standard rank-one Chv\'atal--Gomory (CG) family studied in
\cite{prior:balcan2021cuts}. Given an integer program $Az\le b$ with
$m$ tunable rows, one shared vector $u\in[0,1]^m$ produces one inequality
$\lfloor u^\top A\rfloor z\le\lfloor u^\top b\rfloor$.
The vector $u$ is shared across all training IPs. Each hard IP has
one structurally binary variable and one dummy integer variable fixed to zero,
at most three meaningful tunable rows, and coefficients bounded by eight.
The restricted LP-based branch-and-bound procedure solves the root LP,
adds exactly one aggregated cut, and then continues ordinary LP-based
branching if necessary. Presolve, variable fixing outside the specified
domain, and additional cuts are disabled. The structural binary bounds are
not additional rows with freely tunable multipliers.
\begin{theorem}[Hardness of tuning one aggregated CG cut]
\label{thm:singlecg:main}
For the restricted root-only, no-presolve LP-based B\&B model, deciding
whether a shared CG multiplier vector gives all training IPs one-node
trees (equivalently closes every original LP gap) is NP-complete.
Minimizing average excess tree size or average normalized remaining gap
admits no finite multiplicative approximation and is hard to approximate
within a fixed positive additive constant unless $\mathsf P=\mathsf{NP}$.
There is a reduction from $3$-SAT on $T$ variables and $M$ clauses to
$N=5T+M$ training IPs with $m=2T$ shared multipliers. On the restricted
gadget class, $m\le\Pdim\le O(m\log m)$.
\end{theorem}
\begin{proof}[Proof idea, with full proof in Appendix~\ref{app:cg-single-aggregate}]
Two unary CG gadgets restrict a multiplier coordinate to one of two
intervals $J_0=[1/8,1/7]$ and $J_1=(1/4,2/7]$, representing false and
true. A two-row cut closes its root gap exactly when \emph{exactly one} of the two complementary literals of a
variable is true, and a three-row cut closes it exactly when the
represented clause has a true literal. Thus all gaps close exactly for
satisfiable formulas. For an arbitrary multiplier vector, the number
of failed gadgets controls the number of unsatisfied clauses after
rounding its coordinates to a Boolean assignment. Bounded-occurrence
gap-3SAT then gives a constant additive gap. A failing gadget closes at
most $5/12$ of its original root gap, and every failing B\&B tree has
exactly three nodes. The final statistical upper bound follows from a
finite hyperplane-arrangement argument.
\end{proof}
The multiplier dimension $m$ grows with the SAT instance. The solver
model evaluates one aggregated root cut, without presolve or further cuts.
Corollary~
ef{cor:cg-robust} extends this restricted-model hardness to
designated independent relative coefficient perturbations with radius
$1/40$. The robust construction uses an additional unary gate and
$7T+M$ training programs.
\begin{remark}[Fixed multiplier dimension with bounded coefficients]
With a fixed number $m$ of multiplier coordinates and coefficients of
bounded integer magnitude, the changes in the aggregated CG cut lie on
polynomially many rational hyperplanes in $[0,1]^m$. Their arrangement
has polynomially many cells when $m$ is fixed. For the present
root-only gadget class, evaluating one rational representative of each
cell and every relevant boundary face gives polynomial-time exact ERM.
Thus the hardness above requires a growing multiplier dimension.
The analogous question with a fixed number of coordinates but
binary-encoded coefficients of unbounded magnitude has a different precision cost.
\end{remark}
 The underlying hard IPs themselves
are trivial. The difficulty is the shared configuration. An auxiliary
positive result proves that with upper-bound-type gadgets only, zero
empirical loss reduces to linear feasibility, although exact ERM remains
NP-hard (Proposition~\ref{prop:singlecg:upperonly}).

\section{Power-mean linkage and root-split hardness}\label{sec:linkage-hardness}
For two nonempty disjoint clusters $U,V$, power-mean linkage compares
\[
D_\alpha(U,V)=\left(\frac{1}{|U||V|}
\sum_{u\in U,v\in V}\delta(u,v)^\alpha\right)^{1/\alpha},
\qquad \alpha\ne0.
\]
At $\alpha=0$ we use the geometric mean, and the limits at
$\alpha=\pm\infty$ give single and complete linkage.
For a target root split $L\mid R$ and root children $C,S\setminus C$,
the unnormalized root-Hamming loss is
\[
\ell_{\mathrm{root}}(C;L,R)=
\min\{|C\triangle L|,|C\triangle R|\}.
\]
It counts misclassified leaves after the better labeling of the two
root children and vanishes exactly when their unordered bipartition
matches $L\mid R$. A primitive distance is an input distance between
an unordered pair of individual points, before any merges occur.

The reduction uses two evolving clusters with identical neutral background
blocks. The common contributions to their linkage scores cancel, allowing
small differences in primitive distances to encode assignments as merge
choices. Clause probes then test these assignments. A failed clause forces
an irreversible merge across the prescribed root split. Guards confine
successful parameters to $[1,2]$, while a cleanup cluster completes the
successful hierarchy. Appendix~\ref{app:linkage-complete} gives the full
rational metric, all merge-order inequalities, and tie cases.
\begin{theorem}[Power-mean linkage root-split hardness]\label{thm:linkage:main}
For any fixed deterministic tie-breaking rule, deciding whether there exists
$\alpha\in\mathbb R\cup\{\pm\infty\}$ whose unmodified power-mean
linkage hierarchy has a prescribed root bipartition is NP-hard, even on
a rational metric with $O((n+m)^2)$ points, distances in $[1,2]$ and
$O(n+\log(n+m))$-bit entries.
\end{theorem}
\begin{corollary}[Linkage ERM values]\label{cor:linkage:erm}
For one metric instance, deciding whether the optimal root-Hamming
loss is zero is NP-hard, and its optimal value admits no finite-factor
polynomial-time approximation unless $\mathsf P=\mathsf{NP}$.
With a separately supplied complete similarity graph of weights $1$
and $2$, the minimum Dasgupta cost over linkage exponents is NP-hard
to compute and has no FPTAS for the optimal value unless $\mathsf P=\mathsf{NP}$.
\end{corollary}
The corollary concerns optimal loss values. The reduction uses exponentially
small distance offsets, and exact polynomial-time evaluation for arbitrary
real exponents remains an open question.

Randomly perturbed distances exhibit a different behavior. The following
result controls the number of changes in the \emph{entire} linkage hierarchy
as the exponent ranges from single to complete linkage. The proof uses
one-edge sensitivity and a bound on the speed of merge thresholds.

\begin{theorem}[Smoothed linkage path complexity]\label{thm:linkage:smoothed}
Let $S$ have $n\ge 2$ points. Draw its primitive unordered-pair
distances independently from $[1,2]$, with each density bounded by
$\kappa$. Fix deterministic tie breaking, and let $K(S;\Theta)$ count
finite exponents $\alpha\in\Theta$ at which the complete power-mean
linkage hierarchy changes. For every finite union of intervals
$\Theta\subseteq\mathbb R$,
\[
\mathbb E K(S;\Theta)
\le 2\kappa\binom{n}{2}(n-1)n^2
\int_\Theta\min\!\left\{\frac{(\log 2)^2}{8},
                         \frac{2\log n}{\alpha^2}\right\}\,d\alpha,
\]
where the integrand at $\alpha=0$ is defined by continuity. Consequently,
\[
\mathbb E K(S;\mathbb R)=O\!\left(\kappa n^5\sqrt{\log n}\right).
\]
\end{theorem}
\begin{corollary}[Smoothed linkage ERM]\label{cor:linkage:smoothed-erm}
For $m$ independently smoothed training instances of at most $n$ points
each, any polynomial-time evaluable loss of the complete hierarchy can
be minimized over $\alpha\in\mathbb R\cup\{\pm\infty\}$ in expected
polynomial time in $m,n,\kappa$, in an exact-real model equipped with an
all-real exponential-polynomial root-isolation oracle.
\end{corollary}
The transition bound is independent of the evaluation loss. The oracle
assumption in the corollary is needed to enumerate the parameter cells
and their boundaries. Polynomial expected Turing bit complexity remains
open. Both proofs appear in Appendix~\ref{sec:linkage:smoothed-proof}.

\section{Scalar Lasso validation tuning and path complexity}\label{sec:regression-extension}
\paragraph{Regularization-path tuning.}
For Lasso, the parameter does not directly pick a Boolean word. It
changes which regression coefficients are zero, positive, or negative.
A path with exponentially many such patterns is an encoding device,
but this alone cannot prove hard validation tuning. The key additional
step is to arrange a validation error that tests the encoded signs,
while the optimality conditions prevent a parameter outside the
intended path region from obtaining an unintended perfect score.
For training data $X\in\mathbb Q^{n\times p}$, $y\in\mathbb Q^n$, the Lasso solution path is
\begin{equation}\label{eq:lassoobjective}
\widehat w(\lambda)\in\arg\min_w\left\{\tfrac12\|Xw-y\|_2^2+\lambda\|w\|_1\right\},\qquad \lambda>0.
\end{equation}
Given a validation pair $(r,t)$, ordinary tuning minimizes the
squared prediction error $(r^\top\widehat w(\lambda)-t)^2$ over a
compact interval of positive parameters.

\begin{lemma}[KKT sign readout]\label{lem:regression-readout}
Suppose $X$ has full column rank, $G=X^\top X$, and $c=X^\top y$. On a full-support Lasso path segment with coefficient sign vector $s\in\{\pm1\}^p$,
\begin{equation}
\widehat w(\lambda)=G^{-1}(c-\lambda s).
\end{equation}
For any $u\in\mathbb Q^p$, choose validation vector $r=Gu$ and target $t=u^\top c$. Then
\begin{equation}\label{eq:lasso-readout}
(r^\top\widehat w(\lambda)-t)^2=\lambda^2(u^\top s)^2.
\end{equation}
\end{lemma}
\begin{proof}
The full-support first-order optimality equation is $G\widehat w=c-\lambda s$. Since $G$ is invertible, the claimed solution and readout follow by substitution.
\end{proof}

\begin{lemma}[Antipodal sign saturation of a maximal Lasso path]\label{lem:lasso-antipodal}
Assume the nondegenerate full-column-rank Lasso path attains the maximal
$(3^p+1)/2$ distinct sign patterns in the Mairal--Yu construction
\cite{MairalYu2012}. It then contains exactly one representative of every
antipodal pair $\{s,-s\}$ of nonzero vectors in $\{-1,0,1\}^p$ on a
nonempty open regularization interval. In particular, each full-support
antipodal pair $\{s,-s\}$ is represented.
\end{lemma}
\begin{proof}
Mairal and Yu prove that no nonzero sign state and its negative can both
occur on a full-rank Lasso path, and that a sign state cannot reoccur in
disjoint positive-length intervals. There are exactly $(3^p-1)/2$
nonzero antipodal pairs, plus the zero pattern. A path attaining
$(3^p+1)/2$ distinct sign states must contain zero and one state from
each pair. This count proves the assertion without an extra sign-schedule
induction. The result is an assertion about the \emph{real-arithmetic}
path, not a polynomial-bit construction of its rational coefficients.
\end{proof}

\begin{theorem}[Polynomial-bit Mairal--Yu path]\label{thm:lasso-polybit}
For $p\ge1$, let $B=100p^3$, $a_j=B^{-(j-1)}$, $y=\mathbf 1_p$, and
$X_{ij}=0$ for $i>j$, $X_{jj}=a_j$, and $X_{ij}=2a_j$ for $i<j$.
The $j$-dimensional truncated instance has no positive Lasso kink below
$a_j/[2(2j-1)^2]$, and the full $p$-dimensional path has exactly
$(3^p+1)/2$ sign-pattern segments. Its input has polynomial bit length,
and $\lambda_0=a_p/[4(2p-1)^2]$ is a positive polynomial-bit cutoff
below every kink.
\end{theorem}
\begin{proof}
Factor $X_j=U_jD_j$ where $D_j=\operatorname{diag}(a_1,\dots,a_j)$,
$U_j$ has diagonal $1$ and upper entries $2$, and
$(U_j^{-1})_{ik}=2(-1)^{k-i}$ for $i<k$.
Therefore $\|U_j^{-1}\|_\infty,\|U_j^{-\top}\|_\infty\le2j-1$.
The least-squares vector has coordinates
$w_i^{(0)}=(-1)^{j-i}/a_i$ and full support.
On this sign cell, the KKT candidate is
$w(\lambda)=w^{(0)}-\lambda G_j^{-1}s$, where
$G_j^{-1}=D_j^{-1}U_j^{-1}U_j^{-\top}D_j^{-1}$, and hence
$|[G_j^{-1}s]_i|\le(2j-1)^2/(a_ia_j)$.
For $0<\lambda<a_j/[2(2j-1)^2]$ all coordinates retain their
original sign. Strict convexity and the KKT equations show this is
the actual path, so there is no kink in that range.
Mairal--Yu's adversarial extension applies whenever
$a_{j+1}<\lambda_{\min,j}/(2\|y_j\|_2^2+1)$.
Because $\|y_j\|_2^2=j$, our choice obeys this inequality:
$B>2(2j-1)^2(2j+1)$ for all $j<p$.
Thus their segment recurrence $K_{j+1}=3K_j-1$, $K_1=2$, applies at
all stages, proving $K_p=(3^p+1)/2$.
Each nonzero matrix entry has $O(p\log p)$ bits, giving total
$O(p^3\log p)$ bits. The cutoff has $O(p\log p)$ bits.
\end{proof}

\subsection{Standard scalar Lasso validation loss with an adjacent-pair reduction}
The reduction uses standard squared validation loss without any support restriction or auxiliary sparsity penalty.
A \emph{signed partition} instance consists of integers
$d_1,\ldots,d_r,T$ and asks whether some
$\sigma\in\{-1,+1\}^r$ satisfies
$\sum_{j=1}^r d_j\sigma_j=T$.  This problem is NP-complete
(the target-zero version is the standard \textsc{Partition} problem).

\begin{theorem}[Ordinary scalar Lasso squared-validation zero-loss ERM]
\label{thm:lasso-ordinary-validation}
There is a polynomial-time many-one reduction from \textsc{Signed Partition}
to the following decision problem.  Given a full-column-rank rational
training pair $(X,y)$, a rational validation pair
$(X_{\rm val},y_{\rm val})$, and rational endpoints
$0<\lambda_{\rm lo}<\lambda_{\rm hi}$, decide whether
\[
\exists\lambda\in[\lambda_{\rm lo},\lambda_{\rm hi}]:\quad
 \|X_{\rm val}\widehat w(\lambda)-y_{\rm val}\|_2^2=0,
\qquad
\widehat w(\lambda)=\arg\min_w
\{\tfrac12\|Xw-y\|_2^2+\lambda\|w\|_1\}.
\]
The decision problem is NP-complete.  The reduction uses $p=2r+3$
training variables, $p$ training observations, and $r+3$ ordinary
squared-validation observations. The training design depends only on $r$,
and all input quantities, including both endpoints, have polynomial binary
length. No sparsity restriction or additional validation penalty is imposed.
The result establishes exact-decision hardness.
\end{theorem}
\begin{proof}[Proof outline, with details in Appendix~\ref{app:lasso-adjacent-proof}]
Let $G=X^\top X$, $c=X^\top y$, and
$g(\lambda)=X^\top(y-X\widehat w(\lambda))/\lambda$.
For every rational $u$, the validation row $Gu$ with response $u^\top c$
has residual $-\lambda u^\top g(\lambda)$.  Apply the polynomial-bit
Mairal--Yu construction of Theorem~\ref{thm:lasso-polybit} with
coordinate indices $0,\ldots,2r+2$ and top index $t=2r+2$.  Use the
$r+3$ dual readouts
\[
 g_0-g_t,\quad
 g_{2j-1}+g_{2j}\ (j\in[r]),\quad
 g_{t-1}+g_t,\quad
 \sum_{j=1}^r d_jg_{2j-1}-Tg_t.
\]
In the Mairal--Yu recursion, the newest dual coordinate is strictly
positive. Moreover every nonzero old-coordinate prefix is a signed,
rescaled copy of the preceding-level dual path. These identities imply
that if $g_0\neq0$ and $g_{k-1}+g_k=0$, then
$(g_{k-1},g_k)$ consists of opposite signs of magnitude one.
The top-pair and anchor equations force $g_t=g_0=1$. The remaining pair
equations force $\sigma_j=g_{2j-1}\in\{-1,1\}$.
The final equation becomes $\sum_jd_j\sigma_j=T$, proving soundness
at interior parameters and at path breakpoints.
Conversely, maximal-path antipodal saturation and positivity of the
newest coordinate imply that the full-support pattern
$(1,\sigma_1,-\sigma_1,\ldots,\sigma_r,-\sigma_r,-1,1)$
occurs for every signed assignment $\sigma$. A satisfying assignment
makes all residuals zero. The lower endpoint is chosen below every
positive kink, and rational KKT certificates give NP membership.
\end{proof}
\begin{corollary}[Finite-factor inapproximability of Lasso validation ERM]
\label{cor:lasso-zero-approx}
Unless $\mathsf P=\mathsf{NP}$, no polynomial-time algorithm returning a
rational regularization parameter can achieve a finite multiplicative
approximation factor for the minimum ordinary squared validation loss
on the compact positive interval in
Theorem~\ref{thm:lasso-ordinary-validation}.
\end{corollary}
\begin{proof}
The loss is nonnegative, so a finite multiplicative guarantee must attain
zero whenever the optimal loss is zero. A polynomial-bit rational candidate
can be checked by solving the corresponding rational, full-rank Lasso
KKT system and evaluating the validation loss. Such an approximation
algorithm would therefore decide the NP-complete zero-loss problem.
\end{proof}

\begin{remark}[Lasso hardness, path enumeration, and smoothed precision]
The exact zero-validation result is weak NP-hardness.
The polynomial-bit Mairal--Yu design contains exponentially separated
entries, and no hardness for polynomially bounded dynamic range is
proved here. If the regularization-path breakpoints are supplied,
empirical squared-validation loss can be optimized in time polynomial
in their number (Proposition~\ref{prop:regression-scan}). Under the
Gaussian model of Theorem~\ref{thm:lasso:kac-rice}, their expected number
is polynomial. Combining the latter two observations gives an
expected-polynomial \emph{real-arithmetic path enumeration approach}
when each segment can be generated in polynomial arithmetic work,
but not automatically expected-polynomial Turing bit complexity.
The output-sensitive algorithm and the Gaussian segment-count result control different aspects of the computation.
\end{remark}

\begin{remark}[Why a positive compact parameter interval is necessary]
\label{rem:lasso-positive-domain}
Every constructed validation residual equals $-\lambda u^\top g(\lambda)$,
and $\|g(\lambda)\|_\infty\le1$. Consequently the validation loss tends
to zero as $\lambda\downarrow0$ on both YES- and NO-instances. Over the
open domain $\lambda>0$, the statement
$\inf_{\lambda>0}V(\lambda)=0$ is therefore trivial.
Theorem~\ref{thm:lasso-ordinary-validation} concerns \emph{attainment of
zero} or exact minimization on a closed interval bounded away from zero.
It does not imply a fixed additive approximation gap or hardness of
conventional scalar Lasso AIC.
\end{remark}

\begin{proposition}[Output-sensitive exact validation ERM along regularization paths]\label{prop:regression-scan}
Consider $m$ Lasso training tasks with full-rank rational designs and squared validation losses, all tuned by the same scalar $\lambda$ over a compact interval. If each exact regularization path and its rational breakpoints are supplied, exact global ERM is computable in time polynomial in the aggregate number $K$ of segments, the task dimensions, and input bit lengths. The same claim holds for Elastic Net while tuning $\lambda_1$ with a \emph{fixed} $\lambda_2>0$, assuming the stated solution is unique and the exact path is supplied.
\end{proposition}
\begin{proof}
On each fixed Lasso support/sign interval $J,s$, the active coefficients satisfy $\widehat w_J(\lambda)=(X_J^\top X_J)^{-1}(X_J^\top y-\lambda s)$, and the inactive coordinates equal zero. For fixed-$(\lambda_2>0)$ Elastic Net, replace $X_J^\top X_J$ by $X_J^\top X_J+\lambda_2 I$. Thus every squared validation loss is quadratic in the scalar $\lambda_1$ or $\lambda$, on each segment. Merge all tasks' breakpoints, minimize the sum quadratic on each common interval and at its endpoints, and keep the best candidate. Standard exact rational linear algebra and quadratic minimization suffice when all input coefficients and breakpoints are rational and have the stated finite encodings.
\end{proof}

Mairal and Yu~\cite{MairalYu2012} establish exponentially many Lasso path pieces in the worst case, so Proposition~\ref{prop:regression-scan} is output-sensitive, not a worst-case polynomial-time guarantee. Li and Singer~\cite{LiSinger2018} prove polynomial path-complexity bounds under an explicit smoothed-input model. Under the hypotheses of their theorem, the output-sensitive method consequently admits corresponding \emph{high-probability polynomial} complexity, provided a suitable path enumerator obeys the polynomial per-piece bit-cost and nondegeneracy conditions. A high-probability bound alone is not a proof of polynomial \emph{expected} running time, because rare instances can have exponentially long paths. Theorem~\ref{thm:lasso:kac-rice} bounds the expected segment count under
Gaussian perturbations. With polynomial arithmetic work per segment,
this gives expected-polynomial real-arithmetic path enumeration.
Expected-polynomial Turing bit complexity requires further analysis.

\paragraph{Other regression guarantees and unresolved cases.}
Other statistical bounds for regression tuning depend on the margins of
the coefficient and validation constructions. Ridge paths are rational functions of $\lambda$, and
Proposition~\ref{prop:ridge-P} gives polynomial-time exact
scalar ridge validation ERM on rational compact intervals.
Theorem~\ref{thm:lasso-ordinary-validation} gives zero-loss hardness for
ordinary scalar Lasso validation on a compact positive interval. Hardness
with a fixed number of validation examples, a constant additive gap, or
ordinary scalar Elastic Net remains open. The Lasso hardness proof is detailed in Appendix~\ref{app:lasso-adjacent-proof}.

\section{Precision, perturbations, and tractable configuration}\label{sec:three-axis-smoothing}
A worst-case construction may rely on a parameter interval so narrow
that rounding errors almost surely erase it. Another construction may
remain valid even if every numerical input changes a little. These
situations call for different conclusions. The second supports a
smoothed-hardness reduction when the perturbations are both
independent and large enough in the input's normal scale, while the first
may lead to an efficient average-over-perturbations algorithm. This
perspective follows the classical smoothed-analysis approach of
\cite{SpielmanTeng2004,BeierVocking2006} and is also related to smoothing in online learning
\cite{HaghtalabRoughgardenShetty2020}, equilibrium computation
\cite{ChenDengTeng2009,BoodaghiansEtAl2020}, and local optimization
\cite{ChenEtAl2020}. The perturbation distributions in those works
are not identical to ours.

The code-to-loss compiler has two outputs. Coordinate probes certify statistical capacity, while calibrated joint constraints give ERM hardness. A third property concerns survival under perturbations of the \emph{primitive input}. It is possible for a class to have high worst-case pseudodimension and yet have few parameter regions under input smoothing. It is also possible for its hard Boolean decoding to survive every perturbation in an inverse-polynomial box.

\paragraph{Perturbation assumptions.}
A candidate smoothed result must identify the independently perturbed entries, the treatment of zeros and combinatorial data, the arithmetic model, and whether it bounds output changes or solves exact ERM. In particular, a smoothed bound on the number of distinct outputs is \emph{not} automatically a polynomial Turing-time algorithm for discovering all parameter regions.

\begin{definition}[Bounded relative perturbations]\label{def:relative-smoothed}
For a finite rational description of an instance, replace every designated nonzero numerical entry $v$ by $v(1+\eta_v)$, with independently drawn $\eta_v$ in $[-\rho,\rho]$, independently across training instances unless stated otherwise. Zero entries and explicitly combinatorial structure are unchanged. A bounded-density version assumes the density of each $\eta_v$ is at most $\kappa$ on $[-1/2,1/2]$. We refer to this as \emph{sparsity-preserving multiplicative smoothing}. It differs from additive Gaussian perturbations, from independently drawn pairwise distances, and from perturbing Euclidean point coordinates.
\end{definition}

\begin{theorem}[Box-robust reduction implies hardness under smoothing]\label{thm:smoothed-robust-transfer}
Suppose a polynomial-time reduction maps each instance $\phi$ of an NP-hard promise decision problem to a polynomial-bit DDAD sample $S_\phi$, and the relevant zero/nonzero-optimum or constant-gap distinction is unchanged under \emph{every} perturbation $|\eta_v|\le\rho(n)$ of the designated entries. Assume $\rho(n)\ge1/\poly(n)$, each resulting ERM instance can be evaluated/verified in the stated randomized computation model, and the model admits polynomial-bit simulation of its perturbation experiment. Then an algorithm that, for every base sample and all density parameters $\kappa=\poly(n)$, solves the perturbed ERM problem with probability at least $2/3$ in time $\poly(n,\kappa)$ would imply $\mathsf{NP}\subseteq\mathsf{BPP}$. If the algorithm is always correct whenever it terminates and has polynomial expected running time, the corresponding zero-error conclusion is $\mathsf{NP}\subseteq\mathsf{ZPP}$.
\end{theorem}
\begin{proof}
Choose independent noise uniformly on $[-\rho/2,\rho/2]$. Its density is $1/\rho=\poly(n)$ and its support lies inside the robustness box. By assumption \emph{every} perturbation preserves the promise answer. Generate a perturbed sample, invoke the smoothed ERM algorithm and decode its answer. Repetition amplifies bounded success probability. For zero-error algorithms, restart after a timeout and use correctness on termination, provided exact random sampling and the stated polynomial-bit simulation are available. A continuous-real oracle without an efficient discrete simulation does not alone imply a Turing-model inclusion.
\end{proof}

\begin{proposition}[Smoothed statistical capacity survives a common robust code]\label{prop:robust-fat-smooth}
Suppose one set of parameters $\{\theta_b:b\in\{0,1\}^d\}$ witnesses $\fat_\gamma\ge d$ on a collection of evaluation instances, and for every perturbation in a box their losses change by at most $\varepsilon$ at \emph{all} these parameter/input pairs. Then for each perturbed sample the same parameters witness $\fat_{\gamma-\varepsilon}\ge d$ whenever $\varepsilon<\gamma$. This is a pointwise robustness statement, not a claim about random-data pseudodimension for all possible input distributions.
\end{proposition}
\begin{proof}Use the same thresholds and apply the two-sided error bound to all signs.\end{proof}

\paragraph{Perturbation convention.}
Definition~\ref{def:relative-smoothed} is the default for sparse
algorithmic inputs. Fixed-unit additive errors in scalar batch CG
certify worst-case robustness. The Gaussian Lasso result and the
independent-distance linkage theorem (Theorem~\ref{thm:linkage:smoothed})
use their own distributional hypotheses. In particular, independently
sampling pairwise distances from $[1,2]$ preserves the triangle inequality.

\paragraph{Robustness of the individual constructions.}
For the history-independent matching clause probes, weights enter the loss but not the encoded decisions. Positive perturbations preserve which probe is matched, so its failure cost stays bounded away from zero. The precise normalized margin depends on the chosen normalization and which entries are smoothed. For the single aggregated CG-cut construction, smoothing is harder. The row coefficients and right-hand sides in the rounded cut can move independently, so exact equal-coefficient sums no longer apply. The appendix gives a replacement certificate based on uniform multiplier windows. For the history-dependent B\&B controller, uniform slack at partial LP nodes and incumbent updates yields box robustness under independent relative perturbations of radius $1/[1000(T+1+K)]$ (Theorem~\ref{bbrobust:thm:main}). The theorem uses a fixed clause-reading node selector.

% \subsection{A generic pivotal-coordinate route to smoothed tractability}
\begin{remark}[Using the pivotal-coordinate certificate]\label{thm:smoothed-coarea-general}
The area-formula bound of Theorem~\ref{thm:cert-pivotal} also applies to
nonuniform count and speed envelopes. If at most $R_e(\theta)$ graphs
cross each vertical line and their derivatives are bounded by
$V_e(\theta)$, its right-hand side becomes
$\sum_e\kappa_e\int_I R_e(\theta)V_e(\theta)\,d\theta$.
This is a conditional theorem. Its chargeability and speed hypotheses
must be verified for the actual application.
\end{remark}

\begin{remark}[A smoothed semi-supervised-learning example]\label{rem:ssl-smoothed}
For graph-based semi-supervised labeling formulated as a minimum cut~\cite{prior:balcansharma2021},
independent bounded-density perturbations of the \emph{edge distances}
permit a whole-bandwidth expected-polynomial bound on the number of
optimal-cut transitions for exponential/Gaussian edge weights.
This is an output-region bound under an independent-edge model, not a
polynomial-time exact parameter optimizer and not a result for dependent
Euclidean distances.  The formulation and proof are deferred to
Appendix~\ref{sec:gaussian-smoothed}.
\end{remark}

\begin{remark}[Model-specific smoothing conclusions]\label{rem:smoothed-classification}
Pointwise robust worst-case hardness, expected transition counts, and an
expected-polynomial exact ERM algorithm are different conclusions.  In
particular, additive noise of fixed absolute size on exponentially large
integer coefficients does not constitute inverse-polynomial noise after
normalization.  The assumptions and certified status of each application
are collected in Appendix~\ref{app:smoothed-families}.
\end{remark}

\subsection{Scalar CG precision and expected Lasso path length}\label{sec:round3-new}
The scalar batch-CG policy applies one real parameter to a batch of
\emph{individually rounded} rank-one cuts. It is different from tuning the
multiplier vector of one aggregated cut.

\begin{theorem}[Precision trichotomy for one-variable batch CG]\label{thm:scabs:overview}
For the scalar, individually rounded batch-CG policy specified in
Appendix~\ref{scabs:app:scalar-cg-additive-smoothing}, the following hold.
\begin{enumerate}[label=(\roman*)]
\item Zero empirical remaining-root-gap loss is NP-hard to attain on
one-variable IPs with at most three rows and polynomial-bit, potentially
exponentially large coefficients. The reduction tolerates pointwise
additive errors of magnitude at most $1/10$ in the raw coefficients and
right-hand sides, and nonclosed clauses leave at least $4/5$ of their
original root gap.
\item For binary root-gap closure and the corresponding binary tree-size
loss, exact ERM has an output-sensitive pseudo-polynomial algorithm. Its
number of candidate endpoints is bounded by
$O(\sum_j\min\{\lceil A_j\rceil,\lceil1/s_j\rceil\})$, where
$s_j=1-B_j/A_j$ for a feasible row $-A_jx\le -B_j$.
\item Under independent feasible relative perturbations of those rows
with inverse-polynomial noise radius and the finite-precision assumptions
of Theorem~\ref{thm:cg-relative-tract}, the binary closure/tree-size
ERM algorithm has expected polynomial running time.
\end{enumerate}
The second and third claims do not extend by this argument to exact
\emph{numerical} remaining-gap optimization.
\end{theorem}
\begin{proof}
Part (i) is proved in Appendix~\ref{scabs:app:scalar-cg-additive-smoothing}.
The exact cut-closing interval formula is
Lemma~\ref{lem:cg-one-row-interval}, and the algorithm and noise analysis
for (ii)--(iii) appear in Theorem~\ref{thm:cg-relative-tract}.
\end{proof}

\begin{remark}[Fixed-unit robustness is not smoothed hardness]
The construction in part (i) has coefficients of magnitude
$\Theta(1000^T)$. Dividing all coefficients by their largest magnitude
turns the permitted absolute error $1/10$ into an exponentially small
relative error. Consequently the reduction does not contradict a running
time polynomial in the inverse normalized smoothing scale. Open directions include several-variable scalar batch cuts,
fixed-dimensional aggregated cuts with binary-encoded coefficients of
unbounded magnitude, and the exact numerical remaining-gap objective.
For growing multiplier dimension, the box-robust aggregated-cut
construction is already established in Corollary~\ref{cor:cg-robust}.
\end{remark}

\subsubsection{Expected smoothed Lasso path length by coordinate entry counting}
\begin{theorem}[Smoothed Lasso path length]\label{thm:lasso:kac-rice}
Fix $n\ge d\ge2$, $0<\sigma\le1$, $\|y\|_2=1$, $X=X^h+G$, $\|X^h\|_{\mathrm{op}}\le1$, and independent $G_{ij}\sim N(0,\sigma^2/n)$. Let $K(X,y)$ be the number of nondegenerate linear segments of the scalar Lasso regularization path. Then
\[
 \mathbb E K(X,y)\le C\,\frac{n d^{5/2}}{\sigma^2}\left(1+\log\frac{nd}{\sigma}\right)
\]
for an absolute constant $C$. The proof is given in Appendix~\ref{app:lasso:kac-rice}. It uses independent-column conditioning, a one-dimensional Gaussian zero-crossing identity, and two-column distance small-ball estimates. The theorem concerns the expected \emph{number} of path segments, not automatically the bit-complexity of computing every kink.
\end{theorem}
Unlike the high-probability smoothed path bound of Li and Singer~\cite{LiSinger2018}, this theorem directly controls the expectation of the number of path pieces under its stated Gaussian design model. Its $n d^{5/2}\sigma^{-2}$ dependence is an upper bound, not a proved optimal rate, and the factor $n$ is not known to be necessary. The response $y$ is fixed rather than perturbed. We do not directly compare numerical constants or rates across different noise normalizations. Nor does a path-count bound alone control the bit complexity of finding all breakpoints.

\begin{remark}[Online matching with history-dependent policies]\label{rem:matching-summary}
A single tunable mixture of two fixed, history-dependent scores can realize
$T$ independent decisions in online arrival-weighted matching.  On legal
matching instances, normalized regret yields $\fat_{1/4}=\Pdim=T$.
Unary-probe ERM is polynomial-time, whereas two-literal probes make exact
ERM NP-hard and three-literal probes make zero-loss ERM NP-complete.
When the probe interaction graph has treewidth $w$, dynamic programming
instead gives $2^{O(w)}\poly(T,m)$-time ERM without enumerating all
$2^T$ parameter intervals.  These are properties of this explicitly
history-dependent policy family, not claims about memoryless Ranking.
The complete policy, regret identities, reductions, and tractability proofs
are in Appendix~\ref{sec:matching}.
\end{remark}

\section{Gradient descent capacity, ERM hardness, and smoothing}\label{sec:gd-new}
The familiar constant-step gradient-descent update $x_{t+1}=x_t-\eta\nabla f(x_t)$ can realize the same code-to-probe compiler without an expressly programmed controller. The construction uses an expanding piecewise-quadratic map whose left and right branches both cover a common state interval. Varying one scalar step size changes the first launched state. Successive iterations reveal its symbolic itinerary. This realization is useful because the iteration-count loss itself reads the code through small convergence regions. The complete decoder, arithmetic data, proof of regularity, and hardness reduction are in Appendix~\ref{app:gd-new}.

\begin{theorem}[GD capacity and ERM arity, summary]\label{thm:gd-summary}
For polynomial-description piecewise-quadratic, potentially discontinuous
objectives with capped gradient-norm stopping time, the following worst-case
statements hold on compact rational step-size windows.\par\smallskip\noindent\textbf{(a)} constant-step tuning has $\fat_{1/4}=\Omega(H\log k)$ with $k$ branches and $O(H+k)$ affine boundaries in dimension two.\par\smallskip\noindent\textbf{(b)} an $H$-parameter schedule has $\fat_{1/4}=\Omega(H^2)$ already with two-dimensional objectives.\par\smallskip\noindent\textbf{(c)} tuning a $d$-coordinate initialization has $\fat_{1/4}=\Omega(dH)$.\par\smallskip\noindent\textbf{(d)} squared-error training of a single ReLU neuron with one trainable scalar and four examples has $\fat_{1/4}=\Omega(H/\log H)$. And (e) for two-dimensional GD, ERM over a scalar step size is NP-hard with two-literal convergence sinks, while the corresponding unary-sink class admits polynomial-time ERM. Three-literal sinks make zero-nonconvergence feasibility NP-complete and yield constant-additive inapproximability.
\end{theorem}

The four-example construction in (d) is a continuous ReLU squared-loss objective (with one fixed feature coefficient). Earlier neural-network capacity lower bounds also exploited numerical expressivity \cite{KoiranSontag1997}, while computational hardness of learning a neuron under label noise addresses a distinct problem \cite{DiakonikolasEtAl2022}. The clause-sink objectives of (e), by contrast, can be discontinuous and are not neural-network losses. The $1/4$-fat-shattering claims concern iteration counts divided by $H$ and use a strict gradient threshold, with witness step sizes avoiding piece boundaries, as in the general piecewise-polynomial setting of \cite{SharmaGD2026}. Theorem~\ref{thm:gd-summary} is also not a claim of computational hardness after independently perturbing each training objective's decoder. Agreement of $\Theta(H)$ symbols across separate perturbed decoders is exponentially precision-sensitive; Appendix~\ref{app:gd-new} makes the limitation explicit.

\begin{table}[t]\centering\small
\caption{Gradient-descent instantiations of the encoding/verification framework. All hardness is worst-case unless stated otherwise.}\label{tab:gd-summary}
\begin{tabularx}{\linewidth}{@{}p{.29\linewidth}X p{.25\linewidth}@{}}\toprule
Family & Capacity or ERM behavior & Assumptions\\\midrule
Single scalar step, $k$ branches & $\Omega(H\log k)$ fat dimension & $d=2$, piecewise quadratic\\
Full learning-rate schedule & $\Omega(H^2)$ fat dimension & $d=2$, nonconvex\\
Initialization vector & $\Omega(dH)$ fat dimension & $d+1$ coordinates\\
Four-example ReLU neuron & $\Omega(H/\log H)$ fat dimension & one trainable weight, continuous loss\\
Literal-sink GD probes & Unary ERM is polynomial. Binary ERM is NP-hard. Ternary zero-loss ERM is NP-complete & $d=2$, discontinuous pieces\\
Independent  smoothing & shared-decoder reduction does not transfer & open for these probes\\\bottomrule
\end{tabularx}
\end{table}

\section{Complementary applications}\label{sec:apps}
Two additional models illustrate how the form of the empirical loss
affects configuration complexity. In online matching, one parameter
mixes two fixed history-dependent scores, producing a sequence of
assignment choices. Legal regret probes can test individual choices
or clauses, and bounded interaction width allows dynamic programming
(Appendix~\ref{sec:matching}). For branch-and-bound, a fixed
history-dependent node selector reads clause information encoded in the
integer program and yields NP-hard tree-size tuning. This construction
uses a policy capable of inspecting prior search decisions
(Appendix~\ref{app:bb-erm-detail}). The arithmetic and regression
mechanisms behind the headline results use standard cutting-plane and
Lasso operations. The matching and node-selection examples clarify the
additional expressiveness supplied by search history.

\section{Related work}
\paragraph{Statistical algorithm design and empirical optimization.}
Gupta and Roughgarden~\cite{GuptaRoughgarden2017,GuptaRoughgarden2020}
placed application-specific algorithm selection in a learning-theoretic
framework. Balcan et al.~\cite{BalcanDeBlasioEtAl2021} developed general
sample-complexity guarantees for data-driven algorithm design. Balcan,
Dick, and Vitercik~\cite{BalcanDickVitercik2018} used \emph{dispersion}
to analyze optimization and online learning with piecewise-regular losses.
Related online-learning and meta-learning guarantees for discontinuous
piecewise-Lipschitz losses appear in
\cite{SharmaBalcanDick2020,BalcanKhodakSharmaTalwalkar2021}.
More recent open problems address online optimization over such
functions~\cite{BalcanPegdenSharma2026}.

Previous statistical results cover partitioning and
linkage~\cite{prior:balcan2017}, learning to
branch~\cite{prior:balcan2018,prior:balcan2022bc}, cutting-plane
selection~\cite{prior:balcan2021cuts,prior:khalifelodi2026}, and
parameterized decision-tree splitting
criteria~\cite{prior:balcansharma2024}.
Graph-based semi-supervised learning offers another example in which
tunable algorithms and architectures admit statistical
guarantees~\cite{prior:balcansharma2021,DuHuangSharma2025}.
The Pfaffian framework gives structural generalization bounds and
dispersion results for additional algorithm families, including a
restricted linkage family~\cite{BalcanNguyenSharma2025Pfaffian}.
Our linkage result instead bounds the expected total number of changes
in the full power-mean hierarchy over all real exponents.
Output-sensitive methods minimize empirical losses by processing
parameter cells actually encountered~\cite{BalcanSeilerSharma2024}.
The results here investigate when exact empirical optimization is
computationally difficult and when perturbations make that enumeration
manageable.

\paragraph{Hardness of empirical learning.}
Computational hardness of proper learning and exact fitting has a long
history, including Pitt and Valiant~\cite{PittValiant1988} and Blum and
Rivest~\cite{BlumRivest1988}. Our NP-hardness reductions do not imply
hardness of statistical generalization or of training neural networks
in those classical models. Their distinguishing feature is that a
\emph{shared algorithm configuration} is optimized over different
inputs to an existing algorithm and evaluated using its standard
performance measure. In particular, the Lasso result concerns held-out
squared loss after convex training, rather than computational hardness
of solving one Lasso optimization problem. The aggregated-CG result
concerns one standard cut and bounded input coefficients. For proof
complexity comparisons, branch-and-cut and cutting-plane proof systems
have been studied by~\cite{FlemingEtAl2021}. Our specialized
history-dependent B\&B controllers differ from standard
variable-selection heuristics.

\paragraph{Smoothed complexity and the role of precision.}
Smoothed analysis originates with Spielman and Teng
\cite{SpielmanTeng2004}. Beier and V\"ocking~\cite{BeierVocking2006}
showed a pseudo-polynomial versus smoothed-polynomial equivalence for
certain binary-optimization models. Our one-variable batch-cut results
are an application-specific analogue under different assumptions.
Robust reductions for smoothed complexity have precedents in equilibrium
computation~\cite{ChenDengTeng2009,BoodaghiansEtAl2020}. Different
perturbation models arise in local
optimization~\cite{ChenEtAl2020}. %An inverse-polynomial perturbation
% must be measured relative to normalized data, rather than as a fixed
% absolute error on exponentially large integers.

\paragraph{Regression, clustering, and hyperparameter tuning.}
Prior work established generalization and online-learning guarantees
for tuning Ridge, Lasso, and Elastic Net regularization across
instances~\cite{BalcanKhodakSharmaTalwalkar2022,prior:balcannguyensharma2023}.
Distribution-dependent bounds for regression tuning provide sharper
statistical guarantees under additional assumptions on the task
distributions~\cite{BalcanGoyalSharma2025}.
Mairal and Yu~\cite{MairalYu2012} constructed Lasso paths with
exponentially many segments, and Li and Singer~\cite{LiSinger2018}
studied path complexity under smoothed data. These results motivate
our analysis of the computational complexity of minimizing held-out
squared loss over the regularization parameter. Our proof uses dual
conditions to exclude unintended zero-loss states.

Power-mean linkage belongs to the agglomerative clustering families
studied in algorithm configuration~\cite{prior:balcan2017}. The
prescribed root-split and Dasgupta-cost questions considered here
concern computational optimization over linkage exponents. Dasgupta
introduced the relevant hierarchical clustering
cost~\cite{Dasgupta2016}.
Neural-network hyperparameter tuning has also been studied through
structured parameter-dependent dual
functions~\cite{BalcanNguyenSharma2025Neural}.
For gradient descent, extracting long decision sequences from a
numerical parameter is related to neural-network capacity
constructions~\cite{KoiranSontag1997}, although our iteration mechanism
and four-example squared-ReLU model are different.
Learning-augmented online algorithms
\cite{AnandGeKumarPanigrahi2021,BamasMaggioriSvensson2020,GrigorescuEtAl2022}
study performance with predictions. Offline-to-online hyperparameter
transfer for stochastic bandits~\cite{SharmaSuggala2025} similarly
uses information from previous tasks to improve future decisions.
These settings differ from exact empirical optimization of the
configuration families analyzed here.
\section{Research directions}\label{sec:research-directions}
Our computational hardness results raise questions about the algorithmic features and numerical precision needed for difficult configuration problems.

\paragraph{Can familiar solver rules produce the same hardness?}
The aggregated-cut result uses one standard rank-one cut with a shared
multiplier vector. Its restricted solver model omits presolve and later
cut rounds. Does hardness persist when these operations are included?
For branching, the current clause-reading node selector keeps search
history and inspects coefficients. Can conventional branching scores,
using only local LP information, yield a comparable lower bound on
empirical solve time or tree size?

\paragraph{Which hard examples survive independent noise?}
The scalar batch-cut precision theorem gives both worst-case hardness
and an expected-polynomial algorithm under relative perturbations for
binary root-gap closure. Linkage uses exponentially fine clock offsets,
while the current multi-instance GD construction requires consistent
symbolic trajectories across objectives. For linkage, the expected number of hierarchy changes is already
polynomial under independently sampled bounded-density primitive distances
(Theorem~\ref{thm:linkage:smoothed}). Its polynomial expected Turing
bit complexity remains open, as does robustness under perturbations of
arbitrary base metrics. The aggregated-CG construction remains hard
under the designated independent relative perturbations of
Corollary~\ref{cor:cg-robust}. It remains to understand more realistic
solver models and whether GD hardness survives separate perturbations
of its training objectives.

\paragraph{Can empirical tuning avoid exhaustive path enumeration?}
The Lasso result constructs exponentially many support/sign patterns.
The Gaussian theorem bounds their expected number, but efficient exact
parameter optimization also depends on locating path breakpoints with
controlled arithmetic precision. Can an algorithm avoid enumerating
most of the path, or obtain expected-polynomial bit complexity? For a
fixed number of aggregated-CG multipliers and bounded coefficients,
arrangement enumeration is polynomial. The corresponding problem with
binary-encoded coefficients of unbounded magnitude remains open.

\paragraph{How simple can an ERM-hard configuration family be?}
The Lasso theorem tunes one regularization parameter and the GD
statistical construction uses one learning rate. GD ERM hardness here
uses a two-dimensional objective that may be discontinuous, while a
continuous, four-example ReLU loss yields a capacity lower bound.
Can one obtain hard ERM for such a small continuous neural training
problem, for ordinary scalar Elastic Net, or for well-conditioned
Lasso designs with polynomially bounded dynamic range? A stronger
hardness result in any of these settings would separate the intrinsic
optimization difficulty from the precision of the current witnesses.

\appendix

\section{Decision-code definitions and the probe compiler}\label{sec:compiler}
Let $\X_T$ be an instance family, $\Theta_T$ the parameter domain, and $\F_T=\{f_\theta:\theta\in\Theta_T\}\subseteq[0,1]^{\X_T}$. We use the usual definition of $\fat_\gamma(\F_T)$. Instances $x_1,\ldots,x_d$ are $\gamma$-fat-shattered if there are thresholds $r_i$ such that every $b\in\bits^d$ is realized by some $\theta$ with $b_i=1$ implying $f_\theta(x_i)\ge r_i+\gamma$ and $b_i=0$ implying $f_\theta(x_i)\le r_i-\gamma$.

\begin{definition}[Efficient code and probes]\label{def:probe}
An algorithmic family has an efficient $T$-bit code if there is a deterministic map $B:\Theta_T\to\bits^T$ and a polynomial-time procedure $\mathsf{Encode}(b)$ outputting a rational parameter $\theta_b$ with $O(\poly(T))$ bits and $B(\theta_b)=b$, such that evaluating $f_{\theta_b}(x)$ is polynomial-time for valid $x$.
For a Boolean predicate $P:\bits^r\to\bits$ and an ordered tuple of distinct coordinates $I=(i_1,\ldots,i_r)$, a $\Delta$-\emph{predicate probe} is an efficiently constructible valid instance $x_{P,I}$ satisfying
\[
f_\theta(x_{P,I})=c_{P,I}+\Delta P(B_{i_1}(\theta),\ldots,B_{i_r}(\theta))
\]
for every $\theta\in\Theta_T$, with $\Delta>0$ and a known baseline $c_{P,I}$ independent of $\theta$.
\end{definition}

\begin{insight}
This theorem is a transfer principle, not a black-box claim that every
algorithm implements SAT. If its \emph{actual} instances can charge for
a falsified two- or three-variable clause, optimizing the algorithm's
single shared parameter is as hard as optimizing those clauses. With
only one-variable tests, the very same code can be optimized one bit
at a time.
\end{insight}

\begin{theorem}[Constraint-probe compiler]\label{thm:compiler}
Suppose a bounded-loss DDAD class has an efficient $T$-bit code and permits probes with $c_{P,I}=0$ and a common $\Delta>0$ for the predicate that a supplied Boolean clause is falsified. Then the empirical loss of a collection of clause probes is exactly $\Delta$ times its fraction of unsatisfied clauses. Consequently:
\begin{enumerate}[label=(\roman*)]
\item if probes for all clauses of width at most three are permitted, deciding whether $\min_\theta\widehat L(\theta)=0$ is NP-complete under the rational-witness formulation, and there is no finite multiplicative approximation unless $\mathsf P=\mathsf{NP}$.
\item if width-two probes are permitted, exact ERM is NP-hard.
\item for width-one probes, exact ERM is polynomial-time (assuming no other input types are allowed).
\item if the available probes also read each individual bit with a uniform loss gap $\Delta$, then $\fat_{\Delta/2}(\F_T)\ge T$.
\end{enumerate}
\end{theorem}
\begin{proof}
Given clauses $C_1,\ldots,C_m$, construct $x_{C_j}$. Since $B$ is onto, minimization over $\theta$ is identical to minimization over $b\in\bits^T$ of $\Delta m^{-1}\sum_j\one\{C_j(b)\text{ is false}\}$. In the width-three case, zero loss is equivalent to satisfiability. A satisfiable formula yields a rational parameter $\mathsf{Encode}(b)$ of polynomial length. Evaluating it verifies the zero loss, proving NP membership for this constructed decision language. Finite-factor approximation must find zero loss on zero-optimum instances.
For width two, give each edge $\{u,v\}$ of a \textsc{Max-Cut} graph the two clauses $(x_u\vee x_v)$ and $(\neg x_u\vee\neg x_v)$. Precisely one of the two clauses is unsatisfied if $b_u=b_v$ and neither is unsatisfied otherwise. Thus minimizing empirical loss maximizes cut size. Unary clauses separate over variables and can be optimized independently. Finally, an individual-bit probe gives losses $0$ and $\Delta$ across all codewords. Choosing threshold $\Delta/2$ gives margin $\Delta/2$.
\end{proof}

\begin{remark}[What the compiler does not show]
The predicate-probe condition is substantive. The instances must belong to the original application, and the loss must be its ordinary objective. Parameter codewords alone do not imply NP-hard ERM. Equally, NP-hardness for one evaluation family does not imply hardness on all distributions or for a better structured subset of training inputs.
\end{remark}

\subsection{Quantitative decoding with interference}
\begin{lemma}[Robust observation]\label{lem:robust}
Suppose a code $B$ realizes all $\bits^d$ and evaluation instances satisfy, at the witnessing parameters,
\[
f_{\theta_b}(x_i)=a_i+\Delta_i b_i+h_i(b_{-i})+e_i(b),
\quad \osc(h_i)\le\omega_i,\quad |e_i(b)|\le\varepsilon_i.
\]
If $\Delta_i>\omega_i+2\varepsilon_i$ for every $i$, then
\[
\fat_\gamma(\F)\ge d\quad\text{for }0<\gamma\le\frac12\min_i(\Delta_i-\omega_i-2\varepsilon_i).
\]
\end{lemma}
\begin{proof}
The largest loss with $b_i=0$ is at most $a_i+\max h_i+\varepsilon_i$. The smallest with $b_i=1$ is at least $a_i+\Delta_i+\min h_i-\varepsilon_i$. The midpoint of the two bounds is a common shattering threshold.
\end{proof}

\section{Certificates for consistent decoding and smoothing}\label{sec:unified-certificates}
The key technical complication is \emph{consistency}. A hardness
reduction must work against all legal parameter values, including
values on interval boundaries and values that do not look like the
intended bit strings. We use explicit witnesses to prove that every
assignment can occur, then separate estimates that show why an
unintended parameter cannot evade the clause tests. This is stronger
than showing a picture with exponentially many regions. The next
statements give an exact interface for these two obligations.

The main applications point to two logically distinct tasks. A \emph{statistical certificate} constructs many parameter codewords and evaluation inputs that distinguish their coordinates with a common margin. A \emph{computational certificate} makes the \emph{joint} evaluation of one parameter enforce a hard constraint system. Neither implication holds without the extra observation or constraint-compatibility assumption. We give explicit interfaces and preserve the two different quantitative resource requirements.

\subsection{Parameter codes, loss probes, and calibration}
\begin{definition}[Algorithmic parameter code and resource bound]\label{def:cert-code}
A $T$-bit parameter code is a map $B:\Theta\to\{0,1\}^T$ and a rational witness construction $\operatorname{Enc}:\{0,1\}^T\to\Theta$ satisfying $B(\operatorname{Enc}(b))=b$. A \emph{polynomial-bit certificate} also bounds the encoding length of the instances and witnesses and, if a Turing-model NP statement is sought, supplies a polynomial-time exact verification procedure for the relevant loss threshold. In real-analytic problems a root-comparison oracle is a separate model, not an automatic consequence of polynomial input bit length.
\end{definition}

\begin{definition}[Coordinate observation with interference]\label{def:cert-observe}
Coordinate $i$ is $(\Delta_i,\omega_i,\varepsilon_i)$-observable if an instance $x_i$ has, at code witnesses, losses
\[
f_{\operatorname{Enc}(b)}(x_i)=c_i+\Delta_i b_i+h_i(b_{-i})+e_i(b),
\]
with $\operatorname{osc}(h_i)\le\omega_i$ and $|e_i(b)|\le\varepsilon_i$. The instance must use the model's actual performance measure. An arbitrarily assigned trace loss does not certify an application theorem.
\end{definition}

\begin{proposition}[Robust statistical observation compiler]\label{prop:cert-stat}
If the $T$ coordinates are observable and $\Delta_i>\omega_i+2\varepsilon_i$ for every $i$, then
\[
\fat_\gamma(\mathcal F)\ge T,\qquad
\gamma\le\frac12\min_i(\Delta_i-\omega_i-2\varepsilon_i).
\]
\end{proposition}
\begin{proof}
This is Lemma~\ref{lem:robust} applied to the certified encoder.
\end{proof}

A computational reduction needs more structure. Some parameter values may fall outside the intended bit-code intervals or produce degenerate outputs. A sound reduction must rule out spurious low-loss configurations rather than merely verify the canonical witnesses.
\begin{definition}[Calibrated constraint-probe reduction]\label{def:cert-csp}
Given a Boolean constraint system $\Phi$ with $T$ variables and $M$ constraints, a polynomial-time constructor produces $N\le cM$ valid DDAD training instances $S_\Phi$, including any necessary \emph{calibration inputs}, and a polynomial-time decoder $D:\Theta\to\{0,1\}^T$ (or a decoder defined on all zero-loss parameters). It satisfies:
\begin{enumerate}[label=(\roman*)]
\item \emph{Completeness:} for every satisfying assignment $b$, $\operatorname{Enc}(b)$ yields zero loss on all inputs in $S_\Phi$.
\item \emph{Zero-loss soundness:} if every input has loss zero at $\theta$, then $D(\theta)$ satisfies $\Phi$.
\item \emph{Quantitative soundness, when claimed:} for a universal constant $K$,
\[
\operatorname{Unsat}_\Phi(D(\theta))\le K\cdot \#\{x\in S_\Phi:f_\theta(x)>0\},
\]
and every positive loss on this restricted family is at least $\eta>0$.
\end{enumerate}
The second item can be implemented by an irreversible violation state, a valid constraint implication, or a consistency/calibration gadget. It need not have an explicit Boolean evaluator inside the algorithm.
\end{definition}

\begin{theorem}[Calibrated CSP-to-ERM compiler]\label{thm:cert-csp}
If (i)--(ii) of Definition~\ref{def:cert-csp} hold, zero empirical loss is NP-hard to decide by reduction from the underlying satisfiability problem. If in addition the optimization language has polynomial-bit witnesses and exactly polynomial-time verifiable losses, its zero-loss decision problem is NP-complete. If (iii) holds, $N\le cM$, and the source problem has a perfect-completeness gap $\operatorname{Unsat}_\Phi(b)\ge\epsilon M$ for all $b$ in NO instances, then
\[
\inf_{\theta\in\Theta}\frac1N\sum_{x\in S_\Phi}f_\theta(x)
\ge \frac{\eta\epsilon}{Kc}
\]
in NO instances, whereas the YES optimum is zero. Thus fixed-additive approximation is NP-hard for any threshold smaller than this gap. No finite multiplicative approximation can find a zero-cost solution on all YES instances unless $\mathsf P=\mathsf{NP}$, provided its returned candidate can be evaluated in polynomial time.
\end{theorem}
\begin{proof}
Completeness and soundness give equivalence between source satisfiability and zero empirical loss, proving NP-hardness. Polynomial verifiability and the rational witness condition give NP membership when applicable. For the gap, quantitative soundness implies at least $\epsilon M/K$ positive-loss inputs. Each contributes at least $\eta$, so averaging over $N\le cM$ yields the displayed inequality. The approximation conclusions follow by comparing the zero and positive optimum cases.
\end{proof}

\paragraph{Why the calibration condition is substantive.}
For matching, the parameter code appears in online choices and neighbors of a late probe test a clause. For one aggregated CG cut, unary rounding gadgets constrain each literal multiplier to two intervals, consistency gadgets enforce complements, and a clause gadget tests a sum of at most three coordinates. For linkage, parameter guards, a prescribed merge clock and irreversible cross-target merges play the corresponding roles. In the B\&B clause-reading construction, by contrast, the history-dependent policy itself examines the encoded clause, a strictly stronger score model. For ordinary scalar Lasso, the dual-saturation mechanism (Lemmas~\ref{lem:lasso-replay-adj}--\ref{lem:lasso-adjacent-saturation}) prevents spurious zeros without a full-support assumption. These are not interchangeable algorithmic assumptions.

\subsection{Three locally verifiable code mechanisms}
\begin{proposition}[Direct-product composition]\label{prop:cert-product}
If $p$ parameter blocks independently realize $d_j$ observable bits with a common margin $\gamma$, and the tests for block $j$ are insensitive to all other blocks, then $\fat_\gamma\ge\sum_{j=1}^p d_j$. With bounded cross-block interference, the same conclusion follows from Proposition~\ref{prop:cert-stat}.
\end{proposition}
\begin{proof}
Choose each parameter block to realize its portion of an arbitrary labeling, and concatenate the independently feasible choices. Apply the observation criterion.
\end{proof}

\begin{proposition}[Signed-switch compilation]\label{prop:cert-switch}
Suppose a scalar parameter can implement upward and downward Boolean step functions at any of $K-1$ ordered thresholds and can add their costs within one legal evaluation input, up to additive error $\varepsilon$. If each switch has contribution $\Delta$, then $\lfloor\log_2K\rfloor$ inputs can be fat-shattered at any margin below $\Delta/2-\varepsilon$ on the raw cost scale.
\end{proposition}
\begin{proof}
Place a step between consecutive parameter witnesses. On input $i$, add an upward step at every $0\to1$ transition and a downward step at every $1\to0$ transition of bit $i$ in the binary expansions of the witness indices. The total is a constant plus $\Delta$ times the desired bit, up to the stated error.
\end{proof}

\begin{proposition}[Analytic comparison interpolation]\label{prop:cert-interpolation}
Suppose a legal score comparison can realize the sign of $\sum_{j=1}^K c_j\phi_j(\theta)$ for integer vectors $c$, and there are $K$ parameter witnesses with nonsingular evaluation matrix $V_{ij}=\phi_j(\theta_i)$. Then one can choose integer coefficients producing any prescribed binary sign vector on those witnesses. A uniform action-to-loss decoder gives $\fat_\gamma\ge\lfloor\log_2K\rfloor$ for its certified margin $\gamma$. The indicated rounding construction uses coefficients of magnitude at most $(\max\{1,\|V\|_\infty\}+1)\|V^{-1}\|_\infty+1$. Rescaling the basis may improve this bound.
\end{proposition}
\begin{proof}
For signs $y\in\{-1,1\}^K$, compute $c^*=V^{-1}y$. Choose $L=\max\{1,\lceil\|V\|_\infty\rceil\}$ and round $Lc^*$ coordinatewise to $c\in\mathbb Z^K$. The evaluation error is at most $\|V\|_\infty/2\le L/2$, preserving all signs strictly. The bound follows from $\|c\|_\infty\le L\|V^{-1}\|_\infty+1/2$, and the decoder supplies the performance margin.
\end{proof}

A fourth mechanism is \emph{persistent self-simulation}. Both sides of a parameter-dependent comparison lead to states satisfying a common inductive invariant. This creates $2^T$ execution traces with a succinct depth-$T$ instance and polynomial-bit dyadic witnesses. To infer $\fat_\gamma\ge T$ it still needs valid coordinate probes. In a frontier search such as B\&B, the invariant must cover dormant nodes and incumbent information, rather than just the active state. The matching, gradient-descent, and cutting-plane constructions illustrate several ways to realize the parameter code.

\begin{remark}[Clause-cost normal form and its expressivity requirement]\label{rem:clause-normal-form}
If a polynomial-bit parameter code realizes all $b\in\{0,1\}^T$
and legal instance losses are an additive sum of bounded-arity Boolean
factor costs, ERM reduces to optimizing that factor system, provided the
loss identity holds for every admissible parameter (or non-code values
are calibrated). Unrestricted two-literal clause factors already capture
Max-2-SAT, while unrestricted three-literal factors capture zero-loss
3-SAT. These are hardness statements only when the application can
realize the required clause factors. Factor arity alone is insufficient.
Unary factors with a uniform cardinality budget are solved by sorting
marginal costs, and bounded interaction treewidth permits dynamic
programming in $2^{O(w)}\poly(T,m)$ time. The relevant concrete probes
have different realizability and non-code soundness mechanisms:
onto parameter codes (matching), exclusion (batch CG), slot counting
(GD), and irreversible cross-target merges (linkage).
\end{remark}

\subsection{Proof contracts and the limits of the framework}
A complete certificate specifies its algorithm and solver policy, a polynomial-size parameter encoder and witness bit lengths, exact local comparisons and tie behavior, a persistent global execution invariant, loss decoding for the original objective, quantitative margins, and a Turing or exact-real evaluation model. For computational reductions, it must also establish completeness and soundness for every admissible parameter, and charge any malformed or inconsistent code to an explicit loss. A large set of output trees alone gives neither a fat-shattering bound nor ERM hardness. Similarly, a polynomial number of parameter regions is an output-size statement and becomes an ERM algorithm only with a suitable per-region enumeration and numerical evaluation oracle.

\subsection{A reusable smoothed pivotal-event compiler}
Hardness gadgets can make algorithmic behavior sensitive to fine parameter differences. For expected complexity bounds, conditioning on one independent input coordinate at a time can count output changes without enumerating adaptive execution states.
\begin{theorem}[Pivotal-coordinate area-formula compiler]\label{thm:cert-pivotal}
Let $Z=(Z_1,\ldots,Z_E)$ have independent coordinates, with conditional densities $f_e\le\kappa_e$. Fix a bounded scalar-parameter interval $\Theta$. Assume that, after fixing $Z_{-e}$, the $e$-pivotal events can be covered by finitely or countably many rectifiable graphs $Z_e=\vartheta_{e,j}(\theta)$ such that (i) their number through almost every $\theta$ is at most $R_e$; (ii) $|\vartheta_{e,j}'(\theta)|\le V_e$ almost everywhere. And (iii) every actual parameter value at which the algorithm output changes is charged to at least one such pivotal event, outside a probability-zero degeneracy set. Then the expected number $K(Z;\Theta)$ of output transitions satisfies
\[
\mathbb E K(Z;\Theta)\le |\Theta|\sum_{e=1}^E\kappa_e R_eV_e.
\]
The same conclusion holds with a nonconstant integrable speed envelope $V_e(\theta)$, replacing $V_e|\Theta|$ by $\int_\Theta V_e(\theta)\,d\theta$.
\end{theorem}
\begin{proof}
Fix $e$ and condition on $Z_{-e}$. By the one-dimensional area formula, the expected number of intersections between the random horizontal line $Z_e=z$ and the pivotal graphs is bounded by
\[
\int f_e(z)N_e(z)\,dz
\le\kappa_e\int_\Theta\sum_j|\vartheta_{e,j}'(\theta)|\,d\theta
\le\kappa_e R_eV_e|\Theta|.
\]
Each actual transition is charged to some coordinate, so summing over $e$ and taking expectations proves the claim. Replacing the constant speed bound by an integrable envelope gives the variant.
\end{proof}

\paragraph{Why this is distinct from dispersion.}
The theorem controls changes of a \emph{single algorithm output path} under input smoothing. It can imply the bounded-density of parameter breakpoints needed for dispersion over multiple independent training inputs, but its stronger use is output-sensitive exact ERM once the transitions can be enumerated. For power-mean linkage, Appendix~\ref{sec:linkage:smoothed-proof}
verifies these conditions using a deterministic one-edge execution bound,
an entropy--variance identity, and extremal-edge charging. This yields
Theorem~\ref{thm:linkage:smoothed} across the full real exponent line,
including geometric linkage at $\alpha=0$.

\begin{table}[t]\centering\small
\caption{Common hardness and statistical mechanisms. The same parameter code is used for different proof purposes. The final column describes the application-specific constraints.}\label{tab:unified-mechanisms}
\begin{tabularx}{\linewidth}{@{}p{.20\linewidth}p{.22\linewidth}p{.27\linewidth}X@{}}\toprule
Problem & Code source & Computational loss decoder & Statistical decoder / qualifier\\\midrule
Power-mean linkage & Nested analytic merge comparisons & Irreversible clause probe, supported by a metric, guards, and a full merge schedule & Root-Hamming and Dasgupta-value hardness with exponentially small parameter margins\\
One aggregated CG cut & Independent rounded multiplier sums & Quantization, consistency and at-most-three-row clause tests & Unary root-gap and tree-size tests, with separate arithmetic digit gadgets\\
Online matching & Dyadic history-dependent decisions & Late-neighbor probes realize Boolean clauses & Single-neighbor probes have constant regret margin\\
B\&B node selection & History-dependent frontier code & Clause-reading controller activates an expensive parity trap & Designated parity trap dominates interference\\
Lasso sign path & Active-set/sign changes via KKT & Adjacent-pair and top/anchor readouts force saturated dual signs. Dual replay gives soundness & Ordinary squared validation loss, $r+3$ rows, compact positive interval\\
Decision trees & Nested prunings or analytic split changes & No general clause verifier for usual scalar parameters & Held-out labels, with constant margins in restricted tree families\\\bottomrule
\end{tabularx}
\end{table}

\section{Reusable decision codes, observation, and simulation}
\subsection{The interfaces: what actually needs to be constructed}
Let a parameterized algorithm have parameters $\theta\in\Theta$ and legal problem instances $x\in\mathcal X$, with ordinary loss $f_\theta(x)$. We distinguish (i) \emph{parameter encoding}, (ii) \emph{legal algorithmic realization}, (iii) \emph{coordinate observation}, and (iv) \emph{joint constraint verification}. Encoding alone implies neither pseudo-dimension nor ERM hardness.
\begin{definition}[Polynomial-bit code]
A $d$-bit code is a deterministic $B:\Theta\to\bits^d$ and an encoder $\Enc:\bits^d\to\Theta$ with $B(\Enc(b))=b$. For a Turing-model claim, $\Enc$ must run in polynomial time and output rational parameters of polynomial bit length. The actual algorithm must implement $B$ on its intended instance family, including deterministic tie conventions.
\end{definition}
\begin{theorem}[Robust observation compiler]\label{thm:robust}
The efficient-code witnesses of Definition~\ref{def:probe} satisfy the
fat-shattering bound of Lemma~\ref{lem:robust} whenever their legal
loss probes obey its signal, interference, and error hypotheses.
\end{theorem}

\begin{remark}[Resources that must accompany every capacity theorem]
State separately the bit count $d$, number of problem objects $n$, number of real parameters $k$, total input bit complexity, width of the witness parameter cells, and the \emph{actual loss margin after normalization}. In particular, $2^d$ distinct execution traces alone do not prove $\Pdim\ge d$.
\end{remark}

\subsection{Three reusable parameter-code generators}
\begin{lemma}[Nested-interval/decision-clock encoder]\label{lem:interval}
For every $w\in\bits^{\le d}$, suppose the parameter line contains a nonempty interval $J_w$ such that $J_{w0},J_{w1}$ are disjoint subintervals of $J_w$, and the algorithm at decision $|w|+1$ chooses bit $a$ at every parameter in the interior of $J_{wa}$ whenever its earlier history is $w$. Suppose furthermore that for every $b\in\bits^d$ one can find a rational interior point of $J_b$ with polynomial bit complexity. Then a polynomial-bit $d$-bit code exists. If the pairwise tests have threshold error bounded by $\xi_j$, it suffices to show that every comparison is separated from its threshold by more than $\xi_j$ at each intended witness.
\end{lemma}
\begin{proof}
Choose $\theta_b\in\operatorname{int}(J_b)$, and induct on the prefix length. At stage $j$ it belongs to $J_{b_{1:j}}$ and therefore takes the prescribed branch. The separation hypothesis preserves the decisions in the presence of perturbation.
\end{proof}
\begin{recipe}[Recursive self-simulation]\label{rec:self}
Build a state invariant $q_w$ and a normalized parameter $\nu_w(\theta)$. At each stage, a common normalized threshold divides the viable interval into two cells. Show that each successor has the \emph{same type} of state invariant after renormalization, and that a final rational point can be constructed with polynomial bits. Separately prove that no other candidate action interrupts the intended stage. This is the structure behind adaptive matching and the nested analytic linkage clock. The latter uses the metric-realization and scheduling arguments in Appendix~\ref{app:linkage-complete}.
\end{recipe}
\begin{lemma}[Signed-switch synthesis]\label{lem:switch}
Let $\theta_0<\cdots<\theta_{K-1}$ be $K=2^d$ ordered witnesses. Suppose a legal subgadget can implement either $\ind\{\theta>\tau\}$ or $\ind\{\theta<\tau\}$ for every threshold between consecutive witnesses. Suppose up to $K-1$ such subgadgets compose into one legal evaluation instance whose loss is a common baseline plus $\Delta$ times their sum, with uniform additive error at most $\varepsilon$. Then $d$ evaluation instances can $\gamma$-fat-shatter the $K$ witnesses for $\gamma\le\Delta/2-\varepsilon$, provided this is positive.
\end{lemma}
\begin{proof}
For each binary digit of the witness index, add an upward switch at every $0\to1$ transition and a downward switch at every $1\to0$ transition. As the parameter crosses a boundary, the switch sum changes by exactly the desired $\pm1$, so it equals the requested digit plus a constant independent of the witness. The approximation error reduces the two-class gap by at most $2\varepsilon$.
\end{proof}
\begin{lemma}[Comparison-basis interpolation with integer coefficients]\label{lem:interp}
Suppose $K$ legal comparison features $\phi_j(\theta)$ are available. At $K$ parameter values $\theta_s$, let $V_{sj}=\phi_j(\theta_s)$ be nonsingular. Suppose every integer coefficient vector $C\in\mathbb Z^K$ can be realized by a legal score comparison with sign equal to $\operatorname{sign}(\sum_j C_j\phi_j(\theta_s))$, and the selected action has a loss decoder with values $0,1$. Then $\fat_{1/2}\ge\lfloor\log_2K\rfloor$. More quantitatively, let $Q=\max_s\sum_j|V_{sj}|$ and $L=\max\{1,\lceil Q\rceil\}$. For each target labeling $y\in\{-1,+1\}^K$ choose
\[
C=\operatorname{round}(LV^{-1}y),\qquad
\|C\|_\infty\le L\|V^{-1}\|_\infty+\tfrac12.
\]
Then $\|VC-Ly\|_\infty\le Q/2\le L/2$, so every requested comparison sign is strict.
\end{lemma}
\begin{proof}
The bound follows from coordinate rounding. Each coefficient error is at most $1/2$. Assign the $d$ binary-digit label vectors to the $d$ evaluation instances.
\end{proof}
\begin{remark}
Nonsingularity is not a physical realization theorem. One must prove that \emph{signed} coefficients can be implemented, that the required integer multiplicities and data sizes are affordable, that computed coefficients have polynomial-bit representations, and that the comparison controls the original loss. For Tsallis stumps these are separate class-count and sample-size constraints.
\end{remark}

\subsection{Reusable loss transducers and capacity transfers}
\begin{lemma}[Exact affine-KKT loss transducer]\label{lem:kkt}
Let $G=X^\top X\succ0$, $c=X^\top y$, and suppose a fitted model obeys
\[
G\widehat w(\lambda)+2\eta\widehat w(\lambda)=c-\lambda g(\lambda),
\]
where $\eta\ge0$ is \emph{fixed} and $g(\lambda)$ is a valid $\ell_1$-subgradient. For any rational $u$, form the validation feature $z=(G+2\eta I)u$ and target $t=u^\top c$. Then its ordinary residual is
\[
z^\top\widehat w(\lambda)-t=-\lambda u^\top g(\lambda).
\]
If $g$ is a full-support sign vector and $p$ is a fixed coordinate of sign $+1$, taking $u=(e_i-e_p)/\lambda_0$ yields squared losses $0$ or at least $4$ at all sign witnesses with $\lambda\ge\lambda_0$, according to whether $g_i=+1$ or $g_i=-1$. Therefore these witnesses yield $\fat_2\ge p-1$ for the \emph{raw, unbounded-feature} squared-validation class.
\end{lemma}
\begin{proof}
Multiply the KKT identity by $u^\top$. At the full-support witnesses $g_i,g_p\in\{\pm1\}$, $g_p=1$, and the squared difference is either zero or $4(\lambda/\lambda_0)^2$.
\end{proof}
\begin{lemma}[Quantitative performance transfer]\label{lem:transfer}
Suppose $\gamma$-fat-shattering is witnessed by parameters $\theta_b$ and inputs $x_i$ in a source family. Assume a target family has valid embeddings $\iota(\theta_b)$ and $\tau(x_i)$ satisfying
\[
|f^{\mathrm{target}}_{\iota(\theta_b)}(\tau(x_i))-[a_i f^{\mathrm{source}}_{\theta_b}(x_i)+c_i]|\le\varepsilon_i,
\qquad a_i>0.
\]
Then the target has fat-shattering dimension at least $d$ at any margin
\[
0<\gamma'\le\min_i(a_i\gamma-\varepsilon_i).
\]
\end{lemma}
\begin{proof}
Transfer the source threshold $r_i$ to $a_ir_i+c_i$ and apply the error bound to both labels.
\end{proof}
\begin{lemma}[Direct sum of independent parameter blocks]
If $\Theta$ contains $\Theta_1\times\cdots\times\Theta_s$, and disjoint evaluation-instance families depend on their corresponding parameter block alone and have fat-shattering certificates $(d_j,\gamma_j)$, then the joint class has dimension at least $\sum_jd_j$ at margin $\min_j\gamma_j$.
\end{lemma}
\begin{proof}Choose the parameter blocks independently for each requested labeling.\end{proof}
\begin{lemma}[Stable sign witnesses under positive quadratic regularization]\label{lem:enet}
Let $G\succ0$, $w_0=G^{-1}(c-\lambda s)$ be a full-support Lasso witness with $\min_j|(w_0)_j|=\delta>0$. Let $w_\eta=(G+2\eta I)^{-1}(c-\lambda s)$. If
\[
\frac{2\eta}{\lambda_{\min}(G)}\|w_0\|_2<\delta,
\]
then $w_\eta$ has the same sign vector $s$ and satisfies the full-support Elastic Net KKT equations with quadratic penalty $\eta\|w\|_2^2$. In particular the corresponding sign state persists. Scaling $(X,y)$ by $K$ makes a fixed positive quadratic penalty effectively $\eta/K^2$ after dividing the objective by $K^2$.
\end{lemma}
\begin{proof}
The resolvent identity gives $w_\eta-w_0=-2\eta(G+2\eta I)^{-1}w_0$. The spectral norm bound gives the displayed sufficient condition for coordinatewise sign preservation. Once signs agree, the full-support KKT conditions certify the solution by strict convexity.
\end{proof}
\begin{remark}
The KKT transducer does \emph{not} directly prove hardness of standard scalar Lasso AIC. Its loss is training RSS plus a positive support penalty, not an arbitrary validation readout. Nor does a constant raw squared-loss margin imply a constant margin under bounded validation features or bounded loss normalization.
\end{remark}

\begin{lemma}[Ridge Cauchy interpolation as a loss transducer]\label{lem:ridge}
Consider ridge regression with $G=\operatorname{diag}(\mu_1,\ldots,\mu_p)\succ0$ and training sufficient statistic $c=\mathbf1$. Then $\widehat w_j(\lambda)=(\mu_j+\lambda)^{-1}$. For distinct positive rational nodes $\lambda_1,\ldots,\lambda_p$ and distinct positive rational $\mu_1,\ldots,\mu_p$, the evaluation matrix
\[
V_{sj}=\frac{1}{\lambda_s+\mu_j}
\]
is nonsingular by the Cauchy determinant formula. If arbitrary rational validation features are legal, one can interpolate \emph{any} prescribed rational validation predictions at the $p$ nodes. Choosing $2^d\le p$ nodes and one feature vector per binary digit, with target zero, yields squared losses exactly $0$ or $1$ and hence $\fat_{1/2}\ge\lfloor\log_2p\rfloor$ for this unbounded-validation-feature family. If the rational nodes have polynomial-bit descriptions, the inverse-Cauchy coefficients do as well, by determinant bounds.
\end{lemma}
\begin{proof}
The ridge solution has the claimed coordinates by solving $(G+\lambda I)w=c$. Nonsingularity of $V$ lets us choose $r=V^{-1}y$ for each desired prediction vector $y\in\bits^p$. The squared validation error against target zero equals $y_s^2=y_s$ at the chosen nodes.
\end{proof}
\begin{remark}
To realize $G$ and $c$ by \emph{rational} training data, one may take $\mu_j$ to be rational squares and use a rational diagonal design with matching rational responses. The Cauchy proof is a legal-observation route, not a constant-margin guarantee for uniformly bounded features. It also explains the commonality between ridge (Cauchy basis) and the Tsallis stump construction (scaled Vandermonde/exponential basis). Both reduce the statistical argument to a nonsingular evaluation matrix plus a legal coefficient decoder.
\end{remark}

\subsection{Precision-preserving simulation and scheduling}
\begin{lemma}[Exact cancellation with a robust outer schedule]\label{lem:cancellation}
Fix a compact $\alpha$-interval $I\subset(0,\infty)$ and $c\in[1,2]$. Suppose two clusters $A,B$ have the same cardinality $s$. A probe point has $W$ distances equal to $c$ into each cluster, with at most $E=s-W$ remaining distances in $[1,2]$. For power-mean linkage $D_\alpha$, write $P_A(\alpha)$ and $P_B(\alpha)$ for the sums of the exceptional $\alpha$th powers. Then
\[
s\bigl(D_\alpha(p,A)^\alpha-D_\alpha(p,B)^\alpha\bigr)
=P_A(\alpha)-P_B(\alpha)\quad\text{exactly}.
\]
There is a constant $C_I$ depending only on $I$ such that
\[
|D_\alpha(p,A)-c|,\ |D_\alpha(p,B)-c|
\le C_I E/s\qquad(\alpha\in I).
\]
Thus an exponentially small decision signal $P_A-P_B$ can be preserved \emph{exactly}, while an inverse-polynomial separation between successive baseline scales $c$ controls the global merge order when $E/s$ is sufficiently small.
\end{lemma}
\begin{proof}
The $Wc^\alpha$ baseline terms cancel in the two power sums, and the two normalization denominators are equal. The remaining powers lie in a compact positive interval. The mean-value theorem for the map $x\mapsto x^{1/\alpha}$, uniformly over the compact range of $(\alpha,x)$, bounds the score perturbation by a constant times the exceptional fraction $E/s$.
\end{proof}
\begin{lemma}[Stage-isolation induction]\label{lem:isolation}
Suppose an algorithm has a prescribed sequence of states $q_0,\ldots,q_d$. At state $q_{j-1}$, exactly two intended actions $a_j^0,a_j^1$ encode the next bit, every unintended action has score worse than both intended actions by at least $\eta_j>0$, and the comparison of the intended actions at a designated parameter witness has the desired sign with gap greater than $2\xi_j$. Suppose all score perturbations from padding, rounding, or numerical approximation are strictly smaller than $\min\{\eta_j/2,\xi_j\}$. If either intended action establishes the invariant for $q_j$ and no earlier terminal event occurs, then the entire requested $d$-bit execution is realized at that witness.
\end{lemma}
\begin{proof}
At each stage the isolation gap excludes all unintended actions. The intended comparison retains its sign under two-sided error of size $\xi_j$. The invariant advances to the next state, so induction completes the proof.
\end{proof}
\begin{remark}
The exact-cancellation lemma alone proves only a \emph{local score identity}. The global cluster-pair comparisons, guard/probe overlap and cleanup are certified separately in Appendix~\ref{app:linkage-complete}. This distinction remains important for other linkage variants and for the separate smoothing analysis.
\end{remark}
\begin{lemma}[Rational representatives of nonempty linear decision cells]\label{lem:rational}
Suppose a nonempty parameter cell in a bounded rational polytope is specified by polynomially many rational weak and strict linear inequalities, with polynomial-bit coefficients. Then the cell contains a rational point of polynomial bit length, computable by linear programming. Consequently, polynomial-bit parameter witnesses follow automatically for \emph{linear} decision cells, but not for general transcendental root cells.
\end{lemma}
\begin{proof}
Introduce a common slack $\tau$ in all strict inequalities, constrain $0\le\tau\le1$, and maximize $\tau$. A point satisfying all strict inequalities has a positive common slack since there are only finitely many. A rational basic feasible solution to the resulting bounded linear program has polynomial bit length by determinant bounds and a positive objective, and hence satisfies the original cell inequalities.
\end{proof}
\begin{lemma}[Finite-certificate normalization]\label{lem:normalize}
Suppose an unnormalized loss certificate has high-low gap at least $\Delta$ and perturbation/interference total at most $E<\Delta$. If the normalized loss is $f^{\rm norm}=f^{\rm raw}/H$ for a positive common $H$, the same witnesses have guaranteed fat margin $(\Delta-E)/(2H)$. If the normalization $H$ grows with instance size, a constant \emph{raw} margin need not give a constant normalized margin.
\end{lemma}
\begin{proof}Divide the threshold and the two endpoint inequalities by $H$.\end{proof}

\section{Constraint verification and legal loss gadgets}
\subsection{Constraint probes, calibration, and computational hardness}
\begin{theorem}[Exact clause-probe compiler]\label{thm:clause}
Suppose a polynomial-bit code $B:\Theta\to\bits^d$ is onto. For every Boolean clause $C$ of width at most three, suppose one can construct a legal input $x_C$ with loss
\[
f_\theta(x_C)=\Delta\ind\{C(B(\theta))\text{ is false}\},\qquad\Delta>0,
\]
\emph{for all} $\theta\in\Theta$, including tie cases. Then zero empirical loss is NP-hard via 3-SAT. It is NP-complete if zero loss can be verified from polynomial-size rational witnesses in polynomial time. Exact ERM with width-two clauses is NP-hard via Max-Cut; unary probes permit polynomial-time ERM on their own. The unary loss probes also give $\fat_{\Delta/2}\ge d$.
\end{theorem}
\begin{proof}
The loss sum equals $\Delta$ times the number of violated clauses. For an edge $ij$ of a Max-Cut instance, clauses $(b_i\lor b_j)$ and $(\neg b_i\lor\neg b_j)$ together contribute exactly one violation if the endpoints agree and zero if they differ. Unary clauses decouple. Pseudoshattering follows from coordinate observations and Theorem~\ref{thm:robust}.
\end{proof}
\begin{definition}[Calibrated constraint system]
Let a reduction produce $M$ predicate probes and $N_{\mathrm{cal}}$ calibration probes. For a parameter $\theta$, let $F_{\mathrm{test}}$ and $F_{\mathrm{cal}}$ count the failing probes. A decoder $D(\theta)$ is \emph{chargeably calibrated} if some set $\operatorname{Bad}(\theta)$ of source coordinates obeys
\[
|\operatorname{Bad}(\theta)|\le aF_{\mathrm{cal}}(\theta),
\]
and every source constraint violated by $D(\theta)$ whose variables avoid $\operatorname{Bad}(\theta)$ has its associated predicate probe fail. Let $B$ upper bound the number of source constraints containing any one variable.
\end{definition}
\begin{lemma}[Defect-charge lemma]\label{lem:charge}
Under chargeable calibration,
\[
\Unsat_\Phi(D(\theta))\le F_{\mathrm{test}}(\theta)+aB F_{\mathrm{cal}}(\theta)
\le \max\{1,aB\}F(\theta),
\]
where $F=F_{\mathrm{test}}+F_{\mathrm{cal}}$. If every failing probe incurs loss at least $\delta>0$, the total number of probes $N\le cM$, and a perfect-completeness source gap guarantees $\Unsat_\Phi(b)\ge\epsilon M$ for all $b$ in NO instances, then
\[
\inf_\theta\frac1N\sum_x f_\theta(x)\ge
\frac{\delta\epsilon}{c\max\{1,aB\}}.
\]
For YES instances, if every satisfying assignment has a parameter with all probes zero, the optimum is zero. This yields a constant additive hardness gap when $a,B,c,\delta,\epsilon$ are constants.
\end{lemma}
\begin{proof}
Any violated constraint either touches a bad coordinate (at most $B|\operatorname{Bad}|$ such constraints) or has a failed predicate probe. Charge the former to calibration failures and the latter to predicate failures. The source gap forces at least $\epsilon M/\max\{1,aB\}$ failed probes. Divide their minimum total loss by $N\le cM$.
\end{proof}
\begin{remark}
If the losses have positive baselines, replace the zero-optimum statement by a rigorously proved baseline subtraction or by a threshold with a quantitative gap. Zero optimum and finite-factor inapproximability must not be asserted for an objective that cannot attain zero. NP membership also requires polynomial-time verification of the returned \emph{actual loss}, not only a short parameter description.
\end{remark}

\begin{theorem}[Algebraic-validity decoder for continuous optimization]\label{thm:algebraic}
Let $h:\Theta\to\mathbb R^p$ be a parameter-dependent state and $q:\Theta\to(0,\infty)$ a positive scalar. Suppose legal evaluation instances implement the ordinary loss
\[
L_\theta=q(\theta)^2\|Mh(\theta)-t\|_2^2,
\]
for polynomial-bit rational $M,t$. Assume (i) every satisfying assignment $b$ of a source NP-complete problem has a polynomial-bit parameter witness with $Mh(\theta_b)=t$. And (ii) whenever $Mh(\theta)=t$, a polynomial-time computable map extracts a satisfying assignment $b$ from $h(\theta)$. Then zero-loss feasibility is NP-hard. If the parameter domain is compact, $q$ stays positive, and its zero-loss feasibility admits polynomial-bit, polynomially checkable certificates, the corresponding exact ERM zero-loss decision is NP-complete.
\end{theorem}
\begin{proof}
Positivity of $q$ implies that zero squared loss is equivalent to $Mh=t$. The two implications give completeness and soundness of the reduction. Compactness makes minimum zero equivalent to attainment, and polynomial certificates give membership in NP.
\end{proof}
\begin{recipe}[Two-stage dual-integrality check for Lasso]\label{rec:dual}
Build a maximal Lasso path whose full-support signs include all required binary assignments. Represent validation residuals as $-\lambda u^\top g(\lambda)$ via Lemma~\ref{lem:kkt}. The \emph{validity rows} must force the fractional dual $g(\lambda)\in[-1,1]^p$ to lie on the desired Boolean sign face. Only then add the arithmetic instance-dependent row. In the adjacent-pair construction, the validity rows consist of an anchor equality, pair equations $g_{2j-1}+g_{2j}=0$, and a final top-pair equation. A nonzero anchor plus the specific Mairal--Yu replay invariant forces each pair to be opposite saturated signs. The final row tests $\sum_jd_j\sigma_j=T$. One must separately prove replay at \emph{partial-support states}, full-support sign coverage, a positive rational cutoff $\lambda_{\rm lo}>0$, and polynomial-bit rational KKT verification. This is a different compiler from a literal clause probe and does not transfer automatically to AIC.
\end{recipe}
\begin{lemma}[Bounded-corruption replication]\label{lem:replicate}
Suppose a Boolean-decoding gadget can be inconsistent on at most $r$ source variables for any parameter, and a zero-loss predicate system otherwise faithfully enforces every source constraint. Construct $r+1$ disjoint copies of a source formula $\Phi$. If all probes in all copies have zero loss, at least one copy contains no inconsistent variable and therefore supplies a satisfying assignment to $\Phi$. For optimization, if corruption can improve the total objective by at most $D$, using $R>D/\delta$ disjoint copies preserves an integer-valued source optimum gap $\delta>0$, provided the uncorrupted copies each incur the corresponding true source loss.
\end{lemma}
\begin{proof}
At most $r$ disjoint copies contain a corrupted variable, so one of the $r+1$ copies is uncorrupted. For optimization, the additive benefit of corruption is bounded by $D$ while a source optimum gap $\delta$ replicated $R$ times is $R\delta$.
\end{proof}
\begin{remark}
A useful scalar-CG arithmetic application is to take pairwise coprime frequencies $q_i$. At any $\theta\in(0,1)$ at most one coefficient $q_i\theta$ can be integral, since otherwise a rational number with denominator dividing two coprime integers would be an integer. This creates a one-coordinate corruption bound. The duplication lemma repairs zero-loss soundness. It does not turn a batch of separately rounded cuts into a single aggregated cut.
\end{remark}

\subsection{A library of legal application gadgets}
\begin{lemma}[CG rounding windows]\label{lem:cg}
Let $S$ be a nonempty set of constraint rows and $\sigma=\sum_{t\in S}u_t$ for one shared CG multiplier vector $u\in[0,1]^m$. Consider one-variable integer programs with a dummy zero-valued second variable (to absorb unused rows). For integers $a>\beta\ge1$:
\begin{align*}
L(a,\beta;S):&\quad\max x\ \ \text{s.t. }ax\le\beta,\ x\in\mathbb Z_{\ge0};\\
G(a,\beta;S):&\quad\max(-x)\ \ \text{s.t. }-ax\le-\beta,\ x\in\mathbb Z_{\ge0}.
\end{align*}
After generating one aggregated CG cut from the row multipliers, the root LP becomes integral exactly under the respective conditions
\[
L:\quad\sigma\in[1/a,1/\beta),\qquad
G:\quad\sigma>0\ \text{and}\ \lceil a\sigma\rceil=\lceil\beta\sigma\rceil.
\]
Outside these conditions the root solution is fractional and, without presolve or further cuts, branching on $x$ generates exactly three nodes (including the root).
\end{lemma}
\begin{proof}
For $L$, the cut is $\lfloor a\sigma\rfloor x\le\lfloor\beta\sigma\rfloor$, which forces $x=0$ exactly when the first floor is positive and the second zero. For $G$, the cut is $\lceil a\sigma\rceil x\ge\lceil\beta\sigma\rceil$. Integrality at $x=1$ occurs precisely when the positive integer coefficients coincide. Otherwise the remaining LP optimum is strictly between the integer bounds $0$ and $1$, producing two children.
\end{proof}
\begin{remark}[Structural binary-domain convention]
The $G$-gadget inequality alone leaves its LP feasible region unbounded,
even though the objective has a finite attained maximum. In the restricted
CG model, the nontrivial variable has structural domain $x\in\{0,1\}$,
and its LP relaxation includes $0\le x\le1$. These bounds are not
additional rows whose multipliers can be learned. Introducing freely
weighted bound rows would define a different configuration class.
\end{remark}
\begin{proposition}[CG Boolean gate library]\label{prop:cg}
One unary $L(8,1)$ and one unary $G(7,4)$ constrain each literal multiplier to one of the two intervals
\[
J_0=[1/8,1/7],\qquad J_1=(1/4,2/7].
\]
A two-row $L(3,2)$ closes its root gap iff exactly one of its two inputs lies in $J_1$. A three-row $L(2,1)$ closes iff at least one of its three inputs lies in $J_1$. Thus these are legal quantization, XOR-consistency, and clause-OR gates for a \emph{single aggregated cut per instance}.
\end{proposition}
\begin{proof}
The first two windows intersect in $J_0\cup J_1$. The sums $J_0+J_0$, $J_1+J_1$, and $J_0+J_1$ are separated by the window $[1/3,1/2)$. Three inputs have sum in $[1/2,1)$ iff at least one is in $J_1$.
\end{proof}
\begin{lemma}[Irreversible violation transducer]\label{lem:irreversible}
Suppose an algorithm never splits apart a previously merged component. If a legal constraint probe sends one point designated for target side $L$ into a component containing a point of target side $R$ exactly when a constraint is violated, then any eventual root partition into the two target sides is impossible after that violation. If separate guards and a cleanup invariant guarantee the correct root split when no violations occur, the root-Hamming loss is a zero-loss constraint verifier.
\end{lemma}
\begin{proof}
The mixed component persists until termination and thus cannot be contained in either target root child. The converse relies on the stated guard and cleanup hypotheses and is not implied by irreversibility alone.
\end{proof}
\begin{lemma}[Frontier-cost amplifier]\label{lem:frontier}
Assume a search procedure has $d$ stages, each generating a continuation node and a trap root. Exploring trap $j$ before continuation creates precisely $h_j$ extra descendants, finds no incumbent, and returns to the same stage invariant. If deferred trap roots are pruned after a terminal incumbent without descendant expansion and the backbone has constant size $C$, then its generated-node count is
\[
\operatorname{TreeSize}(\theta_b)=C+\sum_{j=1}^d h_jb_j.
\]
A final trap of size $H$ triggered exactly when a Boolean clause of $b$ is false adds $H\ind\{C(b)\text{ false}\}$.
\end{lemma}
\begin{proof}
Charge only additional descendants to each explored trap. The continuation invariant prevents double counting, and terminal incumbent pruning prevents descendants of deferred traps. The final term follows from the probe rule.
\end{proof}
\begin{remark}[B\&B gadget assumptions]
The current clause-probe implementation uses a history-dependent node selector that reads the clause from instance coefficients and explicitly evaluates it. The lemma by itself does not implement the final predicate using a natural static node score or the min/max branching-score mixture.
\end{remark}
\begin{recipe}[Other legal loss decoders]
Online matching. Reserve a late high-weight arrival adjacent to offline vertices that remain free precisely when a literal is true. Prove the offline optimum is parameter-independent. Linkage. Use a clause probe whose intended merge is into an irreversible wrong-side component. Additionally prove a complete metric distance table, every merge-order inequality, global guards, and cleanup. Decision trees. Select a held-out label where the chosen leaf or split yields binary error. Establish legal training construction, equal dataset sizes, and no premature changes to the tree. Lasso. Use Lemma~\ref{lem:kkt}, or support observations with positive thresholds. The common issue is that the physical algorithm and ordinary loss must realize the desired Boolean gate.
\end{recipe}

\section{Robustness and smoothed complexity results}\label{app:smoothed-certificates}
This appendix establishes the perturbation models, robust CG and
branch-and-bound hardness constructions, and independent-edge
smoothed-complexity bounds. In all smoothed hardness claims, perturbation is applied consistently to the \emph{numerical input} after the reduction, with any declared structural zeros and variable integrality preserved. When a perturbation makes integer constraint coefficients irrational, the cut remains valid because CG rounding only requires that decision variables be nonnegative integers. The computational model must nevertheless specify rational finite-precision perturbation samples to derive a Turing lower bound.

\subsection{A robust single-aggregated-CG gate library}
Fix $\varepsilon=1/40$ and independently perturb each designated nonzero coefficient and right-hand side of the four original one-variable gadgets by a relative factor in $[1-\varepsilon,1+\varepsilon]$. Include an additional unary gate $G(17,15;\{t\})$ for every literal row $t$. Use the unary gates $L(8,1;\{t\})$, $G(7,4;\{t\})$, the complement $L(3,2;\{t,t'\})$, and each 3-literal clause $L(2,1;S)$. Use an explicit binary bound $x\in\{0,1\}$ and a nonnegative dummy variable fixed to zero by dummy rows. Bounds on binary variables are regarded as part of the variable domain, not as extra rows participating in the CG multiplier vector. This resolves the otherwise unbounded feasible relaxation for a $G$-gadget without introducing a new freely weighted cut row.

\begin{lemma}[Perturbed coefficient-to-window inclusions]\label{lem:robust-cg-window}
If $L(a,\beta;S)$ closes its root gap after adding one aggregated CG cut, then its multiplier sum $s_S=\sum_{t\in S}u_t$ must lie in
\[
\left[\frac{1}{a(1+\varepsilon)},\frac{1}{\beta(1-\varepsilon)}\right).
\]
Conversely closure is guaranteed throughout the smaller interval
\[
\left[\frac{1}{a(1-\varepsilon)},\frac{1}{\beta(1+\varepsilon)}\right).
\]
For the perturbed $G$-gadget, closure requires $\lceil A(u)\rceil=\lceil B(u)\rceil>0$, where $A(u)=\sum_{t\in S}a_t'u_t$ and $B(u)=\sum_{t\in S}\beta_t'u_t$ (the expressions are positive after removing the common negative sign). For every admissible perturbation, simultaneously closing the three unary gates $L(8,1)$, $G(7,4)$, $G(17,15)$ forces
\[
u_t\in J_0^\varepsilon\cup J_1^\varepsilon,\quad
J_0^\varepsilon\subseteq\left[\frac{16}{123},\frac{40}{273}\right],\qquad
J_1^\varepsilon\subseteq\left[\frac{32}{123},\frac{80}{273}\right].
\]
\end{lemma}
\begin{proof}
The cut on an $L$ gadget has the form $\lfloor A(u)\rfloor x\le\lfloor B(u)\rfloor$. It forces $x=0$ iff $A(u)\ge1$ and $B(u)<1$. The stated interval inclusions follow by bounding each independently perturbed coefficient. For a $G$ gadget the inequality is $\lceil A(u)\rceil x\ge\lceil B(u)\rceil$. The LP bound closes iff these two positive ceilings agree. Since $7(1-\varepsilon)>4(1+\varepsilon)$, simultaneous ceiling agreement for $G(7,4)$ can only occur at ceiling level $1$ or $2$. Level $1$ forces $u_t\le1/[7(1-\varepsilon)]=40/273$. Combining with $G(17,15)$ forces the latter's ceiling level to be $3$ and thus $u_t>2/[15(1+\varepsilon)]=16/123$. Similarly level $2$ for $G(7,4)$, combined with the possible $G(17,15)$ ceiling levels, forces level $5$, giving $u_t>4/[15(1+\varepsilon)]=32/123$ and $u_t\le2/[7(1-\varepsilon)]=80/273$. The $L(8,1)$ gate excludes $u_t=0$ and any remaining lower intervals. All inequalities are inclusions of the feasible windows, not claims of exact endpoints after arbitrary independent perturbations.
\end{proof}

\begin{proposition}[Uniform complement and clause soundness]\label{prop:cg-robust-soundness}
For all independently perturbed data with relative errors at most $1/40$, if every unary gate for a variable closes, then the complement gadget $L(3,2)$ can close only when exactly one of its two literal multipliers belongs to the upper interval $J_1^\varepsilon$. If all three literal rows of a 3-clause belong to $J_0^\varepsilon$, its $L(2,1)$ clause gadget cannot close. Moreover, assigning $u_t=29/210$ to false literals and $u_t=29/105$ to true literals closes all three unary gates, all consistency gates, and every satisfied clause gate \emph{for every perturbation in the box}.
\end{proposition}
\begin{proof}
Closure of the complement gate implies $u+v\in[40/123,20/39)$. Yet two low values sum to at most $80/273<40/123$, while two high values sum to at least $64/123>20/39$. Consequently exactly one must be high. The three-low clause sum is at most $120/273<20/41$, whereas clause closure requires $\sum_{t\in S}u_t\ge20/41$. The chosen witnesses satisfy each unary ceiling equality uniformly. For $G(7,4)$ their ceilings are respectively $1$ and $2$, and for $G(17,15)$ respectively $3$ and $5$. Their unary $L(8,1)$ gates close by the sufficient interval criterion. A consistent opposite-literal pair sums to $29/70$, which lies strictly inside the sufficient $L(3,2)$ interval. Any satisfied three-literal clause has multiplier sum between $58/105$ and $29/35$, inside the sufficient $L(2,1)$ interval. All inequalities hold after replacing every coefficient by either perturbation endpoint, proving uniform validity.
\end{proof}

\begin{corollary}[Robust NP-hardness in the specified CG model]\label{cor:cg-robust}
For the restricted no-presolve, one-aggregated-CG-cut root-gap or excess-tree-size model, the 3-SAT zero-loss reduction remains correct under all relative perturbations of at most $1/40$. With the usual bounded-occurrence gap-3SAT source and a lower bound of $1/18$ on the \emph{remaining gap fraction per failed gate}, the construction also has a constant NO-instance empirical gap. There are $N_{\mathrm{train}}=7T+M$ gadgets (including the additional unary gate). By Theorem~\ref{thm:smoothed-robust-transfer}, this gives a smoothed hardness consequence under its explicit computational-model and perturbation-simulation assumptions.
\end{corollary}
\begin{proof}
Completeness follows from Proposition~\ref{prop:cg-robust-soundness}. Soundness follows because any zero-loss parameter supplies valid lower/upper unary labels, consistent complement choices, and no false clause. For a bounded-occurrence gap source, assign each bad variable to a failed unary or complement gate. Each such variable can invalidate at most $B_0$ clauses. The defect-charge lemma in the toolkit bounds the number of unclosed instances below by a constant fraction of $M$. A failed $G$-gadget has ceiling coefficients differing by at least one, with the larger coefficient at most $18$, so its normalized residual gap is at least $1/18$. A failed $L$-gadget is either entirely unchanged (remaining gap one), or has positive rounded right-hand side and an integer rounded left coefficient bounded by the constant $a|S|(1+\varepsilon)\le25$, giving a positive constant remaining fraction. For the actual three $L$ types one can use the sharper $1/18$ uniformly. Thus the average failed-gate loss is at least a positive absolute constant.
\end{proof}

\begin{remark}[Scope of coefficient robustness]
This theorem is for independent perturbations of the row data while keeping the row incidence pattern, nonnegativity, integer-variable domain, and root-cut generation rule unchanged. It does not imply robustness to adding arbitrary new nonzero constraints, changing the variable domain, numerical cut rounding, presolve, or arbitrary commercial MILP heuristics. This theorem does not itself cover scalar batch cuts. A separate, robust \emph{additive-noise} scalar batch-CG construction is proved in Theorem~\ref{thm:scabs:overview}. Uniform \emph{multiplicative} robustness of the scalar arithmetic-frequency hardness construction is not proved. Binary coverage ERM instead has the relative-smoothed algorithm of Theorem~\ref{thm:cg-relative-tract}.
\end{remark}

\subsection{Matching under multiplicative weight perturbations}
\begin{proposition}[Perturbation-invariant binary clause probes]\label{prop:matching-smoothed}
Suppose the parameter-coded matching policy compares only fixed scores and history and is independent of the numeric vertex weights, every weight stays positive after perturbation, and the probe construction retains a fixed offline matching that matches the probe and all prefix arrivals. Then the unmatched-probe loss for a clause $C$ remains
\[
f_{\theta}(x_C;w)=\frac{w_{\mathrm{probe}}}{W_{\mathrm{pre}}+w_{\mathrm{probe}}}\,\mathbf1\{C(B(\theta))\text{ is false}\},
\]
for normalized regret, with the normalization total $W_{\mathrm{pre}}+w_{\mathrm{probe}}$ fixed by the instance. Hence all zero-loss and clause-soundness identities survive relative perturbations of magnitude $<1$. If the nominal normalized failed-probe loss equals $1/2$, a uniform perturbation radius $\rho<1$ leaves it at least $(1-\rho)/(2(1+\rho))$ under the stated normalization. The signal for shattering and the constant-gap reduction are correspondingly robust.
\end{proposition}
\begin{proof}The policy's choices, matched-vertex identities, and the formula for a failed probe are unchanged. Only the positive cost of failing to match the probe and the normalization vary. The lower bound follows from the worst-case numerator $1-\rho$ and total weight at most $2(1+\rho)$, or more sharply from the exact ratio.\end{proof}

\subsection{A box-robust B\&B parity-trap realization}
Let $n=T+1$ and let $k_1,\dots,k_n$ be positive odd integers, with $K=\max_j k_j$. Introduce binary selectors $x_j\in\{0,1\}$, binary variables $y_{j,a}\in\{0,1\}$ for $a\in[k_j]$, and continuous $t_h\in[-2,2]$. Consider the MILP
\begin{equation}\label{bbrobust:eq:milp}
\begin{aligned}
\max\quad &\sum_{h=1}^n t_h-\tfrac1{10}\sum_{j=1}^n x_j,\\
\text{s.t.}\quad&t_h\le 2x_h &&(h\in[n]),\\
&t_h+3x_j\le 2 &&(1\le j\le h\le n),\\
&2\sum_{a=1}^{k_j}y_{j,a}=k_jx_j &&(j\in[n]),\\
&y_{j,a}\le2x_j &&(j\in[n],\ a\in[k_j]).
\end{aligned}
\end{equation}
Independent \emph{entrywise} multiplicative smoothing replaces each designated nonzero coefficient or RHS entry $v$ of the objective and these displayed constraints by $v(1+\eta)$, with $|\eta|\le\varepsilon$. Entries may have distinct perturbations. The binary declarations, variable boxes, equality/inequality types, row supports and indexing are structural and are not perturbed. Thus zeros remain zero. We call the perturbed MILP $Q^\eta(k_1,\dots,k_n)$.

The exact LP-based solver has no presolve, additional cuts or primal heuristic. It begins without an incumbent, creates both children at each branching, and branches on the least-indexed fractional $x$ when any selector is fractional, otherwise on any fractional $y$. The node-selection rule is the history-dependent score-mixture rule of Appendix~\ref{app:bb-erm-detail}. The first $T$ continuing stages encode the binary decisions $b_j$ with dyadic comparisons of the scalar $\theta$, and a final rule inspects the recorded word and the sizes $k_j$ (identifiable from the \emph{unperturbed row supports}) to decide whether to explore a designated probe trap. Nodes explored inside an activated trap are exhausted before returning to the continuing spine. Previously deferred traps have lower scores than the continuing node, and once the all-zero incumbent is found the usual LP bound test prunes them.

\begin{theorem}[Uniform box robustness]\label{bbrobust:thm:main}
Let
\[
\boxed{\quad 0\le\varepsilon\le \frac1{1000(n+K)}.\quad}
\]
For \emph{every} perturbation of every designated nonzero entry in the displayed box and every parameter $\theta\in[0,1]$, the solver on $Q^\eta(k_1,\dots,k_n)$ has exactly the same stage decisions, visited parity subtrees, and generated-node count as in the idealized unperturbed reduction. In particular, if $k_j(C)\in\{1,3,5\}$ encodes the signs of a clause $C$ for $j\le T$, the last block has size $K=k_*$, and $b=B(\theta)$ is the induced binary word, then
\begin{equation}\label{bbrobust:eq:identity}
\Tree_\theta(Q_C^\eta)=2T+3+\sum_{j=1}^T h(k_j(C))b_j+h(k_*)\,\mathbf 1\{C(b)=\mathrm{false}\},
\end{equation}
where
\[
h(2r+1)=2\left(\binom{2r+2}{r+1}-1\right).
\]
The equality holds \emph{pointwise} for all perturbations and all $\theta$, including dyadic tie values under the prescribed tie rule.
\end{theorem}

\begin{proposition}[Box-robust computational consequences]\label{bbrobust:prop:hard}
Choose $k_*=O(\log(mT))$ as in the unsmoothed $m$-clause B\&B reduction so that $h(k_*)$ dominates the sum of the $O(mT)$ prefix costs. Then zero-loss/threshold exact ERM and the established constant-additive normalized-gap formulation of the same specialized controller remain NP-hard for \emph{every} perturbation in a box of radius
\[
\varepsilon=\Omega\!\left(\frac1{T+\log(mT)}\right).
\]
In the standard bounded-relative-noise hardness-transfer model, a polynomial-time smoothed ERM algorithm with polynomial dependence on the noise density would imply randomized polynomial time for the corresponding NP-hard promise problem, assuming efficient finite-bit simulation of the perturbation experiment. This does not give a hardness theorem for conventional B\&B branching-variable score mixtures.
\end{proposition}

\subsubsection{The uniform spine lemma}
A continuing-spine node at stage $j$ has $x_1=\cdots=x_{j-1}=0$. For any relaxed selector vector $x\in[0,1]^{n-j+1}$, the parity block can be made feasible by setting
\[
y_{h,a}=x_h\frac{k_h(1+\delta_h)}{2\sum_{b=1}^{k_h}(1+\gamma_{h,b})},\qquad h\ge j,
\]
with the actual independent coefficient perturbations $\delta_h,\gamma_{h,b}$. Since $\varepsilon\le1/100$, this is between $0.49x_h$ and $0.51x_h$, safely below the perturbed $y_{h,a}\le 2x_h$ cap and below one. Earlier fixed-zero blocks have $y=0$. Thus the projection of the LP feasible region onto the free $x$-coordinates is exactly their entire box.

Let $\Phi_{j,\eta}(x)$ be the optimal LP objective after optimizing over $t$ and $y$ with these $x$ fixed. In the unperturbed case put $x^*=(2/5,\dots,2/5)$ and define $f(x)=\min\{2x,2-3x\}-x/10$. The own-coordinate inequalities imply
\[
\Phi_{j,0}(x)\le\sum_{h=j}^n f(x_h),\qquad \Phi_{j,0}(x^*)=\frac{19}{25}(n-j+1).
\]
The latter equality holds because all relevant current-stage upper bounds equal $4/5$ when every free $x_h=2/5$, and earlier fixed-zero variables contribute zero. The piecewise-linear function $f$ has left slope $19/10$ and right slope $-31/10$. Consequently,
\begin{equation}\label{bbrobust:eq:sharp}
\Phi_{j,0}(x^*)-\Phi_{j,0}(x)\ge \frac{19}{10}\sum_{h=j}^n\big|x_h-2/5\big|.
\end{equation}

\begin{lemma}[Uniform LP-selector stability]\label{bbrobust:lem:spine}
At every continuing-spine node and under the perturbation radius in Theorem~\ref{bbrobust:thm:main}, every optimal LP solution satisfies
\begin{equation}\label{bbrobust:eq:closeness}
\sum_{h=j}^n\left|x_h-\frac25\right|\le \frac{400}{19}n\varepsilon<\frac1{40}.
\end{equation}
In particular, each free selector lies strictly inside $(0,1)$, the branching rule picks $x_j$, and the continuing-spine LP bound is positive until the final integral all-zero node.
\end{lemma}
\begin{proof}
After dividing each perturbed inequality by its positive $t_h$ coefficient, the bound $t_h\le 2x_h$ becomes $t_h\le a_hx_h$, with $|a_h-2|\le5\varepsilon$. The bound $t_h+3x_\ell\le2$ becomes $t_h\le b_{h\ell}-c_{h\ell}x_\ell$, with $|b_{h\ell}-2|\le5\varepsilon$ and $|c_{h\ell}-3|\le7\varepsilon$. Hence, for every fixed $x\in[0,1]^{n-j+1}$, the maximizing $t_h$ (the minimum of the resulting affine upper bounds and 2) changes by at most $12\varepsilon$. The lower bound $t_h\ge-2$ never binds. Every candidate upper bound is at least $-1-12\varepsilon>-2$. Perturbing the objective coefficients introduces at most $3n\varepsilon$ additional error. We obtain the uniform estimate
\[
|\Phi_{j,\eta}(x)-\Phi_{j,0}(x)|\le20n\varepsilon.
\]
Let $\widehat x$ maximize $\Phi_{j,\eta}$. Comparing the two objectives at $\widehat x$ and $x^*$ and using \eqref{bbrobust:eq:sharp} gives
\[
\frac{19}{10}\|\widehat x-x^*\|_1\le40n\varepsilon,
\]
which is \eqref{bbrobust:eq:closeness}. Since $n\varepsilon\le10^{-3}$, the right side is at most $0.022<1/40$. Also $\Phi_{j,\eta}(x^*)\ge (19/25)(n-j+1)-20n\varepsilon\ge0.74>0$.
\end{proof}

\subsubsection{All-node parity stability and strict trap bounds}
\begin{lemma}[A trapped selector forces later selectors to zero]\label{bbrobust:lem:trap}
At a child of the continuing spine with $x_j=1$ and $x_1=\cdots=x_{j-1}=0$, and at every descendant that fixes only parity variables of block $j$, the LP upper bound is at most
\[
-(n-j+1)(1-7\varepsilon)<-\frac9{10}.
\]
Whenever the LP is feasible, every optimal LP solution has $x_h=0$ for $h>j$. Thus all other parity blocks have $y_{h,a}=0$, and the only remaining source of nonintegrality is the trapped block $j$.
\end{lemma}
\begin{proof}
For every $h\ge j$, the row $t_h+3x_j\le2$ at $x_j=1$ implies, after perturbation,
\[
t_h\le \frac{2(1+\beta_{hj})-3(1+\alpha_{hj})}{1+\xi_{hj}}
\le\frac{-1+5\varepsilon}{1+\varepsilon}\le-1+6\varepsilon<0.
\]
For $h<j$, the fixed equality $x_h=0$ makes $t_h\le0$. All $t_h$ objective coefficients are positive, and each $x_h$ objective coefficient remains negative. Consequently the LP objective is at most $-(n-j+1)(1-7\varepsilon)$.

To show the asserted optimizer structure, decrease any free later selector $x_h>0$ to zero and simultaneously reset its private parity block to zero. The trapped cap on $t_h$ is negative, so its own upper bound $t_h\le a_hx_h$ remains slack after this change. Every other inequality in which $x_h$ appears is relaxed by decreasing $x_h$. Therefore this change cannot decrease any optimal $t$-coordinate and strictly increases the objective through the negative selector coefficient. Iterating forces every $x_h=0$ for $h>j$. The inequalities $y_{h,a}\le2x_h$ then force all remaining parity blocks to zero.
\end{proof}

\begin{lemma}[Exact parity-tree shape after perturbation]\label{bbrobust:lem:parity}
Let the trapped block have odd size $k=2r+1$. At any partial assignment of its $y$-variables, with $a$ fixed ones and $b$ fixed zeros, its LP is feasible \emph{if and only if} $a,b\le r$, for all perturbations allowed by Theorem~\ref{bbrobust:thm:main}. Each feasible such node has a fractional $y$-coordinate. Hence the trap explores exactly
\[
h(k)=2\left(\binom{2r+2}{r+1}-1\right)
\]
nodes strictly below its root, regardless of the order in which fractional parity variables are selected.
\end{lemma}
\begin{proof}
With $x_j=1$, divide the perturbed parity equality by 2 and write
\[
\sum_{a=1}^k(1+\gamma_a)y_a=q,
\qquad q=\frac k2(1+\delta),
\]
with $|\gamma_a|,|\delta|\le\varepsilon$. The perturbed constraints $y_a\le2x_j$ are redundant because their right-hand sides exceed 1. With $a$ fixed ones and $b$ fixed zeros, feasibility is equivalent to
\[
S_a\le q\le S_a+R,\quad
S_a=\sum_{\text{ones}}(1+\gamma_i),\quad
R=\sum_{\text{unfixed}}(1+\gamma_i).
\]
If $a,b\le r$, both inequalities have slack at least $\frac12-\frac32 k\varepsilon>\frac14$. If $a\ge r+1$, the first inequality is violated by at least $\frac12-\frac32 k\varepsilon>\frac14$. If $b\ge r+1$, the second is violated by the same amount. In particular, no binary assignment to all $k$ variables can satisfy the equality, so every feasible node has a fractional $y$ variable.

Every feasible internal node branches into two children and the children remain feasible exactly until their count of fixed zeros or ones first reaches $r+1$. By the standard ballot/hockey-stick identity the number of leaves, including infeasible ones, is $\binom{2r+2}{r+1}$. A full binary tree with $L$ leaves has $2L-1$ nodes. Subtract its root to obtain the asserted formula. The count is independent of which fractional variable is selected because feasibility depends only on $(a,b)$.
\end{proof}

\subsubsection{From LP invariants to robust smoothed hardness}
\begin{proof}[Proof of Theorem~\ref{bbrobust:thm:main}]
At each stage $j$, Lemma~\ref{bbrobust:lem:spine} guarantees that the least-indexed fractional selector is $x_j$, creating the continuing child $x_j=0$ and the trap $x_j=1$. Lemmas~\ref{bbrobust:lem:trap} and~\ref{bbrobust:lem:parity} guarantee that an activated trap is completely explored with precisely $h(k_j)$ descendants and contributes no integer feasible solution. When its subtree is exhausted, the scorer returns to the continuing child. If instead the continuing child is processed first, the trap is deferred. The controller's score hierarchy keeps it dormant until after the final continuing stage.

The structural supports of the parity equalities are unchanged by the multiplicative perturbations. The policy therefore reads the same block cardinalities $k_j\in\{1,3,5\}$ and evaluates the same source clause, and its arithmetic comparison of $\theta$ against the dyadic midpoint is independent of the perturbed MILP coefficients. Hence the first $T$ stage decisions and the final clause probe coincide with the unperturbed code $b=B(\theta)$.

At the end of the continuing spine, all $x$ and $y$ equal zero and the root LP on this terminal node has $t=0$, objective 0, and integral integer variables. It yields the first feasible incumbent, of value zero. Every deferred trap has a strictly negative LP bound (Lemma~\ref{bbrobust:lem:trap}), so it is immediately pruned without exploring descendants. There are $2n+1=2T+3$ generated spine/trap-root nodes, plus exactly $h(k_j)$ descendants of each trap explored before continuation. This proves \eqref{bbrobust:eq:identity} for all $\theta$ and all perturbations in the box.
\end{proof}
\begin{proof}[Proof of Proposition~\ref{bbrobust:prop:hard}]
The coefficient encoding of a clause is unaltered because the supports and block cardinalities are fixed. The exact count identity \eqref{bbrobust:eq:identity} is therefore identical for all perturbations. Apply the original 3-SAT and gap-3-SAT reductions to this unchanged loss, choosing $k_*=O(\log(mT))$ so $h(k_*)$ dominates the bounded prefix contribution. The resulting MILPs have $O(T+\log(mT))$ variables and polynomially many rows and entries. The noise radius in the theorem is $\Omega(1/(T+\log(mT)))$. The abstract smoothed-hardness transfer applies with inverse-polynomial noise density, subject to its randomized computation/finite-bit simulation assumptions.
\end{proof}

\subsection{Smoothed minimum-cut configuration for graph SSL}\label{sec:gaussian-smoothed}
Fix a \emph{finite} feasible cut family $\mathcal C\subseteq 2^E$ (for semi-supervised labeling it consists of cuts consistent with the prescribed labels), and independent distances $D_e$ supported on $[a,a+B]$ with densities at most $\kappa$. For $s>0$, set
\[
W_C(s)=\sum_{e\in C}e^{-sD_e},\qquad C_s\in\arg\min_{C\in\mathcal C}W_C(s),
\]
with deterministic tie breaking. Write $N=|E|$. We count the parameter values at which $C_s$ changes, not the running time of an algorithm enumerating them.

\begin{theorem}[Whole-line expected transition bound for independent exponential edge weights]\label{thm:gaussian-allbandwidth}
Under the preceding independent bounded-density edge model, the expected number of changes in the optimal cut on $s\in(0,\infty)$ satisfies
\[
\boxed{\mathbb E K\le\kappa B N\left(1+N\log N+\frac{N}{e}\right)=O(\kappa BN^2\log(N+1)).}
\]
For Gaussian graph weights $e^{-d_e^2/(2\sigma^2)}$ with independent primitive distances $d_e\in[1,2]$ of densities at most $\kappa_d$, squaring gives $B=3$ and density at most $\kappa_d/2$. Thus $\mathbb E K=O(\kappa_d n^4\log n)$ on graphs with $N=O(n^2)$ edges, over \emph{all} bandwidths $\sigma>0$.
\end{theorem}
\begin{proof}[Proof outline]
Fix $e$ and all other edge lengths. For each $s$, the best cut containing $e$ has cost $e^{-sD_e}+A_e(s)$, whereas the best cut excluding $e$ has cost $B_e(s)$, both with $A_e,B_e$ independent of $D_e$. Hence there is at most one pivotal threshold $D_e=\tau_e(s)$. At an actual cut transition, charge the tie to an edge of \emph{minimum distance in the symmetric difference} of two competing cuts. On the charged portions of its threshold curve, all other uncancelled distances exceed $\tau_e$, giving the derivative estimates
\[
|\tau'_e(s)|\le
\begin{cases}
NB^2,&0<s<1/(NB),\\
NB/s,&1/(NB)\le s\le1/B,\\
N/(es^2),&s\ge1/B.
\end{cases}
\]
Integrating yields $B(1+N\log N+N/e)$ total charged variation for each edge. The area formula, edge-density bound and pivotal charging prove the expectation.
\end{proof}

\begin{remark}[Computational and statistical interpretation]
The theorem holds for \emph{independent edge distances}, not general Euclidean embeddings with dependent pairwise distances. It counts optimal-cut regions, but does not supply a polynomial-bit parametric min-cut enumerator for transcendental weights. For independent smoothed samples with piecewise-constant bounded labeling loss, the expected growth function is at most $1+O(m\mathbb E K)$ on $m$ examples. Symmetrization gives distribution-specific uniform-convergence rates on the order of $\sqrt{(\log(m(1+\mathbb E K))+\log(1/\delta))/m}$. This is \emph{not} a reduction of the worst-case pseudodimension of the original class.
\end{remark}

\subsection{Full proof of the independent-edge exponential minimum-cut bound}
\begin{proof}[Proof of Theorem~\ref{thm:gaussian-allbandwidth}]
Let $N=|E|\ge2$. For every $e\in E$, condition on the other independent distances. At fixed $s>0$, define
\[
A_e(s)=\min_{C\in\mathcal C:e\in C}\sum_{f\in C\setminus\{e\}}e^{-sD_f},\qquad
B_e(s)=\min_{C\in\mathcal C:e\notin C}\sum_{f\in C}e^{-sD_f}.
\]
If either subfamily is empty, $e$ cannot be pivotal. Otherwise, a tie between the two subfamilies occurs only if
\[
e^{-sD_e}=H_e(s):=B_e(s)-A_e(s)>0.
\]
Since $D_e\mapsto e^{-sD_e}$ is strictly decreasing, at most one distance value $D_e=\tau_e(s):=-\log H_e(s)/s$ is pivotal at any fixed $s$. The minimizing cuts in the two subfamilies are independent of $D_e$. They range over finite collections of exponential polynomials and so, after identifying identical analytic functions and deleting isolated ties, the threshold is piecewise real analytic and rectifiable on every compact parameter interval where $0<H_e(s)<\infty$.

Suppose $s$ belongs to the portion of an $e$-pivotal curve \emph{charged to $e$}, meaning that $e$ is a minimum-distance member of the symmetric difference of the two competing minimizers. Denote their exclusive edge sets by $P$ (from the cut without $e$) and $Q$ (from the cut with $e$, excluding $e$). At $D_e=\tau_e(s)$, set $\delta_f=D_f-\tau_e(s)$ for $f\in P\cup Q$. Then
\[
0\le\delta_f\le B,\qquad
1=\sum_{f\in P}e^{-s\delta_f}-\sum_{f\in Q}e^{-s\delta_f}.
\]
Writing $h(t)=\sum_{f\in P}e^{-t\delta_f}-\sum_{f\in Q}e^{-t\delta_f}-1$, differentiation of $e^{-s\tau_e(s)}=B_e(s)-A_e(s)$ (with fixed other edge lengths) gives the exact identity
\[
\boxed{s\tau'_e(s)=-h'(s)=\sum_{f\in P}\delta_fe^{-s\delta_f}-\sum_{f\in Q}\delta_fe^{-s\delta_f}.}
\]
For $s\ge1/B$, $t e^{-st}\le1/(es)$ for $t\ge0$, hence $|\tau'_e(s)|\le N/(es^2)$. For $1/(NB)\le s\le1/B$, use $t e^{-st}\le B$ to obtain $|\tau'_e(s)|\le NB/s$.

For $s<1/(NB)$, use the inequality $|e^{-s\delta_f}-1|\le sB$. Since $h(s)=0$ and $h(0)=|P|-|Q|-1$ is an integer, $|h(0)|<1$, so $h(0)=0$. Rolle's theorem gives a $\xi\in(0,s)$ with $h'(\xi)=0$. Since $|h''(t)|\le NB^2$ for every $t\in[0,s]$, we obtain $|h'(s)|\le NB^2s$ and therefore $|\tau'_e(s)|\le NB^2$. The three ranges integrate to
\[
\int_{\text{charged part of the $e$-curve}}|\tau'_e(s)|ds
\le B+NB\log N+\frac{NB}{e}.
\]
Note that we integrate the \emph{charged portions} only. The derivative bound need not hold at pivotal ties where a different edge is shortest. Because there are finitely many pairs of competing cuts, the parameter axis can be stratified into finitely many analytic cells, making this use of the area formula legitimate.

At each genuine optimal-cut transition, select two distinct minimizing cuts on opposite sides of the transition and choose an edge $e$ of minimum $D_e$ in their symmetric difference. The conditional $e$-pivotal curve passes through the realized value $D_e$, and this transition is counted on a charged portion of that curve. This genericity claim follows directly from the edge-incidence
structure. For three cuts $C_1,C_2,C_3$ with distinct incidence
vectors, the two difference vectors
$\mathbf1_{C_1}-\mathbf1_{C_2}$ and
$\mathbf1_{C_1}-\mathbf1_{C_3}$ are linearly independent. A line
contains at most two vertices of the Boolean cube. Hence there exist
two edges $e,f$ such that the corresponding $2\times2$ submatrix
has nonzero determinant. The Jacobian of their two cut-cost
differences with respect to $(D_e,D_f)$ is that incidence submatrix
multiplied by a positive diagonal matrix with entries
$s e^{-sD_e},s e^{-sD_f}$. It has rank two for every $s>0$.
Fixing the other distances, the three-way-tie set in
$(s,D_e,D_f)$ is therefore a one-dimensional regular submanifold.
Its projection to $(D_e,D_f)$ has two-dimensional Lebesgue measure
zero. Independence and absolute continuity give probability zero,
after a finite union over cut triples. Cuts of identical edge-incidence
cost are identified in advance. Consequently, the coarea formula and the density upper bound give
\[
\mathbb EK\le\sum_{e\in E}\kappa\int_{\text{charged part}}|\tau'_e(s)|ds
\le\kappa BN\left(1+N\log N+\frac Ne\right),
\]
proving the theorem. For $d_e\in[1,2]$ with density $\le\kappa_d$, the transformed variable $D_e=d_e^2\in[1,4]$ has density $f_{D_e}(t)=f_{d_e}(\sqrt t)/(2\sqrt t)\le\kappa_d/2$, yielding the Gaussian specialization.\qedhere
\end{proof}

\begin{remark}[Critical distinction from computational enumeration]
The preceding proof counts \emph{global optima} of a finite min-cut family. It does not give a polynomial-time method to find all their transition parameters. Indeed, even when the number of optimal-cut changes is small, a routine that isolates roots of all pairwise exponential-polynomial cut costs could examine exponentially many nonoptimal cuts. Closing the output-sensitive \emph{algorithmic} step requires a separate parametric min-cut algorithm or an appropriate optimization/root oracle, with bit-complexity analysis.
\end{remark}

\subsection{Exact rational-function optimization for ridge validation}\label{app:ridge-poly}
\begin{proposition}[Polynomial-time scalar ridge squared-validation ERM]\label{prop:ridge-P}
For rational training and validation datasets of polynomial total bit length, with a rational scalar ridge parameter $\lambda\ge0$, exact global minimization of squared validation loss over a rational compact interval is polynomial-time solvable in the classical bit model when the output may be represented as a real algebraic number with an isolating interval. The same holds for finitely many independent training-validation tasks sharing the scalar $\lambda$.
\end{proposition}
\begin{proof}
For task $i$, the ridge fit is $(G_i+\lambda I)^{-1}c_i$, where $G_i,c_i$ have rational entries. Each coordinate is a ratio of polynomials of degree at most the feature dimension, with coefficients of polynomial bit length by determinant bounds. The squared validation loss is therefore rational in $\lambda$, and the sum across tasks has numerator and denominator of degree polynomial in the total input dimension and polynomial-bit integer coefficients after clearing rational denominators. On $\lambda\ge0$, the denominator is nonzero when the Gram matrices have positive ridge shifts. Handle $\lambda=0$ separately if a training design is singular. The stationary points are the real roots of $P'Q-PQ'$, an explicitly computable univariate integer polynomial of polynomial degree and bit complexity. Standard deterministic univariate real-root isolation and algebraic-number sign comparison find the smallest value among stationary points and interval endpoints in polynomial bit time.\end{proof}

\subsection{Distribution-specific capacity from expected region counts}
\begin{proposition}[Annealed growth from smoothed region counts]\label{prop:annealed-growth}
Let each independently sampled training instance be drawn from a distribution including its smoothing randomness, and suppose its loss as a function of a scalar parameter is bounded in $[0,1]$ and piecewise constant on at most $K(x)+1$ open cells, with finitely many exceptional breakpoint values. If $\mathbb E K(x)\le H$, then the expected number of distinct loss vectors on $m$ independent samples satisfies
\[
\mathbb E\,\Pi(S)\le1+2mH,
\]
where the factor $2$ accounts for values attained only at
breakpoints. Standard symmetrization and Massart's finite-class lemma
then yield an expected supremum deviation
$O(\sqrt{\log(2+mH)/m})$. A high-probability distribution-dependent
bound adds $O(\sqrt{\log(1/\delta)/m})$.
\end{proposition}
\begin{proof}
For a fixed sample, the union of at most $\sum_i K(x_i)$ breakpoint sets splits the scalar parameter line into at most $1+2\sum_iK(x_i)$ cells and singleton points. The loss vector is fixed on each cell, so $\Pi(S)\le1+2\sum_iK(x_i)$. Take expectations. Apply symmetrization to a sample of size $2m$ and condition on it. The empirical Rademacher complexity is bounded by the square root of the log of the number of distinct bounded loss vectors, using Massart's lemma and Jensen's inequality. A bounded-differences inequality gives the high-probability statement.\end{proof}

\begin{remark}
This statement concerns a specific smoothed \emph{distribution}. It does not assert that the unsmoothed function class has a smaller worst-case pseudodimension, and it does not apply to losses varying continuously within a fixed output region without additional discretization or smooth-loss covering estimates.
\end{remark}

\subsection{Application-specific smoothing status}\label{app:smoothed-families}
\begin{center}\small
\begin{tabularx}{\linewidth}{@{}p{.29\linewidth}X@{}}\toprule
Family & Smoothing result and assumptions \\\midrule
Online matching with weight-independent code & Zero-loss and gap certificates are robust to positive weight perturbations. Matching-loss normalization must be specified.\\
Single aggregated CG with shared multipliers & Robust zero-loss/gap reduction after the extra unary rounding gate, under bounded coefficient perturbations and the exact CG/no-presolve model. See Appendix~\ref{app:smoothed-certificates}.\\
History-dependent B\&B clause-reading selector & An explicit inverse-polynomial box-robust theorem with all-node LP bounds and exact parity counts, applicable only to the clause-reading node-selection controller.\\
Power-mean linkage & Theorem~\ref{thm:linkage:smoothed} bounds expected hierarchy transitions by $O(\kappa n^5\sqrt{\log n})$ under independently sampled primitive distances in $[1,2]$. Exact ERM follows in the all-real root-isolation oracle model (Corollary~\ref{cor:linkage:smoothed-erm}).\\
Gaussian min-cut semi-supervised labels & Appendix~\ref{sec:gaussian-smoothed} establishes expected output-region complexity with independent edge smoothing. Parametric optimization and Turing bit complexity remain open.\\
Lasso & Theorem~\ref{thm:lasso:kac-rice} bounds expected Gaussian path length, complementing Li--Singer~\cite{LiSinger2018}. Expected Turing bit complexity of exact breakpoint enumeration remains open.\\
One-variable batch CG & Weak NP-hardness under fixed-unit additive noise. Pseudo-polynomial root-gap ERM and expected-polynomial \emph{binary coverage} ERM under feasible independent relative smoothing (Theorem~\ref{thm:cg-relative-tract}).\\
Scalar ridge squared-validation ERM & Polynomial-time univariate rational-function optimization with rational data. No smoothing is needed (Appendix~\ref{app:smoothed-certificates}).\\
Conventional branching mixtures and fixed-dimensional aggregate CG with binary-encoded coefficients of unbounded magnitude & Smoothed exact ERM remains open under these more general settings.\\\bottomrule
\end{tabularx}
\end{center}\normalsize

\section{Models, statistical constructions, and local compilers}
\subsection{Model and result summary}\label{app:inventory}
\begin{center}\small
\begin{tabularx}{\linewidth}{@{}p{.24\linewidth}p{.28\linewidth}X@{}}\toprule
Family & Tuned parameter and statistical information & ERM status and smoothing assumptions\\\midrule
Online matching & One adaptive history-dependent mixture, with $\fat_{1/4}\ge T$ & Unary probes are polynomial. Ternary clause probes give zero-loss NP-completeness\\
One aggregated CG cut & $m$ shared multipliers, with $m\le\Pdim\le O(m\log m)$ on gadgets & Zero-loss NP-complete, with a robust relative-noise variant using explicit gate calibration\\
Scalar batch CG & One affine tagged parameter, at most three separately rounded cuts & NP-hard zero-loss with additive coefficient-perturbation hardness in its stated model\\
Scalar Lasso validation & One positive penalty with standard squared validation & Zero-loss NP-complete on a compact rational interval. Gaussian-smoothed expected path count\\
History-dependent B\&B & One scalar score mixture with an explicit clause-reading state & Hardness and box robustness in a deliberately expressive no-presolve solver\\
Ridge / supplied-tree pruning & One scalar parameter with logarithmic capacity in stated classes & Polynomial exact ridge rational-function ERM and enumeration for supplied-tree parameters\\
Power-mean linkage and ordinary strong branching & Linkage root-split hardness and expected-polynomial full-dendrogram transitions. Standard-score branching remains model-specific & Linkage oracle-model ERM is established (Corollary~\ref{cor:linkage:smoothed-erm}). Polynomial expected Turing bit complexity and standard-score branching ERM remain open\\\bottomrule
\end{tabularx}
\end{center}
\subsection{Expanded statistical and algorithmic lower-bound constructions}\label{app:previous}
\subsubsection{Definitions and the robust observation lemma}
For $\F\subseteq\mathbb R^\X$, a set $x_1,\ldots,x_d$ is $\gamma$-fat-shattered if there exist thresholds $r_i$ such that, for every $b\in\bits^d$, some $f_b\in\F$ satisfies
\[
 b_i=1\Rightarrow f_b(x_i)\ge r_i+\gamma,\qquad b_i=0\Rightarrow f_b(x_i)\le r_i-\gamma.
\]
Pseudodimension is the analogous strict-sign notion without a fixed positive margin.

\begin{lemma}[Robust observation / interference, restated]\label{prior:lem:decoder}
The robust-interference criterion of Lemma~\ref{lem:robust} applies to
any legal family of polynomial-bit witnesses, with the same margin
$\frac12\min_i(\Delta_i-\omega_i-2\varepsilon_i)$.
In particular, exact bit observations give
$\fat_{\min_i\Delta_i/2}\ge d$, and additive cross-bit terms
$h_i=\sum_{j\ne i}w_{ij}b_j$ satisfy the criterion whenever
$\Delta_i>\sum_{j\ne i}|w_{ij}|+2\varepsilon_i$.
\end{lemma}

\begin{lemma}[Approximate performance-preserving transfer]\label{prior:lem:transfer}
Suppose a source class $\F$ has a $\gamma$-fat-shattering certificate of dimension $d$. If a target class $\mathcal G$ has parameter and instance embeddings such that, on the finite certificate witnesses,
\[
|g_{\iota(\theta)}(\tau(x_i))-(a_i f_\theta(x_i)+c_i)|\le\varepsilon,
\qquad a_i>0,
\]
then $\fat_{a_{\min}\gamma-\varepsilon}(\mathcal G)\ge d$ whenever $a_{\min}\gamma>\varepsilon$.
\end{lemma}
\begin{proof}
Replace each threshold $r_i$ by $a_i r_i+c_i$ and subtract the simulation error.
\end{proof}

\subsection{Locally verifiable encoding compilers}
\subsubsection{Direct products, switches, and interpolation}
\begin{lemma}[Direct product]\label{prior:lem:product}
If $\Theta$ contains $\Theta_1\times\cdots\times\Theta_p$ and the losses on disjoint instance families $\X_j$ depend only on $\theta_j$, with $\gamma_j$-fat-shattering witnesses of sizes $d_j$, then $\fat_{\min_j\gamma_j}(\F)\ge\sum_j d_j$.
\end{lemma}
\begin{proof}Choose the parameter coordinates independently for a requested labeling.\end{proof}

\begin{lemma}[Signed-switch synthesis]\label{prior:lem:switch}
Suppose, for any rational threshold $\tau$ in an interval, there are legal upward and downward gadgets with outputs $\one\{\theta>\tau\}$ and $\one\{\theta<\tau\}$. Assume any $k$ such gadgets can be composed into one valid instance with performance $c+\Delta\sum_{j=1}^k g_j(\theta)+e(\theta)$, $|e|\le\varepsilon_k$, and common $\Delta>0$. Then for $K=2^d$, using at most $K-1$ gadgets per evaluation instance,
\[
\fat_{\Delta/2-\varepsilon_{K-1}}(\F)\ge d,
\]
provided the displayed margin is positive.
\end{lemma}
\begin{proof}
Choose ordered witnesses $\theta_0<\cdots<\theta_{K-1}$. For each bit position, insert an upward switch between witnesses where that bit changes $0\to1$, and a downward switch where it changes $1\to0$. The sum equals a constant plus the bit at every witness. Approximation reduces the gap by at most $2\varepsilon_{K-1}$.
\end{proof}

\begin{lemma}[Effective comparison-basis interpolation]\label{prior:lem:interpolation}
Let $\phi_1,\ldots,\phi_K$ be real functions, and let $V=(\phi_j(\theta_s))_{s,j=1}^K$ be nonsingular at $K$ selected parameters. Suppose every integral coefficient vector $C\in\mathbb Z^K$ can be realized by a legal two-action comparison whose score difference has the same sign as $\sum_j C_j\phi_j(\theta)$ at these parameters. Suppose opposite actions can be decoded into losses $0$ and $1$ at the witnesses. Then $\fat_{1/2}(\F)\ge\lfloor\log_2K\rfloor$. Moreover, the coefficient vectors can be chosen with
\[
\|C\|_\infty\le L\|V^{-1}\|_\infty+\tfrac12,
\qquad L=\max\{1,\lceil\|V\|_\infty\rceil\},
\]
where the matrix norm is the maximum absolute row sum.
\end{lemma}
\begin{proof}
For each of $d=\lfloor\log_2K\rfloor$ binary-digit patterns, complete a target vector $y\in\{-1,1\}^K$. Set $C$ to the componentwise nearest integer to $LV^{-1}y$. Then $\|VC-Ly\|_\infty\le\|V\|_\infty/2\le L/2$, so every prescribed sign is strict. Decode the selected action into binary loss.
\end{proof}
The legality and coefficient-cost assumptions are indispensable. Arbitrary interpolation coefficients need not be expressible by distances, LP coefficients, or valid classifications.

\subsubsection{Self-simulation of adaptive decisions}
Let a deterministic algorithm have a controller state $q$, parameter $\theta\in[0,1]$, binary action $a$, and transition $q'=U(q,a)$ that does not read $\theta$ directly.
\begin{definition}[Renormalizable state template]\label{prior:def:self}
A state family $\mathcal Q$ has compatible intervals $J_q$ and increasing bijections $\nu_q:J_q\to[0,1]$, with a common split $\tau\in(0,1)$, such that (i) $a=\one\{\nu_q(\theta)>\tau\}$ off the boundary; (ii) the successors $q_a$ again belong to $\mathcal Q$, with $J_{q_a}=\nu_q^{-1}(B_a)$ where $B_0=[0,\tau]$, $B_1=[\tau,1]$; (iii) $\nu_{q_a}=R_a\circ\nu_q$ on $J_{q_a}$ for fixed increasing maps $R_a:B_a\to[0,1]$. Assume $q_\emptyset$ has $J=[0,1]$ and $\nu$ the identity.
\end{definition}
\begin{theorem}[Constructive self-simulation]\label{prior:thm:self}
Let $h_a=R_a^{-1}$ and assume $|h_a(u)-h_a(v)|\ge\kappa|u-v|$ for some fixed $0<\kappa<1$. At depth $T$, every $b\in\bits^T$ is realized by
\[
\theta_b=h_{b_1}\circ\cdots\circ h_{b_T}(1/2).
\]
Each witness has a stable parameter neighborhood of radius $\kappa^T/4$. If there are valid evaluation instances obeying Lemma~\ref{prior:lem:decoder} for the realized decisions, their performance class fat-shatters $T$ instances at the margin in that lemma.
\end{theorem}
\begin{proof}
By induction, $h_{b_1}$ places the normalized coordinate in child interval $B_{b_1}$, forcing action $b_1$. The successor normalized coordinate cancels $h_{b_1}$ and equals $h_{b_2}\circ\cdots\circ h_{b_T}(1/2)$. Continue. The leaf interval is the image of $[0,1]$ under the same composition. Repeated lower Lipschitz bounds imply the image of $1/2$ is at least $\kappa^T/2$ from either endpoint. Lemma~\ref{prior:lem:decoder} supplies observation and margin.
\end{proof}
\begin{remark}
Theorem~\ref{prior:thm:self} controls a \emph{single evolving state}. It does not imply that a search procedure which retains multiple live subproblems has the same property. The history-dependent clause-reading construction in Appendix~\ref{app:bb-erm-detail} tracks both deferred frontier nodes and incumbent information.
\end{remark}

\section{Online matching with exact regret probes and structured ERM}

\subsection{An exact separation in online bipartite matching}\label{sec:matching}
\subsubsection{The natural optimization model}
We study \emph{online-arrival-weighted} bipartite matching. Offline vertices are known. Online vertices arrive with their neighborhoods and positive weights. The algorithm irrevocably matches each arrival to at most one free neighbor. Its reward is the sum of weights of matched online vertices, and its loss is normalized regret $f_\theta(x)=1-\ALG_\theta(x)/\OPT(x)$, where $\OPT(x)>0$ is the weight of the maximum offline matching.

For each $T$, construct a family of online instances with offline vertices $\{A_1,B_1,\ldots,A_T,B_T\}$. Their first $T$ arrivals are $u_1,\ldots,u_T$, with
\[
N(u_i)=\{A_i,B_i\},\qquad w(u_i)=1/T.
\]
The final arrival has weight one and a neighborhood specified by the evaluation instance. For any nonempty probe neighborhood containing at most one vertex from each pair, the offline optimal reward is exactly two. Route all prefix vertices to avoid one neighbor of the probe and match that probe.

\subsubsection{Two fixed scores and one tunable weight}
The online policy maintains a dyadic interval $[L,R]$, initialized to $[0,1]$. At arrival $u_j$, put $m_j=(L+R)/2$ and define the scores of $A_j,B_j$ as the convex combination of two \emph{fixed} history-dependent scoring rules:
\[
\begin{array}{c|cc|c}
& s_0&s_1&(1-\theta)s_0+\theta s_1\\\hline
A_j&0&1&\theta\\
B_j&m_j&m_j&m_j
\end{array}
\]
The policy takes the available neighbor with larger score, resolving equality in favor of $B_j$, and updates $[L,R]$ to its upper half if it takes $A_j$, and to its lower half otherwise. The final probe is matched to any free neighbor if one exists. The two scoring rules depend on the maintained state but not directly on $\theta$.

For $b\in\bits^T$ define
\[
\theta_b=\sum_{j=1}^T b_j2^{-j}+2^{-(T+1)}.
\label{eq:theta}
\]
This lies strictly inside the dyadic interval corresponding to $b$, and the comparison $\theta-m_j$ recovers successive digits. Therefore
\[
u_j\text{ matches }A_j\iff b_j=1;\qquad u_j\text{ matches }B_j\iff b_j=0.\label{eq:bits}
\]
This proves a full $T$-bit code using an $O(T)$-bit rational parameter and an $O(T)$-space deterministic online algorithm.

\subsubsection{Literal and clause probes}
Associate variable $x_j$ with bit $b_j$. A positive literal $x_j$ is true precisely when $B_j$ is free at the end of the prefix. A negative literal $\neg x_j$ is true precisely when $A_j$ is free. For a clause $C$ on distinct variables, construct the final probe with one adjacent offline vertex for each literal:
\[
x_j\in C\Rightarrow B_j\in N(v_C),\qquad
\neg x_j\in C\Rightarrow A_j\in N(v_C).
\label{eq:clausegraph}
\]
Then $v_C$ has a free neighbor if and only if at least one literal in $C$ is satisfied by $b$. The policy earns one unit from the prefix and one unit from $v_C$ precisely when $C$ is true. Since the offline optimum is two,
\[
\boxed{f_{\theta_b}(x_C)=\frac12\one\{C(b)\text{ is false}\}.}\label{eq:clauseloss}
\]
The identity also holds for every $\theta$, replacing $b$ with the policy's actual decision code $B(\theta)$. No extra evaluation oracle or nonstandard objective is used.

\begin{theorem}[Joint statistical and ERM hardness]\label{thm:matching}
For every $T$, there is an efficiently described one-parameter policy family for online-arrival-weighted online matching and an instance domain $\X_T$ with at most $3T+1$ vertices and $2T+3$ edges per instance, such that:
\begin{align*}
\fat_{1/4}(\F_T)&=\Pdim(\F_T)=T;\\
\text{width-one probe ERM}&\in\mathsf P;\\
\text{width-two exact ERM}&\text{ is NP-hard};\\
\text{width-three zero-loss ERM}&\text{ is NP-complete}.
\end{align*}
In particular, ERM under normalized matching regret has no finite multiplicative approximation unless $\mathsf P=\mathsf{NP}$, even with a single tunable real parameter and polynomial-time loss evaluation.
\end{theorem}
\begin{proof}
Equation~\eqref{eq:clauseloss} and Theorem~\ref{thm:compiler} yield the computational claims. The rational witnesses~\eqref{eq:theta} give polynomial NP certificates. For every $i$, take the one-literal clause $\neg x_i$, which is false if and only if $b_i=1$. The $T$ corresponding instances have loss $b_i/2$ and are $1/4$-fat-shattered at threshold $1/4$. Conversely, every parameter induces a $T$-bit prefix code, and all probe losses depend only on that code. The class thus contains at most $2^T$ distinct functions on $\X_T$, implying $\Pdim\le T$. For $T$ prefix arrivals and one final probe the graph contains $2T$ offline and $T+1$ online vertices. There are $2T+3$ edges at most.
\end{proof}

\begin{corollary}[Constant additive hardness]\label{cor:gap}
There is an absolute constant $\eta>0$ such that, unless $\mathsf P=\mathsf{NP}$, no polynomial-time algorithm can distinguish empirical regret $\OPT_{\ERM}=0$ from $\OPT_{\ERM}\ge\eta$ for this matching policy and three-literal probe instances. Consequently polynomial-time additive $\eta/3$-approximate ERM is impossible unless $\mathsf P=\mathsf{NP}$.
\end{corollary}
\begin{proof}
Apply the standard constant-gap version of \textsc{3-SAT}. It is NP-hard to distinguish satisfiable formulas from formulas in which every assignment violates at least an $\epsilon_0$ fraction of clauses, for some universal $\epsilon_0>0$ \cite{AroraSafra1998,Hastad2001}. Equation~\eqref{eq:clauseloss} gives empirical regret at least $\epsilon_0/2$ in the no case. Set $\eta=\epsilon_0/2$.
\end{proof}

\paragraph{Why this is not a hardness result for Ranking.}
The two fixed scoring rules deliberately maintain a dyadic interval in their shared history and use it to amplify one scalar coefficient into $T$ bits. The instance and regret are standard online matching, but the policy class is more expressive than familiar memoryless priority or random Ranking policies. The theorem establishes a concrete computational barrier for configurable \emph{adaptive} scores, not for every natural matching heuristic.

\subsection{Tractability without enumerating parameter regions}
Here a tree decomposition of the interaction graph lets us optimize
small overlapping groups of decisions without enumerating every full
codeword. This is the familiar dynamic-programming principle behind
algorithms parameterized by treewidth \cite{CyganEtAl2015}.
\label{sec:tw}
\subsubsection{Constraint width, not just statistical dimension}
The same construction also gives an exact algorithmic characterization in terms of the structure of the probes. Given probe instances $x_{C_1},\ldots,x_{C_m}$ with optional nonnegative rational weights $w_1,\ldots,w_m$, the ERM objective is
\[
\widehat L(\theta)=\frac1{2\sum_j w_j}\sum_{j=1}^m w_j\one\{C_j(B(\theta))\text{ false}\}.
\]
This is a weighted Boolean constraint-satisfaction instance. Its \emph{primal graph} has one vertex per parameter-code bit and an edge between any two bits appearing together in a probe.

\begin{theorem}[Bounded-treewidth ERM]\label{thm:tw}
If the primal graph of the probe family has a supplied tree decomposition of width $w$, exact ERM and an optimal rational configuration parameter can be computed in
\[
\boxed{2^{O(w)}\poly(T,m,L)}
\]
time, where $L$ is the bit length of probe weights. The algorithm need not enumerate the $2^T$ parameter intervals. If a decomposition is not supplied, standard fixed-parameter treewidth algorithms add a factor $2^{O(w)}\poly(T)$.
\end{theorem}
\begin{proof}
Use a standard dynamic program on a nice tree decomposition, recording the minimum total weight of violated clauses conditional on the Boolean assignment to the variables in each bag. Each clause's full scope forms a clique of the primal graph, hence is contained in some bag. Charge its weight exactly once. The tables have $2^{w+1}$ entries per bag, with polynomial-time updates. Reconstruct an optimal codeword $b$, then output its dyadic rational witness $\theta_b$ from~\eqref{eq:theta}. This realizes precisely the minimum empirical loss by~\eqref{eq:clauseloss}.
\end{proof}

\begin{corollary}[Separation from region enumeration]
There are datasets with $T$ independently evaluated single-literal probes for which the algorithm has $2^T$ distinct reachable prefix histories, yet exact ERM is computable in $O(T+m)$ time (up to arithmetic costs).
\end{corollary}
\begin{proof}
The probes' primal graph has treewidth zero. The objective is a sum of independent unary terms. Choose each bit minimizing its local contribution and encode the result in $\theta_b$. All $2^T$ histories remain reachable as the parameter varies, but none need be enumerated.
\end{proof}

\subsubsection{Statistical learnability does not imply efficient optimization}
Because $|\F_T|\le2^T$ on $\X_T$, Hoeffding's inequality and a union bound give simultaneous uniform convergence with
\[
m=O\left(\frac{T+\log(1/\delta)}{\epsilon^2}\right)
\]
independent training instances for loss in $[0,1]$. Conversely, the $T$ singleton probes, under their uniform distribution, yield the standard $\Omega(T/\epsilon^2)$ distribution-free uniform-convergence lower bound at small constant accuracy scales. Thus the statistical sample requirement is $\Theta(T/\epsilon^2)$ up to confidence factors, while the computational ERM problem can be NP-hard even on a dataset of $O(T)$ sparse probes.

\begin{table}[t]\centering\small
\caption{Statistical and computational complexity of one configurable online-matching policy. $T$ is the number of prefix decisions and $w$ is the interaction-graph treewidth.}\label{tab:trichotomy}
\begin{tabularx}{\linewidth}{@{}p{.26\linewidth}X p{.28\linewidth}@{}}\toprule
Property & What is measured & Result \\\midrule
Statistical & Fat-shattering of normalized matching regret & $\fat_{1/4}=\Pdim=T$ \\
ERM, one-literal probes & Independent tests of decision bits & $\poly(T,m)$ time \\
ERM, two-literal probes & Optimization of pairwise interactions & NP-hard (\textsc{Max-Cut}) \\
ERM, three-literal probes & Zero-loss feasibility & NP-complete (\textsc{3-SAT}) \\
ERM, bounded treewidth & Factor-graph structured instances & $2^{O(w)}\poly(T,m)$ time \\
Generalization & Uniform deviation for a $2^T$-member class & $O((T+\log(1/\delta))/\epsilon^2)$ samples \\\bottomrule
\end{tabularx}
\end{table}

\subsection{Additional calculations for the matching compiler}

\subsubsection{A literal-probe trace for a three-variable clause}
Let $C=x_1\vee\neg x_2\vee x_3$ and $b=(0,1,0,\ldots)$. The prefix matches $u_1$ to $B_1$, $u_2$ to $A_2$, and $u_3$ to $B_3$. The probe is adjacent to $B_1,A_2,B_3$, none of which is free, so it remains unmatched and incurs regret $1/2$. For $b=(1,1,0,\ldots)$, $B_1$ is free and the probe is matched, giving regret zero. This shows the clause is tested by availability, not by any extrinsic clause-loss oracle.

\subsubsection{More precise complexity formulation}
For fixed $T$, the loss of any rational $\theta$ is polynomial-time computable by $T$ scalar comparisons and one matching decision. The problem \textsc{Matching-Probe-ERM-Zero} takes $T$ and a list of width-at-most-three probe graphs and asks whether there is $\theta\in[0,1]$ with zero empirical regret. The reduction from \textsc{3-SAT} is polynomial time. Membership in NP follows because any satisfying trace has the explicit dyadic rational witness in equation~\eqref{eq:theta}. The witness's denominator is $2^{T+1}$, so its bit length is linear in $T$. Restricting the parameter search to these rational witnesses leaves the optimum unchanged.

\subsubsection{A sample-complexity check}
For the uniform distribution on the $T$ singleton negative-literal probes, the empirical process is exactly
\[
\sup_\theta|\widehat{\E}f_\theta-\E f_\theta|=\frac14\sum_{j=1}^T\left|\frac{N_j}{m}-\frac1T\right|,
\]
where $N_j$ is the occurrence count of probe $j$. For $Z\sim\mathrm{Bin}(m,1/T)$, variance
$v=m(1/T)(1-1/T)$, the fourth central moment obeys
$\mathbb E(Z-m/T)^4\le3v^2+v$. If $v\ge1$, Paley--Zygmund
applied to $(Z-m/T)^2$ gives
$\mathbb E|Z-m/T|\ge c\sqrt v$ for a universal $c>0$.
For $v<1$, the event $Z=0$ gives
$\mathbb E|Z-m/T|\ge(m/T)(1-1/T)^m$ when $m\lesssim T$.
Summing $T$ identical marginal expectations and dividing by $4m$
gives $\mathbb E\sup_\theta|\widehat{\mathbb E}f_\theta-
\mathbb Ef_\theta|\ge c'\min\{1,\sqrt{T/m}\}$.
The reverse inequality follows from Jensen/Cauchy--Schwarz and the
trivial upper bound $1/2$. Replacing one sampled probe changes the
empirical total-variation expression by at most $1/(2m)$. McDiarmid's
inequality yields
\[
\Pr\{|D-\mathbb ED|>t\}\le2\exp(-8mt^2),
\quad D=\sup_\theta|\widehat{\mathbb E}f_\theta-\mathbb Ef_\theta|.
\]
In particular, for $m\ge T$ the lower bound holds with probability
$1-2e^{-c''T}$ after reducing the universal constant, and for $m\le T/2$
a constant deviation also follows deterministically from the
unobserved probe coordinates. This is a uniform-convergence statement,
not a minimax-ERM lower bound.

\subsection{An adaptive combinatorial optimization application: online matching}
Consider online-arrival-weighted online bipartite matching. The $T$ prefix arrivals $u_j$ have distinct offline neighbor pairs $\{A_j,B_j\}$ and weights $1/T$. Evaluation instance $x_i$ has one additional final arrival $v_i$ of weight 1, adjacent only to $A_i$. An online algorithm scores $A_j$ by $\theta$ and $B_j$ by the midpoint of a stored interval $[L_j,R_j]$, updates that interval to the chosen half, and greedily matches the arrival. These scores arise as the mixture $(1-\theta)s_0+\theta s_1$ with $(s_0(A_j),s_1(A_j))=(0,1)$ and $(s_0(B_j),s_1(B_j))=(m_j,m_j)$.

\begin{theorem}[Constant-margin matching lower bound]\label{prior:thm:matching}
For the policy above, normalized regret $f_\theta(x)=1-\ALG_\theta(x)/\OPT(x)$ satisfies
\[
\fat_{1/4}(\F_T)\ge T
\]
using one real parameter and $O(T)$ vertices and edges per evaluation instance.
\end{theorem}
\begin{proof}
The score comparison $\theta-m_j$ and interval update are exactly the dyadic template, so witness $\theta_b=\sum_j b_j2^{-j}+2^{-(T+1)}$ sends $u_j$ to $A_j$ iff $b_j=1$. All prefix arrivals are matched, earning weight 1. Offline matching sends each prefix arrival to $B_j$ and the final probe to $A_i$, giving $\OPT(x_i)=2$. The algorithm matches the probe iff $b_i=0$, giving $\ALG_{\theta_b}(x_i)=2-b_i$. Thus $f_{\theta_b}(x_i)=b_i/2$, which yields margin $1/4$ about threshold $1/4$.
\end{proof}

\section{Unconditional B\&B ERM hardness with clause probes}\label{app:bb-erm-detail}
A final clause-decoding trap turns a sequence of recorded frontier decisions into empirical tree-size penalties. The fixed node-selection policy may inspect its search history and the encoded clause, while tree size is measured by LP-based branch-and-bound.
\subsection{A B\&B frontier gadget}
Fix $T$ and an odd sequence $k_1,\ldots,k_{T+1}$. Write $n=T+1$. Create binary selectors $x_1,\ldots,x_n$, binary blocks $y_{j,1},\ldots,y_{j,k_j}$, continuous $t_1,\ldots,t_n\in[0,1]$, and continuous auxiliary variables $z_1,\ldots,z_{n-1}\in[0,1]$. Consider the MILP
\begin{align}
\max\quad &\sum_{j=1}^n t_j,\label{bbext:eq:obj}\\
\text{subject to}\quad
&t_j\le 2x_j,\quad t_j\le 2(1-x_j) &&(j\in[n]),\label{bbext:eq:tent}\\
&2\sum_{a=1}^{k_j}y_{j,a}=k_jx_j,\quad y_{j,a}\le x_j &&(j\in[n],\ a\in[k_j]),\label{bbext:eq:parity}\\
&z_j\ge x_j,\quad x_{j+1}+z_j\le1 &&(j\in[n-1]),\label{bbext:eq:excl}\\
&z_j\ge z_{j-1} &&(j=2,\ldots,n-1),\label{bbext:eq:zexcl}\\
&x_j,y_{j,a}\in\{0,1\},\quad 0\le t_j,z_j\le1.\nonumber
\end{align}
Only the $x$ and $y$ variables are integer-constrained. The algorithm has no presolve, cuts, primal heuristics, or integrality-based inference beyond ordinary LP relaxation checks. It creates both children whenever it branches on an integer variable, and prunes a node if its LP is infeasible, integral, or its LP upper bound is no greater than a known incumbent.

\begin{lemma}[Continuing spine and parity trap]\label{bbext:lem:lp}
At the node $x_1=\cdots=x_{j-1}=0$ the LP optimal value is $n-j+1$, and every LP optimum has $x_j=\cdots=x_n=1/2$. Branching on $x_j$ yields: (i) a continuing child $C_j$ with $x_j=0$ and LP bound $n-j$, and (ii) a trap child $H_j$ with $x_j=1$, LP bound $0$, and no integer-feasible completion. In the trap, all selector variables other than $x_j$ are fixed to zero along the continuing spine or implied to be zero by \eqref{bbext:eq:excl}--\eqref{bbext:eq:zexcl}. Its remaining integer infeasibility is exactly the odd-parity equality $2\sum_{a=1}^{k_j}y_{j,a}=k_j$.
\end{lemma}
\begin{proof}
Every free term $t_h$ is at most one, with equality only if $x_h=1/2$. Setting all free selectors and all relevant $z_h$ equal to $1/2$, and setting $y_{h,a}=1/4$ for each free block, attains $n-j+1$. In the continuing child, $t_j=0$, and the same argument holds for all later selectors. In the trap, $x_j=1$ forces $t_j=0$, and the exclusion constraints force every later $x_h=0$ and hence every other $t_h=0$. Earlier selectors already equal zero. The LP of the trap is feasible with $y_{j,a}=1/2$, while integer feasibility would require an integer equal to $k_j/2$, which is impossible because $k_j$ is odd.
\end{proof}

\begin{lemma}[Exact trap count]\label{bbext:lem:trap}
For an odd trap block of size $k=2r+1$, if B\&B branches on an unfixed fractional $y$-variable at every feasible nonintegral node, the number of nodes \emph{strictly below} the trap root is
\begin{equation}\label{bbext:eq:h}
h(k)=2\left(\binom{2r+2}{r+1}-1\right).
\end{equation}
In particular, $h(1)=2$, $h(3)=10$, $h(5)=38$, and $h(k)\ge 2^{(k+1)/2}$.
\end{lemma}
\begin{proof}
A partial assignment with $a$ ones and $b$ zeros remains LP feasible exactly when $a,b\le r$. Otherwise the sum $r+1/2$ is unattainable within the remaining box. No feasible node is integral because the target sum is half-integer. Each terminal root-to-leaf branch therefore ends when the number of zeros or ones first reaches $r+1$. By the hockey-stick identity, there are $\binom{2r+1}{r}$ terminal leaves for each of these two reasons, giving $\binom{2r+2}{r+1}$ leaves in total. A full binary tree with $L$ leaves has $2L-1$ nodes, so its number of nodes strictly below the root is $2L-2$, as claimed. The lower bound follows already because all depth-$r$ partial assignments are LP-feasible.
\end{proof}

\subsection{A fixed polynomial-time node-selection policy}
For the first $T$ stages, the node-selection controller keeps a dyadic interval $[L,R]$, initially $[0,1]$, and records whether each trap was explored before the next continuing node. Let $m_j=(L+R)/2$. Two fixed node-scoring functions $s_0,s_1$ assign scores $(s_0,s_1)(H_j)=(0,1)$ to the current trap and $(s_0,s_1)(C_j)=(m_j,m_j)$ to its continuation. The actual score is the convex mixture $s_\theta=(1-\theta)s_0+\theta s_1$. Previously deferred traps have score $-3$ in both rules.

If $\theta>m_j$, the trap is processed first and its descendants receive score $2$ until the trap is exhausted. Otherwise the continuing node is processed first and the trap is marked deferred. After the stage, the controller updates $[L,R]$ to its right or left half respectively. These priorities are set by fixed, polynomial-time rules that may inspect the earlier exploration history and the MILP coefficients. They do not directly read $\theta$. Ties are resolved toward the continuing node.

The variable-branching rule is fixed. Branch on the least-indexed fractional selector if one exists, and otherwise on a fractional parity variable. The controller's trap-exhaustion mode ensures that the selected parity subtree is completely explored before continuing. These specifications define a deterministic LP-based B\&B algorithm.

For a binary word $b=(b_1,\ldots,b_T)$ the parameter witness
\begin{equation}\label{bbext:eq:wit}
\theta_b=\sum_{j=1}^Tb_j2^{-j}+2^{-(T+1)}
\end{equation}
causes trap $H_j$ to be explored before continuation exactly when $b_j=1$. Conversely, every $\theta$ induces some such binary word, and every word is realized by \eqref{bbext:eq:wit}.

\subsection{Clause probes and exact cost representation}
Let $C$ be a Boolean clause of at most three distinct literals over variables $b_1,\ldots,b_T$. Encode the clause in the \emph{sizes of the parity blocks} as follows:
\begin{equation}\label{bbext:eq:marker}
k_j(C)=\begin{cases}
3,&\text{if positive literal }b_j\text{ occurs in }C,\\
5,&\text{if negative literal }\neg b_j\text{ occurs in }C,\\
1,&\text{otherwise.}
\end{cases}
\qquad j\in[T].
\end{equation}
Set $k_{T+1}=k_*=2r+1$, where $r$ will be chosen later. Denote $H=h(k_*)$. The final trap $H_{T+1}$ is a \emph{probe}.

At the final stage, the fixed scoring rules inspect the at-most-three marked block sizes, compute whether the clause $C$ is false on the stored decision bits $b$, and assign the probe trap higher priority than its continuation \emph{if and only if} $C(b)$ is false. Both scoring rules assign the same final priorities. No additional $\theta$ dependence is introduced. This takes $O(T)$ time, or $O(1)$ after indexing the marked blocks. After processing the final continuing child $x_{T+1}=0$, the algorithm finds a feasible solution with value zero, which is optimal, and prunes all previously deferred traps, all of whose stored LP upper bounds are zero.

\begin{lemma}[Clause-cost normal form]\label{bbext:lem:normal}
For every clause $C$ and every parameter $\theta$ encoding the word $b\in\{0,1\}^T$, the B\&B algorithm generates exactly
\begin{equation}\label{bbext:eq:normal}
\Tree_\theta(Q_C)
= (2T+3)+\sum_{j=1}^T h(k_j(C))b_j+H\,\mathbf 1\{C(b)=\mathrm{false}\}.
\end{equation}
Every node in the search can be processed in polynomial time, and if $r=O(\log(T+m))$, the total number of generated nodes is polynomial in $T+m$.
\end{lemma}
\begin{proof}
There are $T+1$ spine branchings, which generate $2(T+1)+1=2T+3$ nodes, counting the root and every trap root. For $j\le T$, the trap root contributes precisely $h(k_j(C))$ extra descendants if $b_j=1$, and none if deferred. At the final stage the probe root contributes exactly $H$ extra descendants if $C(b)$ is false, and none if the final continuation is processed first. In both cases the final continuing child yields a feasible incumbent of value zero and prunes every deferred trap by its LP bound. Lemmas~\ref{bbext:lem:lp} and~\ref{bbext:lem:trap} give the costs. Since $k_j\in\{1,3,5\}$ for $j\le T$, the prefix adds at most $38T$ nodes. The probe contributes $H=2\binom{2r+2}{r+1}-2\le 2^{O(r)}$ nodes.
\end{proof}

\subsection{Hardness and tractability theorems}
\begin{lemma}[Unary ERM is polynomial-time]\label{bbext:lem:unary}
If every training instance encodes a clause of size at most one, exact ERM is solvable in $\operatorname{poly}(T,m)$ time.
\end{lemma}
\begin{proof}
For a unary clause, $\mathbf1\{C(b)=\mathrm{false}\}$ is either $b_i$ or $1-b_i$, and the empty clause (if permitted) gives a constant. By the exact cost identity \eqref{bbext:eq:normal}, the entire empirical objective is a constant plus $\sum_j d_jb_j$. Choose each $b_j$ separately according to the sign of $d_j$, and output $\theta_b$ from \eqref{bbext:eq:wit}.
\end{proof}

\begin{theorem}[NP-complete ERM threshold decision]\label{bbext:thm:nphard}
For the fixed B\&B policy family above, deciding whether there exists $\theta\in[0,1]$ such that $\sum_{a=1}^m\Tree_\theta(Q_{C_a})\le K$ is NP-complete on the restricted family of clause-probe MILPs with at most three literals per clause. The claim remains true for at most two literals per clause. Each training-instance evaluation at a specified rational parameter takes polynomial time. Every MILP has $O(T+\log(mT))$ variables and constraints and polynomially bounded coefficients.
\end{theorem}
\begin{proof}
For the three-literal case, reduce from 3-SAT on $T$ variables and $m$ clauses. Choose $r=\lceil\log_2(100mT)\rceil$, which ensures $H=h(2r+1)>38mT$. For any word $b$, Lemma~\ref{bbext:lem:normal} gives
\begin{equation}\label{bbext:eq:sum}
\sum_{a=1}^m\Tree_{\theta_b}(Q_{C_a})
=m(2T+3)+\sum_{j=1}^T c_jb_j+H\,\operatorname{Unsat}_\Phi(b),
\end{equation}
where $c_j=\sum_a h(k_j(C_a))\le38m$ and $\operatorname{Unsat}_\Phi(b)$ is the number of false clauses. If $\Phi$ is satisfiable, some $b$ has cost at most $m(2T+3)+38mT$. If $\Phi$ is unsatisfiable, every $b$ has cost greater than this threshold because $H>38mT$. Since every $b$ has the parameter witness \eqref{bbext:eq:wit}, and every parameter encodes a word, this establishes equivalence. Membership in NP follows from the $T+1$-bit rational witness and the explicit polynomial-time formula \eqref{bbext:eq:sum}.

For two-literal clauses, reduce from \textsc{MaxCut}. For each edge $\{i,j\}$, add the two clauses $(b_i\lor b_j)$ and $(\neg b_i\lor\neg b_j)$. An edge contributes one violated clause if its endpoints have equal bits, and no violated clauses otherwise. Choosing $H$ greater than the maximum possible variation of the linear prefix term makes the aggregate objective lexicographically minimize the number of uncut edges, from which a threshold for a specified target cut size follows. The reduction and NP membership are polynomial.
\end{proof}

\begin{theorem}[Constant-additive inapproximability]\label{bbext:thm:gap}
There is a universal constant $\delta>0$ for which, unless $P=NP$, no polynomial-time algorithm can always return a parameter whose average \emph{normalized} B\&B tree size is within additive $\delta$ of the optimal average, even on the above three-literal probe family. Every fixed-parameter B\&B run in this reduction takes polynomial time.
\end{theorem}
\begin{proof}
Use the standard perfect-completeness gap-3-SAT consequence of the PCP theorem. There exists $\epsilon>0$ for which it is NP-hard to distinguish a satisfiable formula from one where every assignment falsifies at least $\epsilon m$ clauses. Let $B=2T+3+38T$ and choose $H\ge 8B/\epsilon$, still with $r=O(\log(T+1))$. Normalize every tree size by the common bound $M=B+H$. From \eqref{bbext:eq:normal}, if $\Phi$ is satisfiable, some parameter has average normalized cost at most $B/M\le\epsilon/8$. In the unsatisfiable case every parameter has average cost at least $\epsilon H/M\ge 7\epsilon/8$. Thus a polynomial-time additive $\delta$-approximation to ERM for any sufficiently small fixed $\delta$, e.g. $\epsilon/4$, would decide gap-3-SAT. The construction assumes no exponential-time hypothesis.
\end{proof}

\begin{theorem}[Capacity and probe-size dependence]\label{bbext:thm:fat}
Fix $T$ and allow any odd-parity probe size $k_*$ with $H=h(k_*)$. Include all such sizes as part of the encoded instance family. The unnormalized tree-size class has $\Pdim=T$ on the clause-probe domain. For a fixed $H\ge4(2T+3+38T)$, Let $\mathcal X_{T,r}$ be the class of encoded MILPs whose clauses have arity at most $r$, for any $r\in\{1,2,3\}$. With $M=H+2T+3+38T$, define $f_\theta(Q)=\Tree_\theta(Q)/M$. Then
\begin{equation}\label{bbext:eq:fat}
\operatorname{fat}_{1/4}(\{f_\theta:\theta\in[0,1]\}\vert_{\mathcal X_{T,r}})=\operatorname{Pdim}(\{f_\theta:\theta\in[0,1]\}\vert_{\mathcal X_{T,r}})=T.
\end{equation}
Thus increasing clause arity from one to two changes ERM from polynomial-time to NP-hard without increasing the pseudodimension. The NP-hardness reduction in Theorem~\ref{bbext:thm:nphard} chooses $H$ as a function of the sample size $m$. Its instances belong to the variable-$H$ class, not to one fixed-$H$ subclass.
\end{theorem}
\begin{proof}
For each $i\in[T]$, use the unary negative clause $C_i=(\neg b_i)$. It is false exactly when $b_i=1$. By \eqref{bbext:eq:normal}, losses for $b_i=0$ are at most $B/M\le1/5$, while losses for $b_i=1$ are at least $H/M\ge4/5$. Threshold $1/2$ separates these with margin at least $1/4$, and every assignment is realized by a parameter. For the upper bound, for any allowed $H$ every parameter produces one of only $2^T$ bit strings, and the loss on any encoded instance depends on the parameter only through that string. This proves $\Pdim\le T$ uniformly for the class allowing variable $H$. Thus the restricted loss class has at most $2^T$ functions and pseudodimension at most $T$, whence the same bound for fat-shattering dimension.
\end{proof}

\begin{corollary}[Exact ERM at bounded interaction treewidth]\label{bbext:cor:tw}
Let $\Phi$ be the collection of training clauses and let $w$ be the treewidth of its primal graph on variables $b_1,\ldots,b_T$, with two variables adjacent when they occur together in a clause. Given a width-$w$ tree decomposition, exact ERM for the corresponding MILPs is solvable in $2^{O(w)}\operatorname{poly}(T,m)$ time.
\end{corollary}
\begin{proof}
The empirical objective \eqref{bbext:eq:sum} is the sum of unary costs $c_jb_j$ and local clause penalties $H\mathbf1\{C_a(b)=\mathrm{false}\}$. Standard dynamic programming on a tree decomposition minimizes this weighted Boolean constraint objective in time $2^{O(w)}\operatorname{poly}(T,m)$. A width-$O(w)$ decomposition can alternatively be found in $2^{O(w)}\operatorname{poly}(T)$ time using Korhonen\cite{Korhonen2021}, rather than assuming that an exact decomposition is obtainable with the same dependence. Recover a minimizing word $b$, then output $\theta_b$ from \eqref{bbext:eq:wit}.
\end{proof}

\section{Hardness of ERM for one aggregated Chv\'atal--Gomory cut}\label{app:cg-single-aggregate}
This section establishes ERM hardness for a single aggregated rank-one
cut. Here a vector
$u\in[0,1]^m$ is applied to the $m$ constraints of each input and produces
\emph{exactly one aggregated CG inequality}. The row coordinate $t$ has the
same multiplier $u_t$ on every training instance. No instance-specific
tags, cut filtering, or history-dependent selector are used. Our fixed
LP-based B\&B solver applies only this root cut, disables preprocessing and
additional separation, and counts all generated nodes including the root.
This is a restricted version of the learning-to-cut configuration model in
\cite{prior:balcan2021cuts}.

\subsection{Model and two one-variable gadgets}
An input $I=(A,b,c)$ is a pure integer maximization program
$\max\{c^\top z:Az\le b,\ z\in\mathbb Z_{\ge0}^n\}$ with integral data and
finite optimal value of its LP relaxation. All our gadget instances
have structural binary domains $x\in\{0,1\}$, $y\in\{0,1\}$.
The corresponding box bounds are enforced in each LP relaxation but are
\emph{not} among the $m$ rows being aggregated by the tunable multiplier
vector. Its parameterized rank-one CG cut is
\begin{equation}\label{eq:singlecg:cut}
  \lfloor u^\top A\rfloor z\le\lfloor u^\top b\rfloor,
  \qquad u\in[0,1]^m.
\end{equation}
Because $z\ge0$ is integer, this is valid: $\lfloor u^\top A\rfloor
z\le u^\top Az\le u^\top b$, and its left-hand side is integral.
Let $I^u$ be the LP with that cut appended, and let $\Tree_u(I)$ count
B\&B nodes after adding the cut. Define its fraction of initial root
integrality gap closed for maximization by
\[
 g_u(I)=\frac{\mathrm{LP}(I)-\mathrm{LP}(I^u)}
                  {\mathrm{LP}(I)-\mathrm{IP}(I)},
 \qquad 0\le g_u(I)\le1,
\]
when $\mathrm{LP}(I)>\mathrm{IP}(I)$. We use the nonnegative empirical
losses $\frac1N\sum_I(\Tree_u(I)-1)$ and
$\frac1N\sum_I(1-g_u(I))$.

For any $a>\beta\ge1$ and nonempty $S\subsetneq[m]$, define two
programs on binary variables $(x,y)\in\{0,1\}^2$, with box bounds
structural rather than multiplier rows. Every row outside $S$ reads
$y\le0$:
\begin{align*}
 L(a,\beta;S):&\quad \max x\quad\text{with } ax\le\beta
 \text{ in each }t\in S;\\
 G(a,\beta;S):&\quad \max(-x)\quad\text{with }-ax\le-\beta
 \text{ in each }t\in S.
\end{align*}
The dummy rows force $y=0$ already in the LP relaxation, so there is one
nontrivial integer variable and a unique LP solution. Let
$\sigma=\sum_{t\in S}u_t$. All terms involving the multipliers on dummy
rows cancel after $y=0$ is substituted.

\begin{lemma}[Root behavior of the two gadgets]\label{lem:singlecg:gadgets}
For every $u\in[0,1]^m$, the program $L(a,\beta;S)$ has root LP optimum
$x_0=\beta/a$, integer optimum $x=0$, and
\[
 \Tree_u(L)=1\iff \sigma\in[1/a,1/\beta).
\]
The program $G(a,\beta;S)$ has root LP optimum $x_0=\beta/a$,
integer optimum $x=1$, and
\[
 \Tree_u(G)=1\iff \sigma>0\quad\text{and}\quad
                         \lceil a\sigma\rceil=\lceil\beta\sigma\rceil.
\]
Each gadget has precisely three generated nodes whenever the root is
not integral. In either case, $\Tree_u(I)=1$ if and only if $g_u(I)=1$.
\end{lemma}
\begin{proof}
The aggregated cut on $L$ becomes
$\lfloor a\sigma\rfloor x\le\lfloor\beta\sigma\rfloor$.
For $\sigma<1/a$ it is vacuous. For
$\sigma\in[1/a,1/\beta)$ it forces $x=0$. For larger $\sigma$
its right-hand side is positive and the optimum remains strictly between
zero and one. Thus its root is integral precisely on the asserted interval.
Otherwise branching on $x$ creates an integral $x\le0$ child and an
infeasible $x\ge1$ child. For $G$, the cut becomes
$\lceil a\sigma\rceil x\ge\lceil\beta\sigma\rceil$.
When $\sigma=0$ it is vacuous. Otherwise its new root optimum is
$\max\{\beta/a,\lceil\beta\sigma\rceil/\lceil a\sigma\rceil\}$.
This equals the integer optimum 1 exactly when the ceilings coincide.
Otherwise the two children $x\le0$ and $x\ge1$ are respectively infeasible
and integral. In both gadgets a fractional root has exactly two children.
Finally a root is integral exactly when the cut closes the full initial LP
integrality gap.
\end{proof}

\begin{lemma}[Two Boolean intervals]\label{lem:singlecg:bool}
Let $J_0=[1/8,1/7]$ and $J_1=(1/4,2/7]$.
\begin{enumerate}[label=(\roman*),leftmargin=*]
\item Both unary gadgets $L(8,1;\{t\})$ and $G(7,4;\{t\})$ have
one-node trees precisely when $u_t\in J_0\cup J_1$.
\item For $v,w\in J_0\cup J_1$, $v+w\in[1/3,1/2)$ if and only if
exactly one of $v,w$ is in $J_1$.
\item For $v,w,z\in J_0\cup J_1$, $v+w+z\in[1/2,1)$ if and only if
at least one of $v,w,z$ is in $J_1$.
\end{enumerate}
\end{lemma}
\begin{proof}
The $L(8,1)$ closure set is $[1/8,1)$. For $G(7,4)$, the two
ceilings agree precisely when no integer $k$ satisfies
$4u\le k<7u$. On $u\in[0,1]$ this gives
$(0,1/7]\cup(1/4,2/7]$ as its closure set. Intersection gives (i).
For (ii), sums of two values both in $J_0$ are at most $2/7<1/3$.
Sums of two values both in $J_1$ exceed $1/2$. Mixed sums lie in
$(3/8,3/7]\subset[1/3,1/2)$. For (iii), three $J_0$ values sum to
at most $3/7<1/2$. A sum with at least one $J_1$ value exceeds
$1/4+1/8+1/8=1/2$ and is less than $3(2/7)<1$.
\end{proof}

\subsection{SAT reduction and exact computational consequences}
\begin{theorem}[Single-aggregated-CG empirical risk minimization]
\label{thm:singlecg:hard}
Given a $3$-CNF formula $\varphi$ with $T\ge3$ variables and $M$
clauses of exactly three literals on distinct variables, we can construct
in polynomial time a training set $S_\varphi$ with $N=5T+M$ IPs,
$m=2T$ shared CG multipliers, one structurally binary variable and
one dummy integer variable fixed to zero per IP, and absolute tunable-row
coefficient values at most $8$, satisfying:
\begin{enumerate}[label=(\roman*),leftmargin=*]
\item Every input has at most three rows other than the dummy $y\le0$
rows, and $\Tree_u(I)\in\{1,3\}$ for every $u\in[0,1]^m$.
\item $\varphi$ is satisfiable if and only if a \emph{single vector} $u$
gives every instance a one-node B\&B tree, equivalently closes all
its initial LP integrality gaps.
\item For every $u$, write $F(u)$ for the number of three-node trees
and $B(\varphi)$ for the largest number of clauses containing a given
variable. There is a polynomial-time Boolean assignment $\tau_u$ that
violates at most $(B(\varphi)+1)F(u)$ clauses.
\item If $\Tree_u(I)=3$, then $g_u(I)\le5/12$.
\end{enumerate}
Consequently zero excess-tree-size and zero remaining-gap ERM are
NP-complete on this restricted family, and neither admits a finite-factor
approximation unless $\mathsf P=\mathsf{NP}$. Under bounded-occurrence
Gap-3SAT, each also admits a constant-additive NP-hardness gap. All of
these statements concern \emph{one aggregated cut per IP}.
\end{theorem}
\begin{proof}
For each Boolean variable $x_i$ allocate two parameter coordinates
$r(x_i)=2i-1$ and $r(\neg x_i)=2i$. Create the following programs:
\begin{center}\small
\begin{tabular}{@{}lcl@{}}\toprule
Input type & Number & Gadget \\\midrule
Unary range $A_\ell$ & $2T$ & $L(8,1;\{r(\ell)\})$\\
Unary periodic $B_\ell$ & $2T$ & $G(7,4;\{r(\ell)\})$\\
Variable consistency $P_i$ & $T$ & $L(3,2;\{2i-1,2i\})$\\
Clause check $Q_C$ & $M$ & $L(2,1;\{r(\ell):\ell\in C\})$\\
\bottomrule\end{tabular}
\end{center}
Lemma~\ref{lem:singlecg:gadgets} proves (i). If $\tau$ satisfies
$\varphi$, set $u_{r(\ell)}=27/100$ when $\ell$ is true under $\tau$
and $u_{r(\ell)}=2/15$ otherwise. These numbers lie in $J_1$ and
$J_0$, respectively, so Lemma~\ref{lem:singlecg:bool} makes every
root integral. Conversely, if every root is integral, the unary tests
force all coordinates into $J_0\cup J_1$, the $P_i$ tests force
exactly one literal of each variable into $J_1$, and the $Q_C$ tests
force every clause to contain a $J_1$ literal. This is a satisfying
assignment, proving (ii).

For (iii), declare $x_i$ \emph{good} if its four unary tests and the
one $P_i$ test succeed. Put $\tau_u(x_i)=1$ precisely when
$u_{2i-1}\in J_1$. Every good variable has consistent literal truth
values under this assignment. If a clause is false, either its $Q_C$
test fails or it contains a bad variable. The five tests belonging to
distinct bad variables are disjoint, so there are at most $F(u)$ bad
variables. Each is contained in at most $B(\varphi)$ clauses. Thus
at most $(B(\varphi)+1)F(u)$ clauses are false.

For (iv), one can enumerate the finitely many attainable
$(\lfloor\beta\sigma\rfloor,\lfloor a\sigma\rfloor)$ pairs
for $L(3,2)$ with $\sigma\in[0,2]$ and for $L(2,1)$ with
$\sigma\in[0,3]$. At fractional roots the maximum respective fractions
of gap closed are $1/4$ and $1/3$. For $L(8,1)$ every nonintegral
root closes zero gap. For $G(7,4)$ put
$p=\lceil4\sigma\rceil$, $q=\lceil7\sigma\rceil$.
At a nonintegral root $p<q$, and the greatest attainable $p/q$ is
$3/4$. If $p\le2$, then $q\ge p+1$. If $p=3$ then
$q\ge4$. And if $p=4$ then $\sigma>3/4$ forces $q\ge6$.
Therefore
\[
  g_u(G)\le\frac{3/4-4/7}{1-4/7}=\frac5{12}.
\]
This completes (iv).

For the NP membership claim, each instance on the reduced family has
a cut and a root LP value computable by exact rational arithmetic.
The witnesses for satisfiable formulas use only the two rational values
$2/15$ and $27/100$. More generally, on arbitrary rational input IPs
the root-cut outcomes depend only on the integer floors of finitely many
linear forms in $u$. Each feasible floor pattern defines a rational
polyhedron with some strict facets and has a rational point of polynomial
binary length, found by introducing a common strict-inequality slack.
This gives a polynomial-size certificate for the one-node decision
problem with a fixed polynomial-time deterministic LP oracle.

For a finite-factor approximation, the optimum of either nonnegative
loss is zero precisely on yes formulas. A finite approximation factor
must output zero on those formulas. Evaluating a returned parameter is
polynomial on these gadgets and would decide 3-SAT.
For constant-additive hardness use bounded-occurrence Gap-3SAT
\cite{ref:Feige1998}, with constants $\epsilon_0>0,B_0$ such that
every no-instance violates at least $\epsilon_0M$ clauses for every
assignment and every variable appears in at most $B_0$ clauses.
Part (iii) gives $F(u)\ge\epsilon_0M/(B_0+1)$. Remove unused variables,
so $T\le3M$ and $N=5T+M\le16M$. Hence
\[
 \frac1N\sum_I(\Tree_u(I)-1)
 \ge\frac{\epsilon_0}{8(B_0+1)},\qquad
 \frac1N\sum_I(1-g_u(I))
 \ge\frac{7\epsilon_0}{192(B_0+1)}.
\]
The yes-instances attain zero, so approximating either objective to
within a sufficiently small fixed additive constant is NP-hard.
\end{proof}

\subsection{Statistical complexity and a convexity--hardness boundary}
\begin{proposition}[Capacity of the restricted aggregated-CG family]
\label{prop:singlecg:pdim}
Fix $m\ge4$. Let $\mathcal X_m$ consist of $L(a,\beta;S)$ and
$G(a,\beta;S)$ with $1\le\beta<a\le8$ and
$1\le|S|\le3$. Then the tree-size loss family satisfies
\[
  m\le\Pdim\{I\mapsto\Tree_u(I):u\in[0,1]^m\}
  \le O(m\log m).
\]
\end{proposition}
\begin{proof}
For a lower bound use the $m$ instances $L(8,1;\{t\})$.
The two choices $u_t=0$ and $u_t=1/2$ respectively give tree sizes
$3$ and $1$, so threshold 2 shatters these $m$ inputs.
For the upper bound, each input depends on $u$ only through
$\sigma_I(u)=\sum_{t\in S}u_t\in[0,3]$ and changes value only where
$a\sigma_I$ or $\beta\sigma_I$ is an integer. Thus each input has
$O(1)$ critical hyperplanes in $\mathbb R^m$. For $N$ inputs,
there are $O(N)$ such hyperplanes and at most $(CN)^m$ arrangement
faces of all dimensions, for an absolute $C$. Shattering $N$ inputs
requires $2^N\le(CN)^m$, which implies $N=O(m\log m)$.
\end{proof}

\begin{proposition}[Upper-bound-only gadgets: easy feasibility, hard ERM]
\label{prop:singlecg:upperonly}
If every training input is of type $L(a,\beta;S)$, deciding whether
one multiplier vector closes every LP gap reduces to feasibility of the
weak-and-strict linear inequalities
\[
   1/a\le\sum_{t\in S}u_t<1/\beta,
   \qquad 0\le u_t\le1,
\]
and is polynomial-time solvable. Nevertheless, exact ERM for excess
B\&B node count is NP-hard for $L$ gadgets alone, and is APX-hard on
a bounded-degree family.
\end{proposition}
\begin{proof}
The zero-loss conditions come from
Lemma~\ref{lem:singlecg:gadgets}. A common slack variable resolves all
strict inequalities. For exact ERM reduce \textsc{Max-Cut} on a graph
$G=(V,E)$ of maximum degree $\Delta$. Use one multiplier per vertex and
let $W=\Delta+1$. For every vertex $v$, include $W$ copies of
$L(7,6;\{v\})$ and $W$ copies of $L(4,3;\{v\})$.
These close precisely on disjoint intervals
$K_0=[1/7,1/6)$ and $K_1=[1/4,1/3)$, respectively. For every edge
$vw$ include $L(3,2;\{v,w\})$, which closes for
$u_v+u_w\in[1/3,1/2)$.
Two multipliers chosen from $K_0\cup K_1$ close this edge gadget
exactly when one is from each interval.

Any multiplier outside $K_0\cup K_1$ closes neither unary type.
Replacing it by $3/20\in K_0$ earns $W$ unary successes while losing
at most $\Delta<W$ edge successes. Therefore some optimum has every
multiplier in $K_0\cup K_1$. At such a parameter the number of closed
gadgets equals $W|V|$ plus the number of edges cut. As each open gadget
contributes two units to the excess node count,
\[
  \min_u N\,L_{\rm TS}(u)=
  2\bigl(W|V|+|E|-\mathrm{MaxCut}(G)\bigr).
\]
NP-hardness follows. For cubic graphs $W=4$ and $|E|=3|V|/2$. Let $C^*$ be the
maximum cut size and let $L^*$ be the minimum unnormalized excess node
count. The exact identity gives
$L^*=2((11/2)|V|-C^*)$. Since $C^*\ge |E|/2=3|V|/4$,
we have $L^*\le(38/3)C^*$. Given an arbitrary multiplier vector $u$,
canonicalize each coordinate outside $K_0\cup K_1$ to $3/20$. This
cannot increase the number of failed gadgets because it gains $W=4$
unary closures while sacrificing at most three edge closures.
Interpret the resulting two intervals as a cut $C(u)$. Then
\[
 C^*-|C(u)|
   =\frac{L(u')-L^*}{2}
  \le \frac{L(u)-L^*}{2}.
\]
This is an L-reduction from cubic \textsc{Max-Cut} to minimizing the
unnormalized excess-tree-size objective, with constants $38/3$ and $1/2$.
The known APX-hardness of cubic \textsc{Max-Cut}
\cite{ref:BermanKarpinski1999} therefore transfers.
\end{proof}

\begin{remark}[Assumptions of the single-cut result]
The theorem tunes $m=2T$ coordinate multipliers and uses one aggregated
CG cut per training instance. It requires neither row tags nor
history-dependent control, unlike the scalar batch-cut reduction.
Every restricted IP has only one nontrivial integer variable, so its
best cut and its three-node fallback search are trivial. Hardness comes
from the overlapping row supports across instances. The theorem does
not settle ERM hardness for a \emph{fixed} number of multiplier
coordinates, nor does it imply hardness for commercial presolve models.
The hard sample size $N=\Theta(m)$ is within a logarithmic factor of
the restricted family's $O(m\log m)$ pseudodimension upper bound. This
alone is not a fixed-accuracy sample-complexity guarantee.
\end{remark}

\section{Lasso and regression-path hardness and smoothed counting}
\subsection{Full proof of ordinary scalar Lasso validation hardness}
\label{app:lasso-adjacent-proof}
\subsubsection{The Mairal--Yu dual replay invariant}
At recursive level $k\ge1$, write $(X^{(k)},y^{(k)})$ for the
$k$-variable triangular design from Theorem~\ref{thm:lasso-polybit}.
Use zero-based indices $0,\ldots,k-1$ and let
\[
 g^{(k)}(\lambda):=\frac{(X^{(k)})^\top
 (y^{(k)}-X^{(k)}\widehat w^{(k)}(\lambda))}{\lambda}
 \in[-1,1]^k.
\]
The KKT conditions say that $g_j=\operatorname{sgn}(\widehat w_j)$
when $\widehat w_j\ne0$. On a zero coefficient, $g_j$ may lie anywhere
in $[-1,1]$.

\begin{lemma}[Dual replay and positivity of the newest coordinate]
\label{lem:lasso-replay-adj}
For every level $k\ge1$ and $\lambda>0$,
\[
 0<g^{(k)}_{k-1}(\lambda)\le1.
\]
For $p>k$, every prefix $(g^{(p)}_0,\ldots,g^{(p)}_{k-1})$ at a given
$\lambda>0$ is either zero or equals
$\sigma g^{(k)}(\mu)$ for some $\sigma\in\{-1,+1\}$ and $\mu>0$.
\end{lemma}
\begin{proof}
At one step of the adversarial Mairal--Yu construction, write the new
column as $\alpha(2y^{(k)},1)^\top$, the new coefficient as $\omega$,
and let $\Gamma=1-2\alpha\omega$.
The new coefficient is nonnegative. If $\omega<0$, then
$\Gamma=1-2\alpha\omega>1$.  The older coefficient block minimizes
Lasso with response $\Gamma y^{(k)}$ and hence equals
$\Gamma\widehat w^{(k)}(\lambda/\Gamma)$.  With
$\widehat w=\widehat w^{(k)}(\lambda/\Gamma)$, the old KKT identity gives
\[
 (y^{(k)})^\top(\Gamma y^{(k)}-X^{(k)}\Gamma\widehat w)
 =\Gamma\bigl(\|y^{(k)}-X^{(k)}\widehat w\|_2^2
       +(\lambda/\Gamma)\|\widehat w\|_1\bigr)\ge0.
\]
The newest normalized dual coordinate is consequently
$\alpha[2(y^{(k)})^\top(\Gamma y^{(k)}-X^{(k)}\Gamma\widehat w)
    +1-\alpha\omega]/\lambda>0$, contradicting its KKT value $-1$
when $\omega<0$.  Thus $\omega\ge0$. If $\omega>0$, KKT forces the newest
dual coordinate to equal $1$. If $\omega=0$, the previous coordinates
solve the old problem at parameter $\lambda$, and KKT for the new column
gives
\[
 g^{(k+1)}_k(\lambda)
 =\frac{\alpha(2(y^{(k)})^\top
 (y^{(k)}-X^{(k)}\widehat w^{(k)}(\lambda))+1)}{\lambda}>0.
\]
Indeed, the old KKT identity implies
$y^\top(y-X\widehat w)=\|y-X\widehat w\|_2^2+
\lambda\|\widehat w\|_1\ge0$.
If $\omega>0$ and $\Gamma\ne0$, the older coefficients solve the
previous-level Lasso with response $\Gamma y^{(k)}$.
The exact rescaling relation is
$\widehat w_{\rm old}=\Gamma\widehat w^{(k)}(\lambda/|\Gamma|)$,
so the old normalized dual block equals
$\operatorname{sign}(\Gamma)g^{(k)}(\lambda/|\Gamma|)$.
If $\Gamma=0$, this old block is zero. If $\omega=0$, the same formula
holds with $\Gamma=1$.
Iterating from the full problem down to a desired prefix proves the
signed-replay alternative. The base newest dual coordinate is positive
for the one-variable instance, completing the induction.
\end{proof}

\begin{lemma}[Saturation from an adjacent zero-sum]
\label{lem:lasso-adjacent-saturation}
Let $g=g^{(p)}(\lambda)$ at $\lambda>0$, and suppose $g_0\ne0$.
For each $1\le k\le p-1$, if $g_{k-1}+g_k=0$, then
\[
 (g_{k-1},g_k)\in\{(1,-1),(-1,1)\}.
\]
\end{lemma}
\begin{proof}
By Lemma~\ref{lem:lasso-replay-adj}, the prefix through index $k$ is
$\sigma g^{(k+1)}(\mu)$ for some positive $\mu$, since it cannot be
zero when $g_0\ne0$. Its last two entries arise from the $(k+1)$-st
Mairal--Yu extension. If its newest coefficient is zero, the last
coordinate is strictly positive and the preceding coordinate is the
strictly positive newest dual coordinate of the unchanged $k$-variable
problem. Their sum cannot vanish. If the newest coefficient is
positive, the new dual coordinate equals $1$. Cancellation forces the
preceding coordinate to equal $-1$. Multiplication by the common sign
$\sigma$ gives the stated opposite saturated pair.
\end{proof}

\subsubsection{Bounded-coefficient dual clause tests}
\begin{corollary}[Sparse dual readouts for scalar Lasso validation]
\label{cor:lasso-1in3}
The ordinary scalar-Lasso zero-validation-loss problem remains NP-hard
when the instance-dependent dual-validation readouts contain at most
four nonzero entries, all in $\{-1,+1\}$. The training design is the
same polynomial-bit maximal-path matrix and the regularization domain
remains a positive compact interval.
\end{corollary}
\begin{proof}
Reduce from positive 1-in-3-SAT with three distinct variables per clause
\cite{Schaefer1978}.
For variable $j$ retain the anchor, adjacent-pair, and top-pair
validation vectors from the ordinary reduction. For each clause
$C=\{a,b,c\}$ add the validation vector
$u_C=e_{2a-1}+e_{2b-1}+e_{2c-1}+e_t$.
The KKT identity makes its ordinary squared validation residual
$-\lambda(g_{2a-1}+g_{2b-1}+g_{2c-1}+g_t)$.
The validity vectors enforce $g_t=1$ and
$g_{2j-1}=\sigma_j\in\{-1,1\}$. Thus the clause residual vanishes
exactly when $\sigma_a+\sigma_b+\sigma_c=-1$, i.e. exactly one of its
three variables has sign $+1$. Completeness follows from the
full-support sign-coverage lemma. Soundness follows from adjacent-pair
saturation, including at partial-support states and breakpoints.
Each dual clause row has four entries equal to $+1$. Its ordinary
validation feature is $Gu_C$ and target $u_C^\top c$, which may have
large rational magnitudes despite the sparse dual description.
This is not a constant additive-gap or bounded-feature claim.
\end{proof}

\subsubsection{The signed-partition reduction and verification}
Let $d_1,\ldots,d_r,T$ be the signed-partition instance, and choose
$p=2r+3$ with top index $t=p-1=2r+2$.
Take the triangular rational $(X,y)$ of
Theorem~\ref{thm:lasso-polybit}. Set $G=X^\top X$ and $c=X^\top y$.
For each $u\in\mathbb Q^p$, the validation feature row is
$(Gu)^\top$ and its response is $u^\top c$. Because
$G\widehat w=c-\lambda g$, its residual is
$-\lambda u^\top g$ at every $\lambda>0$, even on partial-support
segments and at breakpoints.

Use exactly $r+3$ validation vectors:
\[
\begin{aligned}
 u_A&=e_0-e_t,\\
 u_j&=e_{2j-1}+e_{2j}\quad(j=1,\ldots,r),\\
 u_{\rm top}&=e_{t-1}+e_t,\\
 u_P&=\sum_{j=1}^r d_j e_{2j-1}-T e_t.
\end{aligned}
\]
The ordinary squared loss is
\[
 V(\lambda)=\sum_{u\in\{u_A,u_1,\ldots,u_r,u_{\rm top},u_P\}}
 \bigl((Gu)^\top\widehat w(\lambda)-u^\top c\bigr)^2
 =\lambda^2\sum_u(u^\top g(\lambda))^2.
\]

\paragraph{Soundness, including inactive coefficients and kinks.}
If $V(\lambda)=0$ for some $\lambda>0$, the top-pair equation gives
$g_{t-1}+g_t=0$. The newest dual coordinate $g_t$ is strictly positive
by Lemma~\ref{lem:lasso-replay-adj}. At the top recursive extension,
an inactive newest coefficient would make both $g_{t-1}$ and $g_t$
positive. Hence the newest coefficient is active, so $g_t=1$ and
$g_{t-1}=-1$. The anchor equation gives $g_0=g_t=1$.
Lemma~\ref{lem:lasso-adjacent-saturation} now applies to every pair,
yielding signs $\sigma_j=g_{2j-1}\in\{-1,+1\}$ and
$g_{2j}=-\sigma_j$. The partition readout reduces to
$\sum_jd_j\sigma_j=T$. This is a signed-partition certificate.

\paragraph{Completeness.}
For a signed assignment $\sigma$, consider the full-support vector
\[
 s=(1,\sigma_1,-\sigma_1,\ldots,\sigma_r,-\sigma_r,-1,1).
\]
Mairal and Yu's antipodal path bound, saturated by the polynomial-bit
construction, means that precisely one of $s$ and $-s$ occurs. The
last/newest coordinate can be active only with positive sign, so this
representative is $s$. Every full-support segment has $g=s$, and every
validation readout vanishes when $\sum_jd_j\sigma_j=T$.

\paragraph{Positive compact domain and bit complexity.}
Let
$\lambda_{\rm lo}=a_p/[4(2p-1)^2]$ and
$\lambda_{\rm hi}=\|X^\top y\|_\infty$. By
Theorem~\ref{thm:lasso-polybit}, the positive lower endpoint lies
below every positive path kink. The upper endpoint is the all-zero
Lasso threshold. Thus all full-support sign intervals needed for
completeness meet the closed domain. All rational training/validation
entries and endpoints have encoding length polynomial in the signed
partition input size. The construction takes polynomial time and does
not enumerate path segments.

\paragraph{NP membership.}
A witness consists of a rational $\lambda$ and rational Lasso
coefficients $w$. The KKT conditions, with a guessed sign/support
pattern, form a rational system of linear equalities and weak/strict
inequalities in $(w,\lambda)$. The zero-validation condition adds
rational linear equalities in $w$. A feasible system has a feasible
rational point of polynomial binary length by standard rational linear
programming bounds. The KKT and squared-validation equalities can then
be checked exactly in polynomial time. Hence the decision problem is
in $\mathsf{NP}$ and is NP-complete by the reduction.

\paragraph{Limits of the theorem.}
The proof uses $r+3$ validation observations. It does not establish
hardness for a constant number. The loss tends to zero as
$\lambda\downarrow0$ on all encoded instances, so on the open positive
half-line its infimum is always zero. We do not claim a constant
additive-approximation gap, standard scalar AIC hardness, or a
single-parameter result for arbitrary fixed-positive Elastic Net.

\subsection{Regression reductions and path complexity}\label{app:regression-details}
\subsubsection{Parameter-path sign cube and hardness boundary}
Standard scalar Lasso squared-validation zero-loss NP-completeness follows
from Theorem~\ref{thm:lasso-ordinary-validation} and the adjacent-pair
dual argument in Appendix~\ref{app:lasso-adjacent-proof}.
Lemma~\ref{lem:lasso-antipodal} supplies full-support sign coverage up
to a global sign. Positivity of the newest dual coordinate selects
the sign representative
$(1,\sigma_1,-\sigma_1,\ldots,\sigma_r,-\sigma_r,-1,1)$ for
each assignment $\sigma$. The validation equations then require
$\sum_jd_j\sigma_j=T$. This argument does not assume that signed-partition
feasibility is invariant under negation of $\sigma$ when $T\ne0$.
The polynomial-bit construction and positive rational cutoff are given
in Theorem~\ref{thm:lasso-polybit}.

\subsubsection{Why the complexity penalty matters}
The density-penalized tuning objective is
\begin{equation}\label{eq:lasso-penalty}
\Phi(\lambda)=\bigl(r^\top\widehat w(\lambda)-t\bigr)^2+\gamma\bigl(p-\|\widehat w(\lambda)\|_0\bigr),\quad \lambda\ge\lambda_0>0.
\end{equation}
The $\ell_0$ term here \emph{rewards dense solutions}. It is not the conventional AIC/BIC penalty $\gamma\|\widehat w\|_0$, which rewards sparsity. This sign is essential to the proposed reduction. It prevents partial-support solutions from attaining zero even if their validation residual vanishes. The fixed-full-support version dispenses with this term by explicitly restricting feasible parameters. Neither objective alone establishes ordinary validation hardness. The distinct adjacent-pair reduction in Theorem~\ref{thm:lasso-ordinary-validation} does so for a compact positive scalar interval. On a full-support state, $\Phi=\lambda^2(u^\top s)^2$. In a NO instance of \textsc{Partition} and with $\lambda\ge\lambda_0$, this is at least $\lambda_0^2$, while partial support costs at least $\gamma$.

\subsubsection{Smoothed path complexity and expectation}
Mairal and Yu give exponential worst-case Lasso paths.
Li and Singer establish polynomial path complexity under smoothed input
assumptions. Given path breakpoints, Proposition~\ref{prop:regression-scan}
merges the path partitions and minimizes piecewise-quadratic empirical
validation loss. Theorem~\ref{thm:lasso:kac-rice} supplies an expected-polynomial
bound on the number of pieces under Gaussian perturbations. The remaining
problem for expected Turing-time ERM is to locate exact breakpoints
with controlled representation and arithmetic cost.

\subsubsection{Ridge, Elastic Net, and statistical generalization}
Ridge regression $\widehat w(\lambda)=(X^\top X+\lambda I)^{-1}X^\top y$ has rational dependence on the regularization parameter. Fixed-support Elastic Net is affine in its $\ell_1$ coefficient when the $\ell_2$ coefficient is fixed. For a fixed active set $J$ and signs $s$,
\[
\widehat w_J(\lambda_1,\lambda_2)=(X_J^\top X_J+\lambda_2 I)^{-1}(X_J^\top y-\lambda_1 s).
\]
This exact KKT identity does not imply an efficient path enumeration when $\lambda_2$ also varies. A ridge $\Theta(\log p)$ capacity theorem is proved in Theorem~\ref{thm:ridge-cauchy} under explicitly constructed rational tasks. Constant-margin statements requiring uniformly bounded validation features are not asserted. Positive fat-shattering margins for Lasso support-cube constructions still require their own separation hypotheses and verification.

\subsection{Expected Lasso path length from leave-one-column-out crossings}\label{app:lasso:kac-rice}
Let $A=X_{-j}$ and let $r(\lambda)=y-A\widehat w_{-j}(\lambda)$ be the Lasso residual for the design with column $j$ removed. The following argument is for a Gaussian-perturbed, full-rank design, for which ordinary genericity ensures that the path is nondegenerate almost surely.

\begin{lemma}[Entry events are bounded by independent one-column crossings]\label{lem:lasso:entry}
Write $x_j$ for column $j$ of $X$, and let $Z_{j,\pm}$ denote the number of positive solutions of $x_j^\top r(\lambda)=\pm\lambda$, counted at nonzero crossings (including one-sided endpoints). Then, almost surely,
\[K(X,y)\le1+2\sum_{j=1}^d(Z_{j,+}+Z_{j,-}).\]
\end{lemma}
\begin{proof}
If coordinate $j$ is inactive, the unique solution on the remaining coordinates equals the reduced Lasso solution, because zero is KKT-feasible for the deleted column. An entry event is therefore a solution of one of the displayed equations. Along the path from large to small $\lambda$, every exit follows a preceding entry, and each generic segment transition is an entry or exit of some coordinate. Genericity rules out simultaneous entry events and contact intervals. Thus the total number of transitions is at most twice the total number of entry events.
\end{proof}

\begin{lemma}[Gaussian Rice bound for one removed column]\label{lem:lasso:rice}
Condition on $A$ and let $\alpha=\sigma_{\min}(A)>0$, $\beta=\|A\|_{\rm op}$, and $\gamma=\mathrm{dist}(y,\mathrm{col}(A))>0$. If $x_j=\mu+\frac\sigma{\sqrt n}z$ is independent of $A$, with $\|\mu\|\le1$ and $z\sim N(0,I_n)$, then
\[
\mathbb E[Z_{j,+}+Z_{j,-}\mid A]
\le C\left[1+\frac{\sqrt n}{\sigma}\,\beta
\left(1+\frac{(2+\beta+\sigma/\sqrt n)\sqrt d}{\alpha}\right)
\left(1+\log\frac1\gamma\right)\right].
\]
\end{lemma}
\begin{proof}
For $\lambda>\beta$, the reduced Lasso is identically zero and $r(\lambda)=y$, so each sign equation has at most one crossing. On each nontrivial reduced-path segment, $r$ is affine and
$\|r'(\lambda)\|_2=\|A_S(A_S^\top A_S)^{-1}s_S\|_2\le\sqrt d/\alpha$.
Further, $\|r(\lambda)\|\ge\gamma$. If $S\ne\varnothing$, the active KKT equation gives $\lambda\le\beta\|r(\lambda)\|$. If $S=\varnothing$, $\|r\|=1$. Therefore for $0<\lambda\le\beta$,
\[\|r(\lambda)\|\ge\max\{\gamma,\lambda/\beta\}.\]
For $\psi_\pm(\lambda)=x_j^\top r(\lambda)\mp\lambda$, the conditional law of $\psi_\pm(\lambda)$ is Gaussian, with density at zero at most $C\sqrt n/(\sigma\|r\|)$. Conditioning further on $\psi_\pm(\lambda)=0$, Gaussian regression gives
\[
\mathbb E[|\psi_\pm'(\lambda)|\mid\psi_\pm(\lambda)=0,A]
\le1+(2\|\mu\|+\beta+\sigma/\sqrt n)\|r'(\lambda)\|.
\]
Indeed, the conditional mean includes the projection factor $\langle r,r'\rangle/\|r\|^2$ times $\pm\lambda-\mu^\top r$, controlled by $\lambda/\|r\|\le\beta$. The remaining Gaussian standard deviation is at most $(\sigma/\sqrt n)\|r'\|$. Apply the one-dimensional Kac--Rice formula on every affine segment (breakpoint hits have probability zero) and integrate. Finally
\[\int_0^\beta\frac{d\lambda}{\max\{\gamma,\lambda/\beta\}}
\le\beta\left(1+\log\frac1\gamma\right),\]
using $\|y\|=1$ and hence $0<\gamma\le1$.
\end{proof}

\begin{lemma}[Smoothed inverse-singular-value and projection-distance moments]\label{lem:lasso:smallball}
For $A=X^h_{-j}+G_{-j}$, $G_{-j}\in\mathbb R^{n\times(d-1)}$ with independent $N(0,\sigma^2/n)$ entries, $n\ge d$ and $\|X^h\|_{\rm op}\le1$, the variables $\alpha,\beta,\gamma$ above obey
\[
\|\alpha^{-1}\|_{L^{3/2}}\le C\frac{d\sqrt n}{\sigma},\qquad
\|\beta\|_{L^{12}}\le C,\qquad
\left\|1+\log\frac1\gamma\right\|_{L^6}
\le C\left(1+\log\frac{nd}{\sigma}\right).
\]
\end{lemma}
\begin{proof}
For a full-rank $n\times(d-1)$ matrix $A$, the negative-second-moment identity states
$\sum_i\sigma_i(A)^{-2}=\sum_{j=1}^{d-1}\mathrm{dist}(A_j,\mathrm{span}(A_{-j}))^{-2}$.
Thus $\alpha\le t$ implies that for at least one column $j$, the distance to the span of the other columns is at most $\sqrt d\,t$. Conditional on the others, that orthogonal complement has dimension at least $n-d+2\ge2$. Anderson's inequality (or the maximal density of a shifted two-dimensional isotropic Gaussian ball) bounds the probability by $Cnd t^2/\sigma^2$ per column. Union-bounding gives
\[\Pr\{\alpha\le t\}\le\min\{1,Cd^2 n t^2/\sigma^2\},\]
which yields the stated $L^{3/2}$ inverse moment by tail integration. Standard Gaussian spectral-norm concentration, $\|X^h\|\le1$, and $\sigma\le1$ yield the $L^{12}$ moment of $\beta$.

For the projection-distance claim, rotate coordinates so $y=e_n$ and write $A=\binom{B}{a^\top}$, with $B\in\mathbb R^{(n-1)\times(d-1)}$. If $s=\sigma_{\min}(B)$, minimizing $\|Bw\|^2+(1-a^\top w)^2$ gives
\[\gamma\ge \frac{s}{\sqrt{s^2+\|a\|^2}}.\]
The same negative-second-moment argument on $B$ has orthogonal-complement dimension at least $n-d+1\ge1$, so $\Pr\{s\le t\}\le\min\{1,Cd^{3/2}\sqrt n\,t/\sigma\}$. A logarithm of a random variable with this linear lower-tail bound has all fixed moments of size $O(1+\log(nd/\sigma))$. The vector $a$ has bounded fixed moments under the stated assumptions, and $\log(1/\gamma)\le\log(1+\|a\|/s)$. This yields the claimed $L^6$ bound.
\end{proof}

\begin{proof}[Proof of Theorem~\ref{thm:lasso:kac-rice}]
Apply Lemma~\ref{lem:lasso:entry} and the conditional bound of Lemma~\ref{lem:lasso:rice}. In the expectation of the right side, apply H\"older to $\alpha^{-1}$ in $L^{3/2}$, $\beta^2$ in $L^6$, and $1+\log(1/\gamma)$ in $L^6$. Lemma~\ref{lem:lasso:smallball} yields
\[\mathbb E(Z_{j,+}+Z_{j,-})
 \le C\frac{n d^{3/2}}{\sigma^2}\left(1+\log\frac{nd}{\sigma}\right).
\]
Sum over the $d$ columns. The proof uses Gaussian absolute continuity so breakpoint coincidences have probability zero,
and counts one-sided endpoints only once. The bound concerns the expected
number of linear pieces. To upgrade it to an expected-polynomial \emph{Turing} runtime for exact ERM requires output-sensitive path following with polynomial expected bit complexity for the piece endpoints. The expectation bound alone does not supply that computational statement.
\end{proof}

\section{A worst-case batch-CG construction and its precision boundary}\label{scabs:app:scalar-cg-additive-smoothing}

Given $T$ Boolean variables, let $q_i=1000^i$, $1\le i\le T$. For each occurrence of a literal on variable $i$, the corresponding integer program has one binary decision variable $x\in\{0,1\}$, a row
\begin{equation}\label{scabs:eq:row}
  -(10q_i+1)x\le -(10q_i-11),
\end{equation}
and a fixed polarity tag. A clause instance has one row of type \eqref{scabs:eq:row} for each of its one to three literals, \emph{all on the same variable} $x$. Its objective is $\min x$.

The policy is controlled by $\theta\in[0,1]$. For a positive literal its row multiplier is $u_+(\theta)=\theta/10$. For a negative literal it is $u_-(\theta)=(1-\theta)/10$. Each row is rounded separately, producing at most three rank-one CG cuts per IP.

A permitted perturbation replaces each nonzero numerical entry of each row (coefficient and right-hand side) and the objective coefficient by an independently chosen \emph{additive} number of magnitude at most $1/10$. The draws may differ even between different occurrences of the same literal. Zero entries and the binary domain are not perturbed. The results below are pointwise for \emph{every} perturbation in that box. Independent bounded-density draws, including uniform draws, are a special case.

Let $L_C(\theta)$ denote the normalized \emph{remaining} root integrality gap for a clause instance $C$, equal to $1$ before cuts and $0$ when the root LP reaches the integer optimum. Let $\operatorname{TS}_C(\theta)$ be the number of generated B\&B nodes, with a root LP check, no presolve, no additional cuts, no primal heuristic, and standard binary branching.

\begin{theorem}[Box-robust scalar CG hardness]\label{scabs:th:main}
There is a polynomial-time reduction from 3-SAT to a collection of one-variable, at-most-three-row IPs such that, for \emph{every} permitted perturbation of every numerical occurrence,
\[
\min_{\theta\in[0,1]}\sum_C L_C(\theta)=0
\quad\Longleftrightarrow\quad \text{the source formula is satisfiable}.
\]
For each $\theta$ and each clause, $\operatorname{TS}_C(\theta)\in\{1,3\}$, and $\operatorname{TS}_C(\theta)=1\iff L_C(\theta)=0$. Moreover, if $L_C(\theta)>0$, then $L_C(\theta)\ge 4/5$. Therefore exact ERM, finite-factor approximation of nonnegative remaining-gap loss, and constant-additive approximation for both normalized remaining gap and average excess tree size are NP-hard. The coefficient encoding and every canonical satisfying witness have $O(T)$ bits per entry. Each fixed-parameter root evaluation is polynomial-time on rational inputs.

For every perturbation, the associated unary-instance family also satisfies $\fat_{1/2}\ge T$ for normalized remaining-gap loss (with the usual non-strict fat-shattering convention).
\end{theorem}

\subsection{A common interval formula for one-row CG closure}\label{app:cg-one-row-interval}
\begin{lemma}[Exact strong-cut intervals]\label{lem:cg-one-row-interval}
Let $0<B<A$ and let $-Ax\le-B$, $x\in\{0,1\}$, receive a
nonnegative multiplier $u\in[0,u_{\max}]$.  Its rounded CG inequality
$\lceil Au\rceil x\ge\lceil Bu\rceil$ forces $x=1$ exactly for
\[
 u\in\bigcup_{\substack{k\in\mathbb Z_{\ge1}\\
                k<A/(A-B)}}
 \left(\frac{k-1}{B},\frac{k}{A}\right]\cap[0,u_{\max}].
\]
Equivalently, for $s=1-B/A\in(0,1)$ there is one open-left,
closed-right window for each integer $1\le k<1/s$ that intersects the multiplier domain.  There are at most
$\min\{\lceil Au_{\max}\rceil,\lceil1/s\rceil\}$ such windows.
\end{lemma}
\begin{proof}
The two positive ceilings equal $k$ if and only if
$k-1<Bu\le k$ and $k-1<Au\le k$. Since $B<A$, their intersection is
$((k-1)/B,k/A]$. It is nonempty precisely when
$(k-1)A<kB$, or $k<A/(A-B)=1/s$. Any intersecting window has $k-1<Bu_{\max}<Au_{\max}$,
so the number of possible integers $k$ is at most $\lceil Au_{\max}\rceil$. The case $u=0$ yields the vacuous cut $0\ge0$.
\end{proof}

\subsection{A stable unary CG predicate}
Write the perturbed raw row as $-A x\le-B$, with
\[
 A=10(q_i+c),\quad B=10(q_i-1-d),
 \qquad 0.09\le c,d\le0.11.
\]
Let $a=A/10=q_i+c$, $b=B/10=q_i-1-d$, and $\Delta=a-b=1+c+d\in[1.18,1.22]$.
For a literal with $v=\theta$ or $v=1-\theta$, multiplying the original row by $v/10$ and applying CG rounding produces
\begin{equation}\label{scabs:eq:cut}
 \lceil av\rceil x\ge \lceil bv\rceil .
\end{equation}
Define its \emph{strong} state by $\lceil av\rceil=\lceil bv\rceil\ge1$. It then forces $x=1$. If the cut is not strong, it either is vacuous ($v=0$), or it has a strictly smaller right-hand integer than the coefficient.

\begin{lemma}[Strong-cut criterion]\label{scabs:lem:strong}
The interval formula of Lemma~\ref{lem:cg-one-row-interval} gives an equivalent geometric description. For $0<v\le1$, the row cut is strong iff $\Delta v<1$ and, writing $t=\{av\}\in[0,1)$, either $t=0$ or $t>\Delta v$. At $v=0$ the cut is vacuous.
\end{lemma}
\begin{proof}
This is the fractional-part form of Lemma~\ref{lem:cg-one-row-interval} applied to $a,b$ and the multiplier $v$.  Write $av=k+t$ with $0\le t<1$. Equality of the positive ceilings requires $\Delta v<1$. When $t=0$ they both equal $k\ge1$, and for $t>0$ they both equal $k+1$ exactly if $t>\Delta v$.
\end{proof}

\begin{lemma}[Uniform exclusion of contradictory literal tests]\label{scabs:lem:exclude}
Fix $q_i$ and $\theta\in[0,1]$. Choose any independently perturbed positive row with $(c_+,d_+)\in[.09,.11]^2$ and any independently perturbed negative row with $(c_-,d_-)\in[.09,.11]^2$. Their strong-cut predicates cannot both hold, \emph{even if the two rows belong to different training instances}.
\end{lemma}
\begin{proof}
At $\theta=0,1$ one multiplier is zero. Suppose, towards a contradiction, both cuts are strong at $0<\theta<1$. By Lemma~\ref{scabs:lem:strong}, $\Delta_+\theta<1$ and $\Delta_-(1-\theta)<1$. Since $\Delta_\pm\in[1.18,1.22]$, this forces $\theta,1-\theta>1-1/1.18>0.1525$. In particular, both $\Delta_+\theta$ and $\Delta_-(1-\theta)$ exceed $0.18$.

Let $t=\{(q_i+c_+)\theta\}$ and $s=(1-\theta)c_-+\theta c_+\in[0.09,0.11]$. Since $q_i$ is an integer,
\[
 \{(q_i+c_-)(1-\theta)\}=\{s-t\}.
\]
If $t=0$, the positive cut may be strong but the negative fractional part is $s<\Delta_-(1-\theta)$, so the negative cut is not strong. If $0<t<s$, the negative fractional part is $s-t<\Delta_-(1-\theta)$, hence it is not strong. If $t=s$, the negative cut may be strong at an integer but the positive cut is not strong because $s<\Delta_+\theta$.

It remains to consider $t>s$. Both cuts being strong would require
\[
 t>\Delta_+\theta,\qquad
 1+s-t>\Delta_-(1-\theta),
\]
which is impossible because
\begin{align*}
\Delta_+\theta-\big(1+s-\Delta_-(1-\theta)\big)
 &=\theta d_++(1-\theta)d_-\ge0.09.
\end{align*}
The separation is independent of $q_i$ and of all perturbation choices.
\end{proof}

\subsection{Explicit and perturbation-robust Boolean encoding}
\begin{lemma}[Base-$1000$ code]\label{scabs:lem:code}
For every $\sigma\in\{0,1\}^T$, set
\begin{equation}\label{scabs:eq:code}
 \theta_\sigma=\frac3{10}+\sum_{i=1}^T\frac{d_i}{1000^{i+1}},
 \qquad d_i=\begin{cases}150,&\sigma_i=0,\\750,&\sigma_i=1.\end{cases}
\end{equation}
Then $0.3001<\theta_\sigma<0.301$, its bit length is $O(T)$, and for every occurrence of variable $i$ under every permitted additive perturbation, the positive literal cut is strong exactly when $\sigma_i=1$, and the negative literal cut is strong exactly when $\sigma_i=0$.
\end{lemma}
\begin{proof}
For $q_i=1000^i$, the number $q_i\theta_\sigma$ has fractional part
\[
 t_i=\frac{d_i}{1000}+\sum_{j>i}\frac{d_j}{1000^{j-i+1}}.
\]
The tail is smaller than $0.000752$. Consequently, $t_i\in[.150,.151)$ when $\sigma_i=0$, and $t_i\in[.750,.751)$ when $\sigma_i=1$.

For any perturbed row, $a=q_i+c$ with $c\in[.09,.11]$. Because $\theta_\sigma\in[.3001,.301]$, the positive fractional part $\{a\theta_\sigma\}=t_i+c\theta_\sigma$ lies in $[.177,.185)$ (zero bit) or $[.777,.785)$ (one bit). Meanwhile $\Delta\theta_\sigma\in(.354,.368)$. By Lemma~\ref{scabs:lem:strong}, precisely the one bit activates the positive cut.

Similarly, $\{a(1-\theta_\sigma)\}=1-t_i+c(1-\theta_\sigma)$ lies in $(.911,.928)$ (zero bit) or $(.311,.328)$ (one bit), while $\Delta(1-\theta_\sigma)\in(.824,.854)$. Precisely the zero bit activates the negative cut. All comparisons have constant slack, uniformly over perturbations.
\end{proof}

\begin{lemma}[Weak-cut redundancy at code witnesses]\label{scabs:lem:weak}
At any witness $\theta_\sigma$ in Lemma~\ref{scabs:lem:code}, every non-strong literal cut is implied by the original row's LP relaxation.
\end{lemma}
\begin{proof}
A non-strong cut has $\lceil av\rceil=k+1$ and $\lceil bv\rceil=k$, where $v\in[.3,.7]$, so it is $x\ge k/(k+1)$. The original bound is $x\ge b/a$. The new cut is redundant iff $\Delta k\le b$. Since $k\le av\le.7a$, we have $\Delta k\le1.22(.7)a=.854a<b$, because $a\ge1000$ and $b=a-\Delta\ge a-1.22>.998a$.
\end{proof}

\subsection{Root-gap and tree-size clause gadgets}
For a clause $C$ of at most three literals on distinct Boolean variables, use a single $x\in\{0,1\}$, positive objective $\min w x$ (with $w\in[.9,1.1]$ after perturbation), and one independently perturbed row $-A_jx\le-B_j$ per literal. Let $x_0=\max_j B_j/A_j\in(0,1)$, and let $x_\theta$ be the root LP minimizer after the generated cuts.

\begin{lemma}[Robust OR and a constant failed-probe gap]\label{scabs:lem:or}
For every $\theta\in[0,1]$ and every permitted perturbation,
\[
 L_C(\theta)=\frac{1-x_\theta}{1-x_0}=0
 \quad\Longleftrightarrow\quad\text{some literal row of $C$ is strong}.
\]
If no literal row is strong, then $L_C(\theta)\ge4/5$. At the canonical parameter witnesses $\theta_\sigma$, this is the \emph{exact} Boolean loss $L_C(\theta_\sigma)=\mathbf1\{C(\sigma)\text{ is false}\}$.
The unmodified one-variable B\&B creates one node iff the loss is zero, and three nodes otherwise.
\end{lemma}
\begin{proof}
The integer optimum is $w$ because $B_j>0$ and $A_j>B_j$, so the only feasible binary point is $x=1$. A strong cut forces $x=1$, closing the gap. Conversely, without a strong cut, each rounded coefficient either vanishes, producing a vacuous cut, or is a positive integer $M_j=\lceil A_ju_j\rceil$ with right-hand integer $N_j\le M_j-1$. All resulting bounds are strictly below one. Thus $x_\theta<1$.

For a quantitative bound, put $a_j=A_j/10$, $\Delta_j=(A_j-B_j)/10\in[1.18,1.22]$, and $a_{\max}=\max_j a_j\ge1000$. The root gap in the $x$ coordinate is $\delta_0=1-x_0=\min_j\Delta_j/a_j\le1.22/a_{\max}$. For every non-strong cut, $M_j\le a_j+1\le a_{\max}+1$ and $N_j/M_j\le1-1/M_j$, so
\[
1-x_\theta\ge \min\left\{\delta_0,\frac1{a_{\max}+1}\right\}.
\]
Hence
\[
 L_C(\theta)\ge\min\left\{1,\frac{a_{\max}}{1.22(a_{\max}+1)}\right\}>\frac45.
\]
Lemma~\ref{scabs:lem:weak} proves exact $0/1$ values on the canonical witnesses. When $x_\theta=1$ the root LP is integral. Otherwise its optimum is fractional, so a single binary branch creates an infeasible $x=0$ child and a feasible integral $x=1$ child. The generated-node count is respectively one or three.
\end{proof}

\subsection{Hardness and statistical lower bound}
\begin{proof}[Proof of Theorem~\ref{scabs:th:main}]
Given a 3-CNF formula $\Phi$ with $T$ variables and $M$ clauses, build one training IP from each clause as above. For a satisfying assignment $\sigma$, the rational parameter $\theta_\sigma$ in Lemma~\ref{scabs:lem:code} makes at least one literal cut strong in every clause, yielding zero total loss for every perturbation.

Conversely, suppose an arbitrary $\theta\in[0,1]$ closes the gap on every clause. In each clause choose a strong literal. By Lemma~\ref{scabs:lem:exclude}, no variable can have both positive and negative literals strong anywhere in the training collection, even though all row perturbations are independent. Thus the chosen strong literals define a consistent partial Boolean assignment satisfying every clause. Extending it to a total assignment satisfies $\Phi$. This proves the zero-loss equivalence \emph{pointwise} for every perturbation.

For a perfect-completeness bounded-width gap-3SAT instance with every assignment violating at least $\epsilon_0M$ clauses, the strong-literal partial assignment at any $\theta$ can be extended to a total assignment. Every closed clause is satisfied under this extension, hence at least $\epsilon_0 M$ clauses are not closed. By Lemma~\ref{scabs:lem:or}, their average remaining-gap loss is at least $(4/5)\epsilon_0$, and their average excess tree size is at least $2\epsilon_0$. YES instances have both losses zero. This proves constant-additive hardness. A finite-factor algorithm for nonnegative remaining-gap loss would solve zero-loss feasibility.

For fat-shattering, use $T$ unary clause instances $C_i=(x_i)$, each with an independently perturbed row. At $\theta_\sigma$, Lemmas~\ref{scabs:lem:code}--\ref{scabs:lem:weak} give $L_{C_i}(\theta_\sigma)=1-\sigma_i$, so the thresholds $1/2$ and margin $1/2$ witness $\fat_{1/2}\ge T$.

The input is polynomially constructible because $q_i=1000^i$ has $O(i)$ bits and $\theta_\sigma$ has $O(T)$ bits. All fixed-parameter evaluations on rational data involve only polynomial-bit multiplication and integer ceilings. For the original rational training instances, zero-loss feasibility has a polynomial-bit parameter certificate. The finitely many cut statuses are constant on cells delimited by rational breakpoints of the form $k/a_j$, $k/b_j$, $1-k/a_j$, or $1-k/b_j$ (including interval endpoints), all with polynomial bit length. Each nonempty cell has a rational point of polynomial bit length, and endpoints are handled separately. Thus the restricted unperturbed zero-loss problem is NP-complete.
\end{proof}

\subsection{Pseudo-polynomial ERM and relative-smoothed tractability}
\begin{theorem}[Output-sensitive and relatively smoothed binary CG coverage ERM]\label{thm:cg-relative-tract}
Consider any collection of one-binary-variable rows $-A_jx\le-B_j$,
with $0<B_j<A_j$, whose independently rounded cuts use one scalar
$\theta$ through fixed affine maps $u_j(\theta)$ into $[0,1]$.
Assume the loss of each training IP depends only on which rows
produce a \emph{strong} cut, as for zero-gap coverage (the indicator
of an open root integrality gap) or the three-node versus one-node
B\&B tree-size loss. Let $s_j=1-B_j/A_j$. With rational data,
exact ERM for these binary losses is computable in time
polynomial in the input bit length and
\[
  \sum_j\min\{\lceil A_j\rceil,\lceil1/s_j\rceil\},
\]
by sorting the strong-cut endpoints and evaluating each open cell and
endpoint.

Suppose the positive row coefficients are independently perturbed as
$A_j'=A_j(1+\xi_j)$, $B_j'=B_j(1+\zeta_j)$, with independent uniform
$\xi_j,\zeta_j\in[-\rho,\rho]$, $0<\rho\le1/4$, and 
condition each row on preserving $0<B_j'<A_j'$. Then the expected
number of \emph{strong-cut} candidate endpoints is
\[
 O\!\left(\sum_j\frac{1+\log(2+A_j)}{\rho}\right).
\]
For rational finite-precision smoothing with at least
$\lceil\log_2(2+\max_j A_j)\rceil+O(\log(1/\rho)+\log m)$
random bits of resolution per noise variable and exact rational arithmetic,
the discretization term in the small-slack tail is negligible, and this gives expected-polynomial-time exact ERM for the stated \emph{coverage/tree-size} losses whenever $1/\rho$ and the number of rows are polynomially bounded. It does not imply the same expectation bound for minimizing the exact, continuously valued remaining-root-gap loss, whose non-strong cuts can still change the LP optimum. Nor does it give an algorithm for a single aggregated cut with shared multiplier coordinates or for general multi-variable IPs.
\end{theorem}
\begin{proof}
By Lemma~\ref{lem:cg-one-row-interval}, the status of row $j$ changes
as \emph{strong versus non-strong} only at endpoints $(k-1)/B_j$ and $k/A_j$ with $1\le k<1/s_j$. Applying the inverse of the affine map $u_j$ gives rational $\theta$-endpoints. Between consecutive endpoints every strong-cut status, and hence every stipulated binary loss, is unchanged. Endpoint
values are evaluated separately to respect the open-left/closed-right
convention.  This proves the output-sensitive bound.

For the expectation, condition on $A_j'$ and write
$s_j'=1-B_j'/A_j'$.  The random variable $s_j'$ is affine in
$\zeta_j$ with conditional density at most
$A_j'/ (2\rho B_j)$ on its support.  If $s_j'$ is in $(0,t)$, then
$B_j'/A_j'$ lies in $(1-t,1)$, so the relevant conditional probability
is $O(t/\rho)$ uniformly over the base ratio $B_j/A_j<1$.
Furthermore, conditioning on $A_j'>B_j'$ has probability at least
$1/2$, because the identically distributed independent multipliers
satisfy $\Pr(\xi_j\ge\zeta_j)=1/2$ and $A_j>B_j$.
Thus $\Pr(s_j'<t\mid s_j'>0)=O(t/\rho)$.
Using $A_j'\le(1+\rho)A_j$ and integrating the tail of
$\min\{1/s_j',(1+\rho)A_j+1\}$ yields
$O((1+\log(2+A_j))/\rho)$. Summation gives the expected event count.
The endpoint sweep uses polynomial arithmetic per event. For a uniform $b$-bit noise grid, a small interval contains at most its continuous-density mass plus one grid point, adding $O(2^{-b}/\rho)$ to the tail estimate. Choosing $b\ge\log_2(2+\max_j A_j)+O(\log(1/\rho)+\log m)$ keeps its contribution to the expected interval count polynomial.
\end{proof}

\begin{remark}[Numerical root-gap ERM is a distinct objective]
The preceding relative-smoothed bound counts \emph{closure} windows,
not every change of the rounded coefficient pair.
For example, with $A=10$, $B=5$, both $u=0.21$ and $u=0.41$ generate
non-strong cuts but yield different bounds on $x$:
$\lceil2.1\rceil x\ge\lceil1.05\rceil$ gives $x\ge2/3$,
whereas $\lceil4.1\rceil x\ge\lceil2.05\rceil$ gives $x\ge3/5$.
Exact numerical remaining-gap ERM can nevertheless be performed in
pseudo-polynomial time by also scanning every floor/ceiling breakpoint
$k/A_j$ and $k/B_j$ in the bounded multiplier domain, with
$O(\sum_j(1+A_j+B_j))$ candidate events.  An analogous
expected-polynomial bound under relative smoothing does not follow
from the closure-window argument alone.
\end{remark}

\begin{proposition}[The cut-closing frequency obstruction]\label{scabs:prop:obstruction}
For a one-variable row $-Ax\le-B$ with $0<B<A$ and multiplier $u>0$, a CG cut can force $x=1$ only if $(A-B)u<1$ and $\lceil Au\rceil=\lceil Bu\rceil$. In particular, in a high-frequency construction with $A,B=\Theta(q)$ and $u=\Theta(1)$, a uniform multiplicative perturbation radius $\rho$ cannot preserve the predicate for all perturbations once $\rho q=\Omega(1)$, because independent perturbations of $A$ and $B$ can make $A-B$ negative (destroying $x=1$ feasibility) or greater than $1/u$ (precluding every strong cut).
\end{proposition}
\begin{proof}
If $\lceil Au\rceil=\lceil Bu\rceil$, the interval $(Bu,Au]$ does not cross an integer, requiring $(A-B)u<1$. Relative changes $A(1\pm\rho)$ and $B(1\mp\rho)$ change $A-B$ by approximately $\pm2\rho q$, whereas the nominal difference in the construction is the constant $12$. For $q\gg1/\rho$ the claimed opposite outcomes are both permitted.
\end{proof}
\begin{remark}[Precise assumptions]
The fixed-unit robustness construction uses independent \emph{absolute additive} errors, not independent relative errors $v\mapsto v(1+\eta)$. For these feasible one-variable rows, Theorem~\ref{thm:cg-relative-tract} gives an expected-polynomial algorithm for zero-gap coverage and binary tree-size ERM under independent relative smoothing. Exact numerical remaining-gap ERM, multi-variable rows and one aggregated cut are not settled by that interval count. A single \emph{aggregated} CG cut is also different from the batch of at most three individually rounded cuts used here. The polarity-dependent affine maps $\theta/10$ and $(1-\theta)/10$ are part of the fixed policy. A common identical multiplier for every row is a narrower model.
\end{remark}

\begin{proposition}[Polynomial-magnitude scalar breakpoint boundary]
Consider an affine one-parameter family of one-round CG cuts with rational coefficients and RHSs of numerical magnitude at most $H$, and assume that a fixed configuration can be evaluated in polynomial time. If the affine multipliers stay in a fixed bounded interval and $H$ is polynomially bounded in the binary input size, then exact ERM is polynomial-time via enumeration of all floor/ceiling breakpoints, for any polynomial number of rows and training instances.
\end{proposition}
\begin{proof}
Each rounded expression has the form $\lfloor (a\theta+b)c\rfloor$ with rational constants and bounded coefficient magnitude, and changes at $O(H)$ rational parameter values on the bounded interval (with the affine slope also bounded). Collect all such boundaries for the whole training collection, and evaluate one rational representative in every open cell and at every boundary. The total number of candidates and all arithmetic bit lengths are polynomial.
\end{proof}

\section{Gradient-descent realization with a full-branch step-size code}\label{app:gd-new}
This section uses capped gradient-norm stopping time, not validation error, for ordinary gradient descent on piecewise-quadratic objectives. The construction realizes full-branch symbolic dynamics directly through the GD update. Throughout, $x_0$ is the initial iterate, a successful check at $x_t$ has cost $t$ (equivalently $t+1$ in the one-based convention of \cite{SharmaGD2026}), and the nonconvergence cost is $H$. Divide costs by $H$ for statistical assertions. A \emph{regular parameter} is one whose run avoids all piece boundaries through the horizon. The 2026 upper-bound model explicitly assumes such boundary avoidance. Our completeness witnesses are regular, and the hardness reductions below quantify over regular parameters. An explicit piece-priority convention can alternatively extend the algorithm to boundary parameters, but must be stated as part of that variant's input model.

\subsection{A full-branch decoder lemma}
This decoder is a quantitative pullback realization of the nested-interval code in Lemma~\ref{lem:interval}. The nontrivial step here is controlling the parameter derivative and retaining strict branch margins.
For $0<\delta\le 1/4$, let $J=[-\delta,1+\delta]$ and $J'=[-\delta/2,1+\delta/2]$. Denote the open branches by $B_0^\circ=(0,1/2)$ and $B_1^\circ=(1/2,1)$.
\begin{lemma}[Expanding parameter decoder]\label{gdnew:lem:decoder}
Suppose for $t\ge 1$ the maps $g_{t,\eta}$ are affine on each branch, their branch images contain $J$, and $|\partial_xg_{t,\eta}|\ge\lambda>1$ and $|\partial_\eta g_{t,\eta}|\le K$. Suppose $x_1(\eta)$ is continuously differentiable and monotone with $x_1(W)\supseteq J'$, $|x_1'|\ge G$, $(\lambda-1)G>K$, and $K<\delta G$. Then each $w\in\{0,1\}^L$ has a nonempty, positive-length parameter interval on which $x_t(\eta)\in B_{w_t}^\circ$ for $1\le t\le L$. These intervals may be chosen disjoint, with a subinterval on which $x_{L+1}$ is monotone, has derivative of magnitude at least $G$, and covers $J'$. If branch derivatives are also bounded by $\Lambda$ and $|x_1'|\le G_{\max}$, the witness interval contains a rational with $O(L\log(\Lambda+K/G)+\log G_{\max}+B)$ bits when the data have $B$-bit rational coefficients and polynomially bounded denominators.
\end{lemma}
\begin{proof}
Inductively take a monotonic parameter interval for which $x_t$ covers $J'$ and all earlier branch inequalities are strict. For each branch choose the subinterval where $x_t$ runs through its closure. Its length is at most $1/(2G)$. Along its interior the derivative of $x_{t+1}=g_{t,\eta}(x_t)$ has constant sign and magnitude at least $\lambda G-K>G$. At a fixed parameter in that interval, the branch image contains $J$. Varying the parameter changes the branch image by at most $K/(2G)<\delta/2$, so the actual endpoint values of $x_{t+1}$ strictly bracket $J'$. Take the two interior points where $x_{t+1}$ equals the endpoints of $J'$. These yield a \emph{closed interval strictly inside the chosen branch} for $x_t$, so the earlier branch inequalities remain strict. In particular, it is incorrect to use the closed branch intervals $[0,1/2]$ and $[1/2,1]$ themselves and claim both are disjoint while respecting a fixed boundary convention. The derivative upper bound gives an interval width at least $\Omega((\Lambda+K/G)^{-L}/G_{\max})$ after a fixed rational interior trimming. Binary search and rational separation provide the stated witness length.\end{proof}

For the concrete decoder put $\eta_0=1/9$, $c=11/16$, and define $F'$ on $(0,1)$ by
\begin{equation}\label{gdnew:eq:gradient}
F'(x)=\begin{cases}\frac{145}{4}x-\frac{91}{8},&0<x<1/2,\\36(x-c),&1/2<x<1.\end{cases}
\end{equation}
The base map $g(x)=x-\eta_0F'(x)$ has branches
\begin{equation}\label{gdnew:eq:maps}
 g_0(x)=\frac{91}{72}-\frac{109}{36}x,\qquad g_1(x)=\frac{11}{4}-3x.
\end{equation}
Their images strictly contain $[-1/5,6/5]$. At the respective branch endpoints the first gives $[ -1/4,91/72]$, the second $[-1/4,5/4]$. On a sufficiently small window about $\eta_0$ the branch slopes have magnitude at least $2.9$, and $\sup_{[0,1]}|F'|\le 12$. Set $x_0=-\eta_0G$ in a launch region with $F'(x_0)=-G$. Then $x_1(\eta)=G(\eta-\eta_0)$. Take $G$ polynomially larger than $H$ and the required branch and clock constants, and $W=[\eta_0-0.1/G,\eta_0+1.1/G]$. The lemma now supplies all finite binary itineraries with $O(L+\log G)$-bit witnesses. The same argument works for $k$ affine full branches by partitioning $[0,1]$ into $k$ equal intervals and taking $F'=(x-g(x))/\eta_0$ on each interval.

\subsection{Statistical lower bounds for capped iteration count}
\begin{theorem}[Three matching capacity lower bounds]\label{gdnew:thm:capacity}
For piecewise-quadratic, possibly discontinuous objectives and the normalized capped iteration-count loss, the following families exist, with regular witnesses and rational polynomial-bit descriptions.
\begin{enumerate}[label=\textup{(\roman*)}]
\item For constant step size and $H\ge4$, $k\ge2$, a family of $\lfloor(H-2)/2\rfloor\lfloor\log_2 k\rfloor$ objectives in dimension $2$, each described using $O(H+k)$ affine boundaries, is $1/4$-fat-shattered.
\item For an $H$-entry learning-rate schedule, $\binom{\lfloor H/2\rfloor}{2}=\Omega(H^2)$ two-dimensional objectives, each with $O(H)$ affine boundaries, are $1/4$-fat-shattered.
\item For a tunable initialization vector of $d$ coordinates and a fixed step size, $d\lfloor H/2\rfloor$ objectives in dimension $d+1$ are $1/4$-fat-shattered.
\end{enumerate}
These respectively match the $H\log(pd\Delta)$, $H^2\log(pd\Delta)$, and $dH\log(pd\Delta)$ \emph{upper-bound dependences} in \cite{SharmaGD2026} up to logarithmic factors, for the stated families and resource parameters.
\end{theorem}
\begin{proof}
Introduce a clock coordinate $y$ with objective term $-y$, so $y_t=\sum_{u=1}^t\eta_u$. Let the stage-$t$ band be $[(t-1/3)\eta_0,(t+1/3)\eta_0]$. Taking each learning rate within $\eta_0/(4H)$ of $\eta_0$ ensures $y_t$ belongs to stage $t$ for all $t\le H$. Since the $y$ gradient is $-1$ outside a designated readout, a stopping threshold $\vartheta=1/2$ precludes early stopping, even at critical points of $F$.

For (i), use the $k$-branch decoder in the $x$ coordinate and $f=F(x)-y$ in coding regions. Index the $k$ branches by binary words of length $q=\lfloor\log_2 k\rfloor$. For each stage $1\le t\le\lfloor(H-2)/2\rfloor$ and bit $j\le q$, build one objective that replaces the coding piece by the constant zero exactly in stage-$t$ band when branch $j$'s label bit is one. The cost is then $t$ or $H$, according to that bit. Every matrix of target bits is obtained by choosing a word of $k$-ary symbols and applying the full-branch lemma. The normalized low cost is at most $1/2$, the high cost is $1$, and threshold $3/4$ gives margin $1/4$.

For (ii), take an instance indexed by $1\le s<t\le\lfloor H/2\rfloor$. Keep $x$ stationary until clock band $s-1$. During that band set $\partial_xf=-G$, so the \emph{update with step size $\eta_s$} launches $x_s=x_0+G\eta_s$. Later steps $\eta_{s+1},\eta_{s+2},\ldots$ drive the decoder and stage $t$ is the readout. The launch band is therefore indexed by $s-1$. Choose $\eta_{\lfloor H/2\rfloor},\ldots,\eta_1$ in reverse order. At the moment $\eta_s$ is chosen all later rates are fixed, so the future maps have $K=0$ as functions of $\eta_s$. The lemma realizes every prescribed continuation word. Parameters fixed later in the reverse induction do not affect instances launched later in chronological time. One binary readout for every pair $(s,t)$ gives the count.

For (iii), let the $s$-th tunable initialization coordinate evolve by the same base map $g$, and let instance $(s,t)$ inspect only that coordinate at stage $t$ and stop if it is in $B_1^\circ$. The other coordinates remain irrelevant to that instance. Choose initial coordinate $s$ in the corresponding inverse cylinder of an arbitrary desired word. Independent choices for $s\in[d]$ realize all $d\lfloor H/2\rfloor$ bits.\end{proof}

\subsection{Four-example ReLU realization}
The following is an extension beyond the deliberately discontinuous two-dimensional objectives. It uses one trainable scalar $w$ in $h_w(u,v)=\max\{0,uw+v\}$ with the coefficient of $v$ fixed to one. This is not a claim about a neuron with no fixed offset feature. For training pairs $(u_j,v_j,y_j)$ put
\[
 f_S(w)=\frac12\sum_{j=1}^4\bigl(\max\{0,u_jw+v_j\}-y_j\bigr)^2.
\]
The loss is continuous and piecewise quadratic, though it need not be continuously differentiable at the ReLU kinks. Define the two base-map inverse-cylinder intervals
\[
 C(M)=\bigl\{w:g^r(w)\in[-1/10,11/10]\text{ along the branch word }M\bigr\}=[a_M,b_M].
\]
All endpoints are rational with $O(r)$ bits.

\begin{theorem}[Four training examples suffice for nearly linear capacity]\label{gdnew:thm:relu}
For every $m\ge2$ there are $m$ training sets of four rational examples, a common scalar initialization, an interval of positive step sizes, and $H=O(m\log m)$ such that the normalized stopping-time class $1/4$-fat-shatters these training sets. Thus $\Pdim\ge\fat_{1/4}=\Omega(H/\log H)$ for this one-trainable-weight ReLU model. All data have bit length polynomial in $m$.
\end{theorem}
\begin{proof}
Put $L=\lceil\log_2m\rceil$, $r=4+4L$, $\varepsilon=4^{-r}/1000$, $G=9/\varepsilon$ and $w_0=-G/9$. For $1\le i\le m$, form the length-$r$ marker
\[
 M_i=1110\prod_{\beta\ \text{a bit of }(i-1)}(10\beta0),\qquad \Phi=(10)^{r/2}.
\]
Represent $i-1$ in exactly $L$ bits (indexing by $i$ itself would fail when $m$ is a power of two). The string $111$ occurs only at the start of a marker in any concatenation of the $M_i$ and $\Phi$. Consequently every marker occurrence is aligned with a block and the markers are distinct. Every run of identical symbols has length at most three. Write $[a_i,b_i]=C(M_i)$ and $c=11/16$. The four examples for $S_i$ are
\begin{center}\small
\begin{tabular}{@{}llll@{}}\toprule
role & $u$ & $v$ & $y$\\\midrule
launch & $-1$ & $-1$ & $(\tfrac{149}{36}-1)G+\tfrac{83}{8}$\\
left branch & $-1/2$ & $1/4$ & $27$\\
left sink & $-6$ & $6a_i$ & $6(a_i-c)$\\
right sink & $6$ & $-6b_i$ & $6(c-b_i)$\\\bottomrule
\end{tabular}
\end{center}
Direct differentiation of the squared ReLU loss gives $f'_{S_i}(w)=-G$ at $w=w_0$, equals \eqref{gdnew:eq:gradient} on $(0,1/2)$ and on $(1/2,a_i)\cup(b_i,1)$, and vanishes identically on $(a_i,b_i)$. The launch example is inactive once $w>-1$. The second example is active exactly below $1/2$. The sink examples are respectively active below $a_i$ and above $b_i$. Thus all instances share the same decoder until one is absorbed at its own flat sink.

For a requested label vector $s\in\{0,1\}^m$, concatenate $B_i=M_i$ if $s_i=1$ and $B_i=\Phi$ otherwise, and append enough alternating $10$ symbols to define at least $H+4$ itinerary positions. The decoder lemma gives a regular $\eta$ realizing this entire word. The full parameter window satisfies $|\eta-\eta_0|\le1.1/G$, and the relevant branch states lie in $[0,1]$. Thus the uniform perturbation bound $1.1\cdot12/G=13.2/G$ gives, on any $r$ consecutive prescribed branch steps,
\[
 \bigl|g_\eta^r(w)-g^r(w)\bigr|
 \le \frac{13.2/G}{(3.04-1)}(3.04^r-1)<10^{-3};
\]
for the displayed choice of $G=9/\varepsilon$ and $r\ge8$. The same estimate applies to every shorter prefix. Each base branch sends its endpoints at least $1/4$ beyond $[0,1]$. Inverting the guarded interval $[-1/10,11/10]$ therefore places every intermediate state at least $0.04$ from its branch boundary (the last inverse step has margin exceeding $0.049$). Consequently $[a_i,b_i]$ has strict intermediate-itinerary margins. These estimates establish both (a) any witness whose itinerary begins $M_i$ enters $(a_i,b_i)$, and (b) an orbit point in $(a_i,b_i)$ has the $\eta$-itinerary $M_i$. The $10^{-3}$ bound is smaller than these explicit $0.04$ branch-prefix margins. At a marker that is present, instance $S_i$ therefore stops at $t_i=1+(i-1)r$. At an absent marker, it never reaches its own sink.

It remains to preclude stopping at an ordinary decoder critical point. The fixed points of the two branches are $z_0=91/290$ and $c=11/16$. Every length-four window in the prescribed word contains both symbols. Checking the fourteen length-four binary cylinders other than $0000$ and $1111$ using the rational inverse maps of \eqref{gdnew:eq:maps} gives
\[
 \min |F'(w)|=\frac{455868}{1295029}>0.352
\]
for the corresponding \emph{base-map} cylinders with a fifth iterate still in $[0,1]$. The preceding $10^{-3}$ perturbation estimate changes this bound by less than $0.04$, leaving a lower bound above $0.3>\vartheta=1/4$. The four-symbol padding covers the final iterations. Hence no pre-sink stopping is possible. Taking $H=2mr+8$ makes each present marker stop before $H/2$, whereas an absent marker incurs cost $H$. The threshold $3/4$ gives margin $1/4$ for every $s$.\end{proof}

\begin{corollary}[Uniform-convergence lower bound]\label{gdnew:cor:sample}
For the uniform distribution on the $m$ four-example training sets of Theorem~\ref{gdnew:thm:relu}, any distribution-free uniform-convergence guarantee with constant success probability and error $0<\epsilon<1/8$ requires $\Omega(m/\epsilon^2)=\Omega(H/(\epsilon^2\log H))$ task samples (in the usual nontrivial regime).
\end{corollary}
\begin{proof}
For a multinomial sample with counts $N_i$, the ability to choose independently between cost $1$ and cost at most $1/2$ on every instance implies
\[
 \sup_\eta\bigl|\widehat{\mathbb E}[c_\eta/H]-\mathbb E[c_\eta/H]\bigr|\ge\tfrac14\sum_{i=1}^m\left|N_i/n-1/m\right|.
\]
The standard multinomial $\ell_1$ deviation lower bound is of order $\min\{1,\sqrt{m/n}\}$ with constant probability. This yields the claim.\end{proof}

\begin{corollary}[Three examples for initialization]\label{gdnew:cor:reluinit}
With fixed step size $\eta_0$, three squared-ReLU examples and a tunable scalar initialization $w_0\in(0,1)$ yield $\Pdim=\Omega(H/\log H)$.
\end{corollary}
\begin{proof}Drop the launch example and use the common inverse-cylinder initial state to encode the chosen marker word. The remaining derivatives and sink tests are unchanged.\end{proof}

\subsection{Two-dimensional ERM hardness, arity and finite precision}
For $n$ Boolean variables build markers for all $2n$ positive and negative literals, each of length $r=4+4\lceil\log_2(2n)\rceil$, and let $S_\ell\subset(1/2,1)$ be their rational sink intervals. Set
\[
 T_*=(n-1)r+1,\qquad Y_*=(T_*+\tfrac12)\eta_0.
\]
Choose $G$ sufficiently large that the full-branch lemma and the $r$-step marker margins hold and $|\eta-\eta_0|\le\eta_0/(4H)$ throughout the tuning window. For any literal set $\Sigma$ of size at most three, take the objective with pieces
\begin{equation}\label{gdnew:eq:literal-objective}
 f_\Sigma(x,y)=\begin{cases}
 0,& y\le Y_*\ \text{and}\ x\in\bigcup_{\ell\in\Sigma}S_\ell,\\
 F(x)-y,& x\in(0,1)\quad\text{otherwise},\\
 -Gx-y,& x\le x_0+1\quad\text{(launch)},\\
 -y,&\text{otherwise}.
 \end{cases}
\end{equation}
The clock makes the gradient norm at least one off the sinks, even where $F'(x)=0$. Every objective has $O(|\Sigma|+1)$ affine boundaries, degree at most two, and dimension two. It stops exactly when a marker from $\Sigma$ begins by stage $T_*$.

\begin{lemma}[Slot-counting without persistent state]\label{gdnew:lem:slots}
For any regular parameter, a sink hit at stage $t$ implies that the corresponding length-$r$ marker begins at itinerary position $t$. Distinct marker occurrences cannot overlap. Therefore at most $n$ marker occurrences can begin at positions $1,\ldots,T_*$.
\end{lemma}
\begin{proof}The first assertion is the two-sided marker-margin property from the preceding proof. The word $111$ marks the start of each marker, and no other $111$ occurs inside any marker, so two occurrences cannot overlap. The $q$-th start is at least $1+(q-1)r$; hence $q\le n$ when its start is at most $(n-1)r+1$.\end{proof}

\begin{theorem}[Arity dichotomy for learning-rate ERM]\label{gdnew:thm:hard}
For the regular-parameter, piecewise-quadratic GD family \eqref{gdnew:eq:literal-objective}, with a rational tuning window $W$ and a polynomially bounded horizon $H$:
\begin{enumerate}[label=\textup{(\roman*)}]
\item Zero nonconvergence loss is NP-complete when each instance has at most three literal sinks.
\item Exact ERM for the number of nonconverging instances is NP-hard already with two sinks per instance. Exact ERM for total iteration count is also NP-hard for the same two-sink family.
\item There exists an absolute $c>0$ for which additive-$c$ approximation of the optimal \emph{average} nonconvergence loss is NP-hard, using at most three sinks.
\item With one sink per instance, ERM for either loss is polynomial-time on this restricted family.
\end{enumerate}
These are worst-case results. They do not establish analogous hardness for squared-error neural-network training objectives or independent per-instance smoothing.
\end{theorem}
\begin{proof}
For (i), reduce from 3-SAT. Add one variable instance with sink set $\{x_j,\neg x_j\}$ for every variable and one clause instance with sink set its at most three literals. A satisfying assignment supplies $n$ nonoverlapping true-literal markers, concatenated in one word; Lemma~\ref{gdnew:lem:decoder} realizes that word with a rational step size and every instance stops. Conversely, if every variable instance stops, the $n$ variable instances force at least one distinct marker per variable. Slot counting permits at most $n$, so each variable has exactly one of its two literal markers and all stopped clause instances are satisfied. For the constructed family, the YES region is a finite union of univariate semialgebraic sign cells. Branch iterations have degree at most $H$ in $\eta$ and coefficients of polynomial bit length, so standard univariate root separation yields a regular rational witness of polynomial bit length whenever a feasible open sign cell exists. Exact rational GD simulation is polynomial in the encoded input and in $H$. Thus the restricted promise family has polynomially checkable NP certificates.

For (ii), reduce from \textsc{Max-2-SAT}. If the formula has $M$ clauses, make $M+1$ copies of every two-sink variable instance and one two-sink clause instance per clause. Some encoded assignment loses at most $M$ clause instances. Any parameter that misses a variable instance loses at least $M+1$ copies. The optimum hence enforces one literal per variable and its number of failures equals the minimum number of unsatisfied clauses. For iteration-count ERM choose $H>N T_*$, where $N$ is the number of instances, so one additional nonconvergence outweighs every possible change in the aggregate stopping times of successful runs.

For (iii), start with a bounded-occurrence gap-3SAT family. Each variable has at most $B$ occurrences, and in NO instances every assignment falsifies at least $\alpha M$ of the $M$ clauses. Make $B$ variable copies per variable. For an arbitrary parameter let $P$ be the \emph{distinct} literal markers starting by $T_*$, $Z$ the number of uncovered variables, and $D$ the number whose two literals appear. From $|P|=n-Z+D\le n$ follows $D\le Z$. Choose an assignment agreeing with the singly covered variables. Resolving the $D$ doubly covered variables arbitrarily can destroy at most $BD$ clauses previously hit by their two literal markers. Thus at least $\alpha M-BD$ clause instances must fail, and at least $BZ$ variable-copy instances fail. The total is at least $\alpha M+B(Z-D)\ge\alpha M$. Because $n\le3M$, there are at most $(3B+1)M$ instances, yielding the constant gap $\alpha/(3B+1)$. This is the slot-counting specialization of the defect-charge principle in Lemma~\ref{lem:charge}, with the doubly covered coordinates charged to uncovered variables. 

For (iv), when all sink sets are singleton, each distinct marker can satisfy only its own group of duplicate instances. At most $n$ distinct markers fit. Choose the $n$ literal markers with largest frequencies and place them earliest in decreasing frequency order, which also minimizes the aggregate stopping-time sum. The full-branch lemma computes a rational realization by recursive interval refinement.\end{proof}

\begin{remark}[Independent perturbations and what is \emph{not} proved]\label{gdnew:rem:smoothing}
If each instance's launch or decoder coefficients are perturbed independently by $\zeta$, the same $\eta$ may produce different length-$T$ words. An $O(\zeta)$ initial discrepancy is expanded by roughly $3^T$. Thus the preceding two-dimensional, multiple-instance slot-counting proof is \emph{not} a smoothed-hardness proof under independent perturbations. This is a failure of this reduction, not a theorem that smoothed learning-rate ERM is tractable or that all smoothed-hardness constructions are impossible. A \emph{different}, single-objective construction can record the bits in auxiliary GD coordinates and evaluate all clauses with one terminal objective. Because there is then only one decoder, relative perturbations of coefficients inside that objective need not destroy cross-instance consistency. Any such claim must specify the compact factorized representation, terminal-zero preservation, perturbation of boundaries and the finite-precision model. The independently perturbed objective-copy setting is left open here.
\end{remark}

\section{A complete rational-metric reduction for power-mean linkage}
\label{app:linkage-complete}
We construct a symmetric rational metric and prove the full greedy merge schedule for power-mean linkage. The exponent ranges over the standard power-mean family, with deterministic tie-breaking. The hardness result concerns worst-case root-split realizability. Theorem~\ref{thm:linkage:smoothed} separately controls expected hierarchy changes under independent primitive-distance smoothing. Polynomial expected Turing bit complexity remains open.

For two disjoint nonempty clusters $U,V$ and $\alpha\in\mathbb R\setminus\{0\}$ write
\begin{equation}\label{lz:eq:mean}
 D_\alpha(U,V)=\left(\frac1{|U||V|}\sum_{u\in U,v\in V}\delta(u,v)^\alpha\right)^{1/\alpha}.
\end{equation}
At $\alpha=0$ and $\alpha=\pm\infty$ use the geometric mean and minimum/maximum, respectively. At each step, unmodified agglomerative linkage merges the pair of clusters with smallest $D_\alpha$, with any fixed deterministic tie-breaking.  \textsc{ZR2} is the problem of deciding whether some $\alpha\in\mathbb R\cup\{\pm\infty\}$ realizes a specified root bipartition $L\mid R$.

\begin{theorem}[Worst-case root-split linkage hardness]\label{lz:thm:zr2}
\textsc{ZR2} is NP-hard under polynomial-time many-one reductions, already for a symmetric rational metric on $O((n+m)^2)$ points whose nonzero distances lie in $[1,2]$ and have $O(n+\log(n+m))$ binary bits. Here $n,m$ are the variable and clause counts in the source \textsc{3-SAT} instance. The reduction is valid for every fixed deterministic tie-breaking rule.
\end{theorem}
The result is an NP-\emph{hardness} result for real-parameter existence, not an NP-completeness claim. Polynomial-bit witnesses and exact polynomial-time linkage comparisons for arbitrary real exponents are not established.

\subsection{Full metric and parameter scales}
Reduce from \textsc{3-SAT} with exactly three distinct variables in each clause. Let $N=n+1$ assignment rounds, the first a dummy, and $k(u)=u+1$ for variable $u$. Fix the rational clock constants
\[
\begin{aligned}
 a&=\frac{11}{10},\qquad b=\frac32,\qquad \tau=\frac1{25},\qquad
 q=\frac{2a\tau}{1-\tau^2}=\frac{55}{624},\qquad
 \mu=\frac{a(1+\tau^2)}{1-\tau^2}=\frac{3443}{3120},\\
 \lambda&=\frac1{2000},\qquad R_0=32,\qquad o_j=\lambda R_0^{-j}.
\end{aligned}
\]
In particular, $\mu^2=a^2+q^2$ and $a-q>1$. Set
\[
 \gamma=\frac1{100(N+1)},\quad c_k=\frac54+4k\gamma,
 \qquad \Delta=\frac1{100(m+1)},\quad d_t=\frac{31}{20}+t\Delta,
\]
\[
 W=1200(N+m+2)^2,\quad T=2W+2N+m+12,\quad H_0=300T,
\]
\[
 h_t=d_t+\frac1{10},\quad \ell_t=d_t-\frac1{10},\quad
 s_t=W+N+t+1,\quad \rho_t=d_t+\frac1{5s_t}.
\]
The target $R$ consists of the two sentinels $r_1,r_2$, and all other points belong to $L$. The six \emph{preformed groups} are the seeds $A^0,B^0$, each with $W$ neutral points and two anchors, the sentinel group $R=\{r_1,r_2\}$, the two safe pairs $\Sigma^-,\Sigma^+$, and the cleanup group $\mathcal H$ with $H_0$ points. Each preformed group has all its internal distances equal to $1$. In addition there are singletons $p_k,q_k$ for $k\in[N]$, clause probes $z_t$ for $t\in[m]$, and guards $g^-,g^+$. The total is $T+H_0$ points.

Every unlisted distance between distinct points is $2$. The following table specifies all overrides. All entries are symmetric.  Denote by $a^A_1,a^A_2,a^B_1,a^B_2$ the four seed anchors.
\begingroup\small
\begin{longtable}{@{}p{.57\linewidth}p{.37\linewidth}@{}}\toprule
Unordered pair family & Assigned distance \\\midrule
Within any one preformed group & $1$\\
$p_k,q_k$ to anchor $a_1^A$ & $a-q$\\
$p_k,q_k$ to anchor $a_2^A$ & $a+q$\\
$p_k,q_k$ to either $B^0$ anchor & $\mu$\\
$p_k$ to a neutral point of $A^0$ or $B^0$ & $c_k$\\
$q_k$ to a neutral point of $A^0$ or $B^0$ & $c_k+\gamma$\\
$p_j$ to $p_k$ or $q_k$ for $j<k$ & $b+o_j$\\
$q_j$ to $p_k$ or $q_k$ for $j<k$ & $b-o_j$\\
$z_t$ to any point of $A^0$ & $d_t$\\
$z_t$ to any point of $B^0$ & $9/5$\\
$z_t$ to $p_k$ or $q_k$ when $k$ is absent from clause $C_t$ & $d_t$\\
$z_t$ to $(p_{k(u)},q_{k(u)})$ for a variable occurring in $C_t$ with $\beta_u=+1$ & $(h_t,\ell_t)$\\
$z_t$ to $(p_{k(u)},q_{k(u)})$ for a variable occurring in $C_t$ with $\beta_u=-1$ & $(\ell_t,h_t)$\\
$z_t$ to $z_{t'}$ for $t<t'$ & $d_{t'}$\\
$z_t$ to either sentinel & $\rho_t$\\
$g^-$ to either point of $\Sigma^-$ & $81/50$\\
$g^-$ to $(r_1,r_2)$ & $(38/25,43/25)$\\
$g^+$ to the two points of $\Sigma^+$ & $(248/125,136/125)$\\
$g^+$ to $(r_1,r_2)$ & $(8/5,8/5)$\\
A point of $\mathcal H$ to any other point of $L$ & $43/25$\\
\bottomrule
\end{longtable}
\endgroup
The clause row is static input data, not a distance that depends on the realized clustering.  Define $\sigma_u=+1$ when $p_{k(u)}$ joins evolving seed $\mathcal A$ and $\sigma_u=-1$ otherwise. If $\beta_u$ is the value falsifying the literal of variable $u$ in $C_t$, assign $(\delta(z_t,p_{k(u)}),\delta(z_t,q_{k(u)}))=(h_t,\ell_t)$ for $\beta_u=+1$, and the reversed pair for $\beta_u=-1$. Thus the member that eventually joins $\mathcal A$ has distance $h_t$ exactly when this literal is false. This convention also makes the probe distance $d_t$ to \emph{both} members of every nonclause pair. No side-dependent distance is assigned after the run.

All specified distances lie in $[1,2]$. In particular every three distinct points obey $\delta(x,z)\le2\le\delta(x,y)+\delta(y,z)$, so the table is a metric. The numbers $o_j$ have $O(n)$ bits. All other entries have $O(\log(n+m))$ bits. Consequently the complete distance matrix is constructible in polynomial Turing time.

\subsection{Clock and exact cancellation}
For a vector $x=(x_1,\ldots,x_k)\in\{-1,+1\}^k$, put
\begin{equation}\label{lz:eq:clock}
 B_0(\alpha)=(a-q)^\alpha+(a+q)^\alpha-2\mu^\alpha,\quad
 G_j(\alpha)=(b-o_j)^\alpha-(b+o_j)^\alpha,\quad
 E_x(\alpha)=B_0(\alpha)+\sum_{j\le k}x_jG_j(\alpha).
\end{equation}
\begin{lemma}[Nested analytic assignment clock]\label{lz:lem:clock}
For every $x\in\{\pm1\}^{\le N}$, $E_x$ has one zero $\alpha_x\in(1.9,2.1)$ and is negative below and positive above it on $[1,3]$. The strict decisions $\operatorname{sign}E_{x_{<j}}(\alpha)$ generate a nonempty open interval for every prescribed word $x\in\{\pm1\}^N$. Whenever $x_1=-1$, that interval is contained in $(1,2)$.
\end{lemma}
\begin{proof}
The identity $\mu^2=a^2+q^2$ yields $B_0(2)=0$. Moreover $B_0(1)<0$ and $B_0(3)>0$. The exact displayed constants give the safe inequalities
\[
 \max_{[1,1.9]} B_0<-0.001458,
 \quad \min_{[2.1,3]} B_0>0.001643,
 \quad \min_{[1.9,2.1]} B_0'>0.01368.
\]
These can be verified with elementary logarithm bounds. For example, $B_0''$ is positive on $[1,3]$. The term $(a+q)^\alpha\log^2(a+q)$ alone exceeds $0.035$, whereas $2\mu^\alpha\log^2\mu<0.027$, and the other positive summand may be discarded. Hence $B_0$ is strictly convex, the maximum on $[1,1.9]$ is at an endpoint, and the quoted inequalities reduce to evaluations at $1,1.9,2.1$ and to $B'_0(1.9)>0.01368$. These bounds hold uniformly over the specified intervals.

For $\alpha\in[1,3]$, direct differentiation and $o_j\le 1/64000$ give
\[
 2o_j\le |G_j(\alpha)|\le14o_j,\qquad |G_j'(\alpha)|\le11o_j,
 \qquad\sum_{j\ge1}o_j=\frac1{62000}.
\]
Thus the perturbation of $B_0$ is less than $2.26\cdot10^{-4}$ and the perturbation of $B_0'$ less than $1.78\cdot10^{-4}$. Hence every $E_x$ is strictly increasing on $[1.9,2.1]$, has opposite signs at the endpoints, and remains strictly negative/positive on the outer parts of $[1,3]$. Its zero is unique. Moreover,
\[
 \sum_{r>j}|G_r(\alpha)|\le14\frac{o_j}{R_0-1}<2o_j\le|G_j(\alpha)|.
\]
At the zero $\alpha_y$ of any prefix $y$, every descendant $E_{y,x_j,\ldots}$ has sign $x_jG_j(\alpha_y)$, since subsequent terms are too small to reverse it. Because $G_j<0$ and all relevant functions cross from negative to positive, the descendants' zeros remain strictly on the side of $\alpha_y$ designated by $x_j$. It follows inductively that each finite decision word has a nonempty interval of parameters, with later switch points lying strictly between earlier switch points. The empty-prefix root is $\alpha_\varnothing=2$. Selecting the dummy first decision $x_1=-1$ restricts the entire surviving cell to $(1,2)$.
\end{proof}

\begin{lemma}[Choice and immediate return]\label{lz:lem:choice}
Suppose that, before round $k$, the two evolving seed clusters $\mathcal A,\mathcal B$ have equal size $s=W+2+k-1$, and each contains one member of every earlier pair. Let $x_j=-1$ iff $p_j\in\mathcal A$. For every $\alpha\in[1,2]$:
\begin{enumerate}[label=\textup{(\roman*)}]
\item the sign of $D_\alpha(p_k,\mathcal A)-D_\alpha(p_k,\mathcal B)$ is the sign of $E_{(x_1,\ldots,x_{k-1})}(\alpha)$.
\item after $p_k$ joins either seed, $q_k$ strictly prefers the other seed over the enlarged seed containing $p_k$.
\end{enumerate}
\end{lemma}
\begin{proof}
The $W$ neutral terms in the two $\alpha$-power sums are identical. Equal cardinality cancels the normalization. The anchors contribute $B_0$ and earlier pair $j$ contributes $x_jG_j$. This proves (i) even when its value is zero. For (ii), suppose $p_k$ joins $\mathcal A$. If $U_A,U_B$ are the $q_k$ power sums to the \emph{old} seeds, then $U_A-U_B=E_x\le0$. Since $\delta(q_k,p_k)=2$, the desired return inequality is equivalent to
\[
 s(U_B-U_A)+U_B<s2^\alpha.
\]
Here $|E_x|<0.008$, $U_B\le s(1.51)^\alpha$, and $2^\alpha-(1.51)^\alpha\ge0.49$ for $\alpha\in[1,2]$, proving strictness. The other case is symmetric.
\end{proof}

\subsection{Clause gate, guards, and quantitative score bounds}
\begin{lemma}[Strict clause threshold]\label{lz:lem:clauses}
For $s\ge3$, $c\in[1.55,1.56]$, $h=c+0.1$, $\ell=c-0.1$, $\rho=c+1/(5s)$ and $\alpha\in[1,2]$, define
\[
 Q_j=\frac{(s-3)c^\alpha+jh^\alpha+(3-j)\ell^\alpha}{s},\qquad j=0,1,2,3.
\]
Then $Q_0<Q_1<Q_2<\rho^\alpha<Q_3$.
\end{lemma}
\begin{proof}
The $Q_j$ increase strictly in $j$. Set $\Xi=\alpha c^{\alpha-1}\ge1$. Taylor's formula and $\frac12\alpha(\alpha-1)\xi^{\alpha-2}\le1$ for $\xi\ge1$ yield
\[
 s(Q_2-c^\alpha)\le 0.1\Xi+0.03\le0.13\Xi,
 \quad s(\rho^\alpha-c^\alpha)\ge0.2\Xi,
 \quad s(Q_3-c^\alpha)\ge0.3\Xi.
\]
This establishes the two strict threshold inequalities.
\end{proof}

For later use, if a multiset has at least $w$ entries equal to $c\in[1,2]$ and at most $e$ other entries from $[1,2]$, then for $1\le\alpha\le2$ its power mean $M_\alpha$ obeys
\begin{equation}\label{lz:eq:padding}
 |M_\alpha-c|\le\frac{3e}{w+e}\le\frac{3e}{w}.
\end{equation}
Indeed the $\alpha$-power average differs from $c^\alpha$ by at most $3e/(w+e)$, and the derivative of the inverse transformation $u\mapsto u^{1/\alpha}$ is at most one for $u\ge1$.

\begin{lemma}[Global guard confinement]\label{lz:lem:guard}
If some linkage run realizes the target root split, then its exponent belongs to $[1,2]$.
\end{lemma}
\begin{proof}
All preformed groups form internally before any outside merge, because each internal linkage is $1$ and every external distance is strictly greater than $1$. For $\alpha<1$, the lower guard has linkage $M_\alpha(1.52,1.72)<1.62$ to the preformed sentinel but linkage at least $1.62$ to every pure-$L$ cluster. If it cannot merge into $R$ because $R$ was already mixed into $L$, the target has already failed. Otherwise its first attachment mixes $L$ and $R$, contradicting the target split. For $\alpha>2$, the upper guard has sentinel linkage $1.6$, whereas its safe-pair linkage $M_\alpha(248/125,136/125)>1.6$ since its quadratic mean is exactly $1.6$. Its other pure-$L$ links are at least $1.72$. The same argument applies. Both comparisons remain strict at $\pm\infty$.
\end{proof}

\begin{lemma}[Quantitative uniform successful-prefix schedule]\label{lz:lem:schedule}
Fix $\alpha\in[1,2]$ and any deterministic tie rule. Until the first $L$--$R$ merge, the run obeys the following schedule. First the six preformed groups assemble. Next each round $k=1,\ldots,N$ attaches $p_k$ to the seed selected by $E_x(\alpha)$ and $q_k$ to the opposite seed. The probes $z_t$ are processed in increasing order. $z_t$ joins $\mathcal A$ iff $C_t$ is true under the decoded assignment, and joins $R$ otherwise. The upper guard may attach to $\Sigma^+$ anywhere in the probe phase. If every clause is true, the lower guard attaches to $\Sigma^-$ and the cleanup cluster absorbs the remaining $L$-clusters before the final $L\mid R$ merge, except that a guard may instead make a premature $L$--$R$ merge at an endpoint tie $\alpha\in\{1,2\}$.
\end{lemma}
\begin{proof}
Let $\epsilon_1=3(N+1)/W$. The explicit $W$ satisfies $\epsilon_1\le\gamma/4$ and $9/W\le\Delta/4$. All comparisons below are uniform over $[1,2]$ by \eqref{lz:eq:padding} and the power-mean monotonicity $D_\alpha\ge D_1$.

\emph{Preformation.} The six internal cliques form first at score $1$; every unlisted external pair has length $2$, and every specified external distance exceeds $a-q>1.01$.

\emph{Assignments.} At the start of round $k$, the seed cardinalities are both $W+2+k-1$. Each candidate $p_k$--seed score is in $[c_k-\epsilon_1,c_k+\epsilon_1]$, and each $q_k$--seed score is in $[c_k+\gamma-\epsilon_1,c_k+\gamma+\epsilon_1]$. The next pair begins at $c_{k+1}=c_k+4\gamma$, and all nonseed competitors have score at least $1.45$ (unprocessed pair-to-pair distances exceed $1.49$). Since $c_N+\gamma+\epsilon_1<1.30$, $p_k$ must merge first. Lemma~\ref{lz:lem:choice} selects its seed and sends $q_k$ strictly to the other one, restoring equal sizes. Across the two evolving seeds at least $W^2$ cross distances remain $2$ and at most $W(2N+m+4)+(N+m+2)(N+2)$ differ; \eqref{lz:eq:padding} gives $D_\alpha(\mathcal A,\mathcal B)>1.9$. The seeds' sentinel links exceed $1.99$ on any successful prefix.

\emph{Probes.} If $z_1,\ldots,z_{t-1}$ have joined $\mathcal A$, then $|\mathcal A|=s_t$ and $D_\alpha(z_t,\mathcal A)^\alpha=Q_j$ with $j$ equal to the number of false literals in clause $C_t$. Its linkage to the pure sentinel is $\rho_t$. Lemma~\ref{lz:lem:clauses} therefore sends it to $\mathcal A$ for $j\le2$ and to $R$ for $j=3$. Every later probe has its best possible competing score at least $d_{t+1}-9/W\ge d_t+3\Delta/4$, while $z_t$ has a score at most $d_t+9/W\le d_t+\Delta/4$. Indeed, its $s_t-3$ baseline distances to $\mathcal A$ equal $d_t$, and the three exceptions lie within $0.1$. The $B$-seed link is at least $1.8-9/W>1.79$; the cleanup link is $1.72$. The lower guard has minimum link $1.62$, the two seed clusters have mutual linkage above $1.9$, and all other unrelated components have minimum link at least $1.72$.

The only possible interleaving is the upper guard. Its safe link is
\[
 D_\alpha(g^+,\Sigma^+)=M_\alpha(248/125,136/125)\in[1.536,1.6],
\]
which is strictly smaller than $D_\alpha(g^+,R)=1.6$ for $\alpha<2$. Its three-point union $G^+=\Sigma^+\cup\{g^+\}$ has link $2$ to every remaining pure-$L$ cluster except cleanup, and link to $R$ at least $(2\cdot1.6+4\cdot2)/6>1.866$ (because $\alpha\ge1$). It cannot disturb the probes. At $\alpha=2$ the tie may instead merge $g^+$ with $R$, an immediate failure that does not create an incorrect successful root split.

\emph{Lower guard.} After the last probe, $g^-$ has safe linkage $1.62$, and its sentinel linkage is
$M_\alpha(1.52,1.72)\ge1.62$, strictly greater when $\alpha>1$.
Hence it joins $\Sigma^-$ unless an endpoint tie chooses the sentinel.
Its safe union $G^-=\Sigma^-\cup\{g^-\}$ has sentinel linkage at least
$(1.52+1.72+4\cdot2)/6>1.873$. Both guard unions have linkage $2$
to other pure-$L$ clusters except cleanup.

\emph{Cleanup.} At this point the pure-$L$ clusters are $\mathcal A,\mathcal B,G^-,G^+$ and $\mathcal H$.  At each step let $\mathcal H'$ contain $\mathcal H$ and any already absorbed $L$-clusters. Since $|\mathcal H'\setminus\mathcal H|\le T$, the dominant $H_0$ cross edges of length $1.72$ and \eqref{lz:eq:padding} imply
\[
 D_\alpha(\mathcal H',X)\le1.72+3T/H_0\le1.73
 \quad\text{for every remaining pure-$L$ cluster }X.
\]
Similarly $D_\alpha(\mathcal H',R)\ge2-3T/H_0\ge1.99$. Every other $L$--$R$ link is at least $1.866$ and every other surviving $L$--$L$ link is at least $1.9$. Therefore the next four merges join $\mathcal H'$ to the other four pure-$L$ clusters, leaving precisely the two clusters $L$ and $R$. The next merge is the root merge. This proves every listed phase and all tie qualifications.
\end{proof}

\subsection{Hardness consequences and computational assumptions}
\begin{proof}[Proof of Theorem~\ref{lz:thm:zr2}]
Given a satisfying Boolean assignment $\sigma$, choose the assignment clock word $x=(-1,-\sigma_1,\ldots,-\sigma_n)$. Lemma~\ref{lz:lem:clock} gives a nonempty \emph{open} interval of exponents in $(1,2)$ realizing that word. All clock and guard inequalities are strict on a suitable subinterval. The global schedule then joins every probe to $\mathcal A$, completes both guards, and collects all of $L$ before its final merge with $R$. Conversely, suppose some exponent yields root split $L\mid R$. Lemma~\ref{lz:lem:guard} restricts it to $[1,2]$. The schedule supplies a well-defined Boolean assignment even at a clock tie. If any clause is false, its first falsified probe joins $R$ and irreversibly mixes target sides before the root. This contradicts the prescribed split. The reduction has polynomial input size and polynomial-bit rational distances as established above.
\end{proof}

\begin{corollary}[Zero-loss root-Hamming value hardness]\label{lz:cor:hamming}
For one linkage instance, deciding whether the minimum target-root-Hamming loss is zero is NP-hard. In a model in which the \emph{optimal loss value} is returned, no polynomial-time finite-factor approximation exists unless $\mathsf P=\mathsf{NP}$.
\end{corollary}
\begin{proof}
The loss equals zero precisely for the target root split. Any finite multiplicative factor must preserve an optimal value of zero.
\end{proof}

\begin{corollary}[Dasgupta-value hardness and no FPTAS]\label{lz:cor:dasgupta}
Let the linkage metric $\delta$ be as in Theorem~\ref{lz:thm:zr2}, and supply separately the complete similarity graph with edge weight $2$ within $L$ and within $R$, and weight $1$ across $L,R$. Computing the minimum Dasgupta cost over linkage exponents is NP-hard. Unless $\mathsf P=\mathsf{NP}$, no FPTAS for the \emph{optimal value} exists in this input/output model.
\end{corollary}
\begin{proof}
For a unit-weight clique on $s$ leaves, every binary hierarchy contributes the fixed cost $C(s)=(s^3-s)/3$. Let $C_*=C(|L|)+C(|R|)$. The graph $K_L\cup K_R$ has cost $C_*$ exactly for root split $L\mid R$. Otherwise one of the two target cliques crosses the root. Restricting the tree to the clique leaves and suppressing unary vertices shows that crossing the original root inflates at least one clique-edge LCA cluster size by the size of the opposite target side. Therefore the cost exceeds $C_*$ by at least $\min(|L|,|R|)\ge1$. Add the unit-weight complete graph, which contributes the tree-independent constant $B=(|S|^3-|S|)/3$, to obtain the stated similarity weights. The YES optimum is $B+C_*$ and the NO optimum at least $B+C_*+1$. Both are integers bounded by $O(|S|^3)$. An FPTAS with $\epsilon<1/(2(B+C_*+1))$ would compute the correct threshold in polynomial time, deciding \textsc{3-SAT}.
\end{proof}

\paragraph{Precision and smoothing qualifications.}
The results concern exact \emph{values} of objectives over a real scalar exponent. They do not assert $\mathrm{NP}$ membership of \textsc{ZR2}, nor that a returned arbitrary real parameter can be checked in polynomial bit complexity. Comparing sums of rational numbers raised to an arbitrary real power requires an explicit output and evaluation model. The construction uses exponentially small offsets $o_j$ and does not
establish hardness under independently perturbed metric entries. The
separate counting argument below gives polynomial expected hierarchy
complexity under bounded-density smoothing.

\subsection{Smoothed complexity across all linkage exponents}\label{sec:linkage:smoothed-proof}
We prove Theorem~\ref{thm:linkage:smoothed} for the standard
agglomerative power-mean family. Let $S$ have $n\ge2$ points and write
$E=\binom S2$. For every $e=\{u,v\}\in E$, the distance $\delta_e$
is independent, supported on $[1,2]$, and has density at most $\kappa$.
Every resulting distance matrix is a metric since
$\delta(x,z)\le2\le\delta(x,y)+\delta(y,z)$ for distinct points.
Set $Z_e=\log\delta_e\in[0,B]$, where $B=\log 2$. Its density is at
most $2\kappa$. We use fixed deterministic tie breaking.

For two disjoint nonempty clusters $U,V$, write
\[
 D_\alpha(U,V)=
 \begin{cases}
 \bigl((|U||V|)^{-1}\sum_{u\in U,v\in V}
       \delta_{uv}^{\alpha}\bigr)^{1/\alpha},&\alpha\ne0,\\
 \exp\bigl((|U||V|)^{-1}\sum_{u\in U,v\in V}
       \log\delta_{uv}\bigr),&\alpha=0.
 \end{cases}
\]
The limits at $\pm\infty$ give single and complete linkage.
For an interval set $\Theta$, write $K(S;\Theta)$ for the number
of finite exponents in $\Theta$ where the full merge hierarchy changes.

\begin{lemma}[One-edge execution bound]\label{lem:linkage:one-edge}
Fix a finite exponent $\alpha$, one primitive edge $e=\{u,v\}$,
and all other primitive distances. As $Z_e$ varies, there are at most
$n-1$ threshold values at which the complete hierarchy can change.
\end{lemma}
\begin{proof}
Until the clusters containing $u$ and $v$ coalesce, exactly one
candidate merge score depends on $Z_e$, namely the linkage between
those two clusters. Temporarily forbid that merge. All allowed
comparisons are independent of $Z_e$, producing a fixed constrained
execution with at most $n-1$ merge slots. At each slot, the forbidden
score is strictly increasing in $Z_e$, whereas the best allowed score
is fixed. The forbidden merge can therefore enter at at most one
threshold per slot. Once $u$ and $v$ coalesce, $e$ is internal to a
cluster and cannot affect subsequent linkage scores. The selected
slot determines the complete remaining hierarchy, giving the bound.
\end{proof}

\begin{lemma}[Log-linkage derivative bounds]\label{lem:linkage:log-derivative}
For a candidate merge $P=(U,V)$, let $E(P)$ be its set of cross
edges and put $r=|E(P)|\le n^2$. Define $F_P(\alpha)=\log
D_\alpha(U,V)$ and
\[
 p_f(\alpha)=\frac{e^{\alpha Z_f}}
 {\sum_{g\in E(P)}e^{\alpha Z_g}}.
\]
Then $F_P$ is analytic through $\alpha=0$, and
\[
 \frac{\partial F_P}{\partial Z_f}=p_f(\alpha),\qquad
 0\le F'_P(\alpha)\le
 \min\!\left\{\frac{B^2}{8},\frac{\log r}{\alpha^2}\right\}.
\]
The second bound at $\alpha=0$ is understood by continuity.
\end{lemma}
\begin{proof}
Let $Q_P(\alpha)=\log\bigl(r^{-1}\sum_{f\in E(P)}
 e^{\alpha Z_f}\bigr)$. Then $Q_P(0)=0$ and
$F_P(\alpha)=Q_P(\alpha)/\alpha$, whose singularity at zero is
removable. The derivative with respect to $Z_f$ follows by direct
differentiation. If $H(p)=-\sum_f p_f\log p_f$ denotes entropy, then
\[
 F'_P(\alpha)=\frac{\log r-H(p(\alpha))}{\alpha^2}.
\]
The numerator lies between zero and $\log r$. Moreover,
$Q''_P(\alpha)=\operatorname{Var}_{p(\alpha)}(Z)$, and integrating
$\frac{d}{d\alpha}(\alpha Q'_P-Q_P)=\alpha Q''_P$ gives
\[
 F'_P(\alpha)=\int_0^1 t\,
 \operatorname{Var}_{p(t\alpha)}(Z)\,dt.
\]
Since $Z\in[0,B]$, its variance is at most $B^2/4$.
This proves the uniform $B^2/8$ bound, including at zero.
\end{proof}

\begin{lemma}[Pivotal curves and extremal charging]
\label{lem:linkage:pivotal-curves}
Fix $e$ and condition on $Z_{-e}$. Away from isolated changes of
the constrained execution, an $e$-pivotal tie between two available
merges $P$ and $Q$ is locally a differentiable curve
$Z_e=\vartheta(\alpha)$. If $e$ is the largest primitive edge of $P$
when $\alpha>0$, or the smallest when $\alpha<0$, then
\[
 |\vartheta'(\alpha)|\le n^2
 \min\!\left\{\frac{B^2}{8},\frac{2\log n}{\alpha^2}\right\}.
\]
At almost every fixed $\alpha$, at most $n-1$ $e$-pivotal
curves meet the vertical line of that exponent.
\end{lemma}
\begin{proof}
At a fixed clustering state, the cross-edge sets of two different
candidate merges are disjoint. Suppose $e\in E(P)$ and
$e\notin E(Q)$. The derivative of the tie equation
$F_P(\alpha)-F_Q(\alpha)=0$ with respect to $Z_e$ is
$p_e(\alpha)>0$. The implicit function theorem therefore gives
\[
 \vartheta'(\alpha)=-
 \frac{F'_P(\alpha)-F'_Q(\alpha)}{p_e(\alpha)}.
\]
If $e$ is extremal in the stated direction, it maximizes its
softmax weight over $E(P)$, so $p_e(\alpha)\ge1/r\ge1/n^2$.
Both derivatives in the numerator are nonnegative and bounded by
Lemma~\ref{lem:linkage:log-derivative}. Their difference is at most
the larger one in absolute value. Since $r,|E(Q)|\le n^2$, the
claimed speed bound follows. The number of $e$-pivotal threshold
values at fixed $\alpha$ is bounded by
Lemma~\ref{lem:linkage:one-edge}. There are finitely many possible
clustering states and candidate pairs. Under independent continuous
distances, their distinct analytic score functions have no identical
comparison, and the relevant pivotal curves can be decomposed into
finitely many smooth pieces on each compact exponent interval,
up to null sets of degenerate inputs.
\end{proof}

\begin{proof}[Proof of Theorem~\ref{thm:linkage:smoothed}]
For each edge $e$, condition on all other log-distances. A random
horizontal line at height $Z_e$ intersects an $e$-pivotal curve
$Z_e=\vartheta(\alpha)$ in an expected number bounded by the
one-dimensional area formula. Since the conditional density of
$Z_e$ is at most $2\kappa$, Lemmas~\ref{lem:linkage:one-edge}
and~\ref{lem:linkage:pivotal-curves} give
\[
 \mathbb E[N_e^{\rm ext}(Z_e;\Theta)\mid Z_{-e}]
 \le 2\kappa(n-1)n^2
 \int_\Theta\min\!\left\{\frac{B^2}{8},
                          \frac{2\log n}{\alpha^2}\right\}d\alpha,
\]
where $N_e^{\rm ext}$ counts intersections on which $e$ has the
extremal property of Lemma~\ref{lem:linkage:pivotal-curves}.

Consider a finite exponent $\alpha_*\ne0$ at which the hierarchy
changes. At the first merge where the executions on the two sides
of $\alpha_*$ differ, two candidate merges $P,Q$ tie. Select the
largest log-distance edge in $E(P)$ for $\alpha_*>0$, or the
smallest for $\alpha_*<0$. Its value appears in $P$ and not in
$Q$, and the tie is a pivotal event for that coordinate. Under
product continuous smoothing, simultaneous independent ties and
other higher-order coincidences have probability zero. A transition
at $\alpha=0$ also has probability zero, since equality of two
geometric linkage scores is a nontrivial affine relation among
independent log-distances. Thus every actual hierarchy transition
is charged to at least one extremal pivotal event, almost surely.
Summing over $\binom n2$ edges gives the asserted integral bound
for bounded $\Theta$. Monotone convergence extends it to unbounded
interval unions. Finally,
\[
 \int_{-\infty}^{\infty}
 \min\!\left\{\frac{B^2}{8},
                     \frac{2\log n}{\alpha^2}\right\}d\alpha
   =2B\sqrt{\log n},
\]
which yields the full-domain claim.
\end{proof}

\begin{proof}[Proof of Corollary~\ref{cor:linkage:smoothed-erm}]
For two candidate merges at a fixed state, equality of their power
means at $\alpha\ne0$ is equivalent to a nonzero exponential
polynomial
\[
 \sum_{j=1}^{M}c_j e^{\beta_j\alpha}=0,
 \qquad M\le2n^2.
\]
After equal exponents are combined, such a polynomial has at most
$M-1$ real roots. The equation always has a formal root at
$\alpha=0$, where the geometric-mean scores must instead be
compared directly. Under the continuous-input assumptions,
distinct candidate merges have no identically equal score functions.

An all-real root-isolation oracle partitions a connected parameter
cell into intervals on which the order of the currently available
merge scores is fixed, together with isolated tie points. Simulate
the selected merge on each interval, combine adjacent pieces
whenever the next selected merge agrees, and recurse until the
hierarchy is determined. Simulate tie points separately under the
fixed rule. This enumerates all maximal cells of fixed complete
hierarchy, including possible boundary-only outcomes. If the
hierarchy has $K$ finite transition exponents, there are at most
$K+1$ open hierarchy cells and $K$ transition points. Compressing
unary continuations gives $O(n(K+1))$ execution-tree nodes, with
polynomially many oracle calls and comparisons per node. The limits
$\alpha=\pm\infty$ are evaluated by single and complete linkage.

For $m$ training instances, enumerate their hierarchy cells, take
the union of boundary exponents, and evaluate the empirical loss at
one representative of each resulting open cell and at each boundary
and endpoint. The expected sum of the individual transition counts
is $O(m\kappa n^5\sqrt{\log n})$, so the resulting exact-ERM
procedure uses expected polynomially many oracle operations. This
conclusion assumes exact-real access and the stated root-isolation
oracle. It does not assert polynomial bit complexity for isolating
all roots or comparing arbitrary real candidate exponents.
\end{proof}

\section{Supplementary regression capacity and tuning bounds}
The following special cases complement the ordinary Lasso validation
hardness theorem in the main text.

\begin{proposition}[Full-support scalar Lasso PARTITION reduction]
\label{prop:lasso-partition}
For full-column-rank rational Lasso instances with the positive rational
cutoff $\lambda\ge\lambda_0$, it is NP-complete to decide whether a
full-support-restricted squared validation loss can attain zero.
The same holds for
$\Phi(\lambda)=(r^\top\widehat w(\lambda)-t)^2+
\gamma(p-\|\widehat w(\lambda)\|_0)$ with any positive polynomial-bit
rational $\gamma$. This result does not concern unrestricted validation
ERM or conventional AIC.
\end{proposition}
\begin{proof}
Given positive integers $v_1,\dots,v_p$ for \textsc{Partition}, use the
polynomial-bit design of Theorem~\ref{thm:lasso-polybit}, set $u=v$,
$G=X^\top X$, $c=X^\top y$, $r=Gu$, and $t=u^\top c$.
On any full-support sign interval $s$, the KKT equation yields
$(r^\top\widehat w(\lambda)-t)^2=\lambda^2(u^\top s)^2$.
The maximal path contains one representative of every antipodal pair
$\{s,-s\}$ by Lemma~\ref{lem:lasso-antipodal}, and the chosen cutoff
intersects every full-support segment. Thus a zero-loss full-support
parameter exists exactly when $u^\top s=0$ for a sign vector, which is
precisely \textsc{Partition}. The density reward excludes every
non-full-support solution from zero loss.
For NP membership guess $s$ and check whether the one-variable rational
KKT system $s_i[G^{-1}(c-\lambda s)]_i>0$ for all $i$ admits a
$\lambda\ge\lambda_0$ with $u^\top s=0$. These are polynomial-bit
linear inequalities, so rational feasibility is polynomially verifiable.
\end{proof}
\begin{remark}[Limits of the Lasso result]
The cutoff $\lambda_0>0$ is essential to avoid an unattained
zero-limit at $\lambda\downarrow0$. The reduction from \textsc{Partition}
proves weak NP-completeness for the stated zero-loss variants, not
strong NP-hardness or hardness of AIC/unrestricted validation tuning.
\end{remark}

\begin{theorem}[Scalar ridge capacity]\label{thm:ridge-cauchy}
For every integer $p\ge2$, there are $d=\lfloor\log_2p\rfloor$ rational full-rank ridge-training/one-point-validation tasks, each with $p$ coefficients and $p$ training samples, such that their common scalar regularization-parameter family satisfies
\[
\fat_{1/2}(\F)\ge d,
\qquad
\Pdim(\F)=\Theta(\log p)
\]
for nonnegative squared validation loss. The training Gram matrices have eigenvalues bounded above and below by universal positive constants, the training data and validation data have rational encodings of polynomial bit length, and the $2^d$ parameter witnesses lie in $[1,2]$. These claims allow large validation-feature norms. Normalizing the features to unit Euclidean norm generally decreases the shattering margin.
\end{theorem}
\begin{proof}
Choose rational and pairwise distinct $q_j=1+j/(p+1)$ and set $X=\operatorname{diag}(q_1,\ldots,q_p)$ and $y=(1,\ldots,1)^\top$. The full-rank ridge solution is $w_j(\lambda)=q_j/(q_j^2+\lambda)$. Choose $p$ distinct rational parameters $\lambda_s=1+(s+1)/(p+1)$, $s=0,\ldots,p-1$, and form the $p\times p$ matrix
\[
M_{sj}=\frac{q_j}{q_j^2+\lambda_s}.
\]
After dividing each column by $q_j$, this is a Cauchy matrix whose row and column nodes are all distinct. Its determinant is nonzero by the Cauchy determinant identity. For each $i\in[d]$, choose a binary target vector $v^{(i)}\in\{0,1\}^p$ with $v_s^{(i)}=\operatorname{bit}_i(s)$ for $0\le s<2^d$. Solve $Mr^{(i)}=v^{(i)}$ exactly over the rationals and use one validation example $(r^{(i)},0)$. Its squared validation loss at each witness is
\[
f_{\lambda_s}(x_i)=\bigl((r^{(i)})^\top w(\lambda_s)\bigr)^2=(v_s^{(i)})^2=v_s^{(i)}.
\]
Thus the $d$ tasks are $1/2$-fat-shattered. All entries of $M$ have polynomial binary length, and rational Gaussian elimination yields $r^{(i)}$ of polynomial binary length, although their numerical magnitude can be large. The Gram eigenvalues $q_j^2$ lie between 1 and 4.
For the upper bound, on any full-rank $p$-variable ridge task, a one-point squared validation loss is a ratio of polynomials with numerator degree at most $2p$ and a denominator strictly positive for $\lambda>0$. After subtracting any fixed threshold and clearing the denominator, there are at most $2p$ real parameter crossings unless the resulting polynomial is identically zero. On $m$ tasks there are at most $O(mp)$ events, and the scalar switch-count argument implies $2^m\le O(mp)$ whenever $m$ tasks are pseudo-shattered. Therefore $m=O(\log p)$.
\end{proof}

\begin{proposition}[Quantitative additive ERM for strongly convex Elastic Net]\label{prop:enet-grid}
Let $\lambda_2\ge\eta>0$ be fixed and tune $\lambda_1\in[0,L]$ in the Elastic Net objective
\[
\tfrac12\|Xw-y\|_2^2+\lambda_1\|w\|_1+\tfrac{\lambda_2}{2}\|w\|_2^2.
\]
For a one-point squared validation loss $(r^\top\widehat w(\lambda_1)-t)^2$, it is Lipschitz in $\lambda_1$ with constant at most
\[
G=\frac{2\sqrt p\,\|r\|_2}{\eta}
\left(\frac{\|r\|_2\|X^\top y\|_2}{\eta}+|t|\right).
\]
For $m$ independent tasks with common $\lambda_1$ let $\overline G=m^{-1}\sum_aG_a$. Then uniform-grid ERM with mesh at most $\epsilon/\overline G$, evaluating each task to sufficient accuracy, returns an $\epsilon$-additive solution using $O(1+L\overline G/\epsilon)$ parameter evaluations. The same result holds for Lasso when $X^\top X\succeq\eta I$ on every task.
\end{proposition}
\begin{proof}
The strongly monotone KKT operator yields $\|\widehat w(\lambda)-\widehat w(\lambda')\|_2\le\sqrt p|\lambda-\lambda'|/\eta$. Taking the KKT inner product with $\widehat w$ also gives $\|\widehat w(\lambda)\|_2\le\|X^\top y\|_2/\eta$. Hence the validation prediction is Lipschitz with constant $\|r\|_2\sqrt p/\eta$ and is bounded in absolute value by $\|r\|_2\|X^\top y\|_2/\eta$. The difference-of-squares identity gives the displayed $G$. Average over tasks and compare the optimal parameter with its nearest grid point. Approximate solves introduce a separately budgeted additive error. The Lasso variant follows by using the smallest Gram eigenvalue as the strong-convexity parameter.
\end{proof}

\paragraph{Acknowledgments.}
The author thanks Yury Makarychev for early-stage discussions on linkage ERM
hardness and Yuyao Wei for early-stage discussions on Lasso ERM hardness.

\paragraph{AI methodology.}
ChatGPT was used to assist with literature review and with developing
high-level proof ideas for the computational hardness results.
The author independently verified the mathematical statements, proofs,
and references.

\end{document}